\documentclass[twocolumn,twoside]{IEEEtran}
\usepackage{psfrag}
\usepackage{amsmath}
\usepackage{amssymb}
\usepackage{graphicx}
\usepackage{subfigure}
\usepackage{verbatim}
\usepackage[thinlines,thiklines]{easybmat}
\usepackage{latexsym}
\usepackage[dvipsnames]{xcolor}
\usepackage{cite}
\usepackage{stmaryrd}
\usepackage{multirow}
\usepackage{textcomp}
\usepackage[ruled]{algorithm2e}
\IEEEoverridecommandlockouts

\newbox\tempbox

\renewcommand{\thefootnote}{\fnsymbol{footnote}}

\def\refpath{../../../Documents/ref/}

\newcommand{\bs}[1]{\ensuremath{{\boldsymbol{#1}}}}
\def\tr{\mathop{\rm tr}\nolimits}

\def\diag{\mathop{\rm diag}\nolimits}

\def\diag{\mathop{\rm diag}\nolimits}

\def\argmin{\mathop{\rm argmin}\nolimits}

\DeclareUnicodeCharacter{00A0}{~}

\newtheorem{thm}{Theorem}
\newtheorem{lem}{Lemma}

\newtheorem{assum}{Assumption}

\def\atan2{\mathrm{atan2}}

\begin{document}

	\bibliographystyle{IEEEtran}
	
\title{Gaussian Processes Model-based Control of Underactuated Balance Robots~\thanks{The preliminary version of this paper was presented in part at the 2019 IEEE International Conference on Robotics and Automation, May 20-24, 2019, Montreal, Canada. This work was partially supported by the National Science Foundation under awards CMMI-1762556 and CNS-1932370 (J. Yi).}}

\author{Kuo Chen, Jingang Yi\thanks{K. Chen and J. Yi are with the Department of Mechanical and Aerospace Engineering, Rutgers University, Piscataway, NJ 08854 USA (e-mail: {kc625@scarletmail.rutgers.edu}; {jgyi@rutgers.edu}).}, and Dezhen Song\thanks{D. Song is with the Department of Computer Science and Engineering, Texas A\&M University, College Station, TX 77843-3112, USA (e-mail: dzsong@cse.tamu.edu).}}

\maketitle

\begin{abstract}
Ranging from cart-pole systems and autonomous bicycles to bipedal robots, control of these underactuated balance robots aims to achieve both external (actuated) subsystem trajectory tracking and internal (unactuated) subsystem balancing tasks with limited actuation authority. This paper proposes a learning model-based control framework for underactuated balance robots. The key idea to simultaneously achieve tracking and balancing tasks is to design control strategies in slow- and fast-time scales, respectively. In slow-time scale, model predictive control (MPC) is used to generate the desired internal subsystem trajectory that encodes the external subsystem tracking performance and control input. In fast-time scale, the actual internal trajectory is stabilized to the desired internal trajectory by using an inverse dynamics controller. The coupling effects between the external and internal subsystems are captured through the planned internal trajectory profile and the dual structural properties of the robotic systems. The control design is based on Gaussian processes (GPs) regression model that are learned from  experiments without need of priori knowledge about the robot dynamics nor successful balance demonstration. The GPs provide estimates of modeling uncertainties of the robotic systems and these uncertainty estimations are incorporated in the MPC design to enhance the control robustness to modeling errors. The learning-based control design is analyzed with guaranteed stability and performance. The proposed design is demonstrated by experiments on a Furuta pendulum and an autonomous bikebot.
\end{abstract}


\section{Introduction}

\begin{figure*}[htb!]
	\centering
	\subfigure[]{
		\label{pendulum}
		\psfrag{T1}[][]{\small\color{yellow} \small $\theta$}
		\psfrag{T2}[][]{\small\color{yellow} \small $\alpha$}
		\includegraphics[width=1.48in]{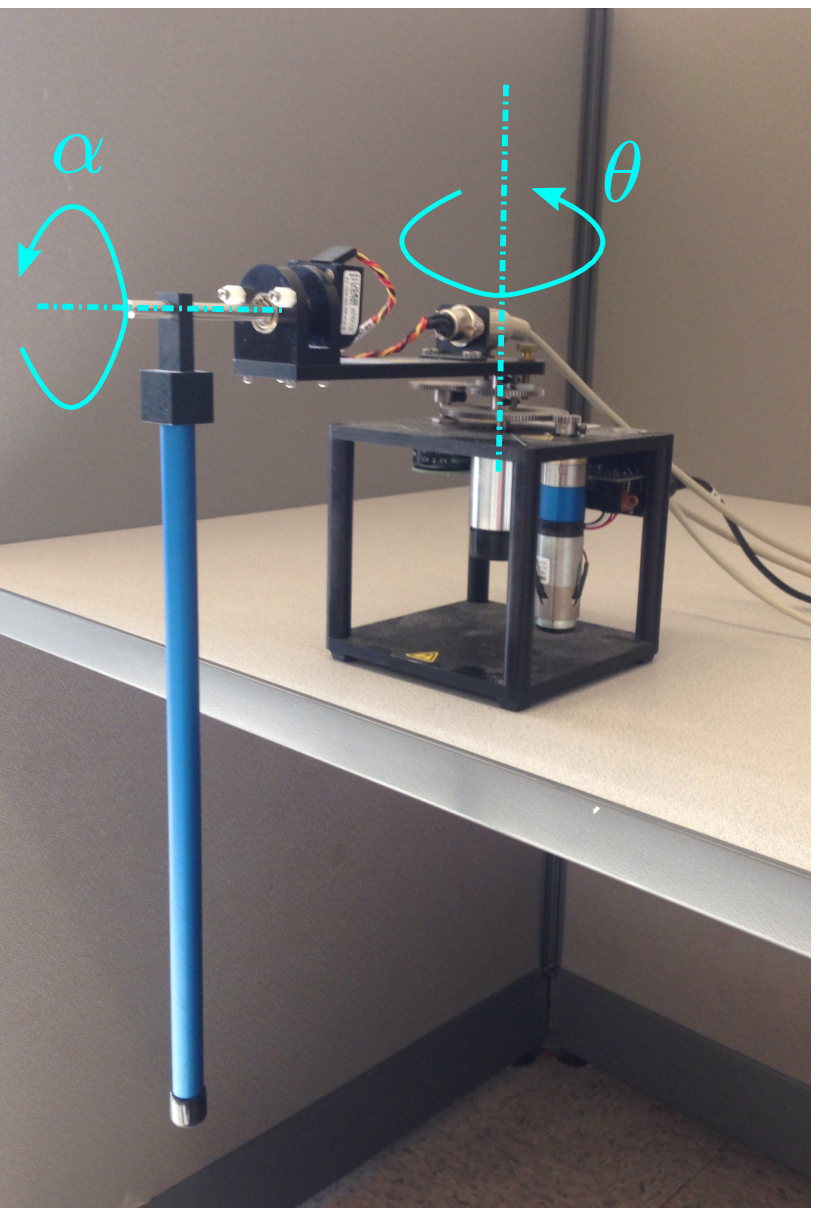}}
	\hspace{0mm}
	\subfigure[]{
		\label{bikebot}
		\includegraphics[width=3.08in]{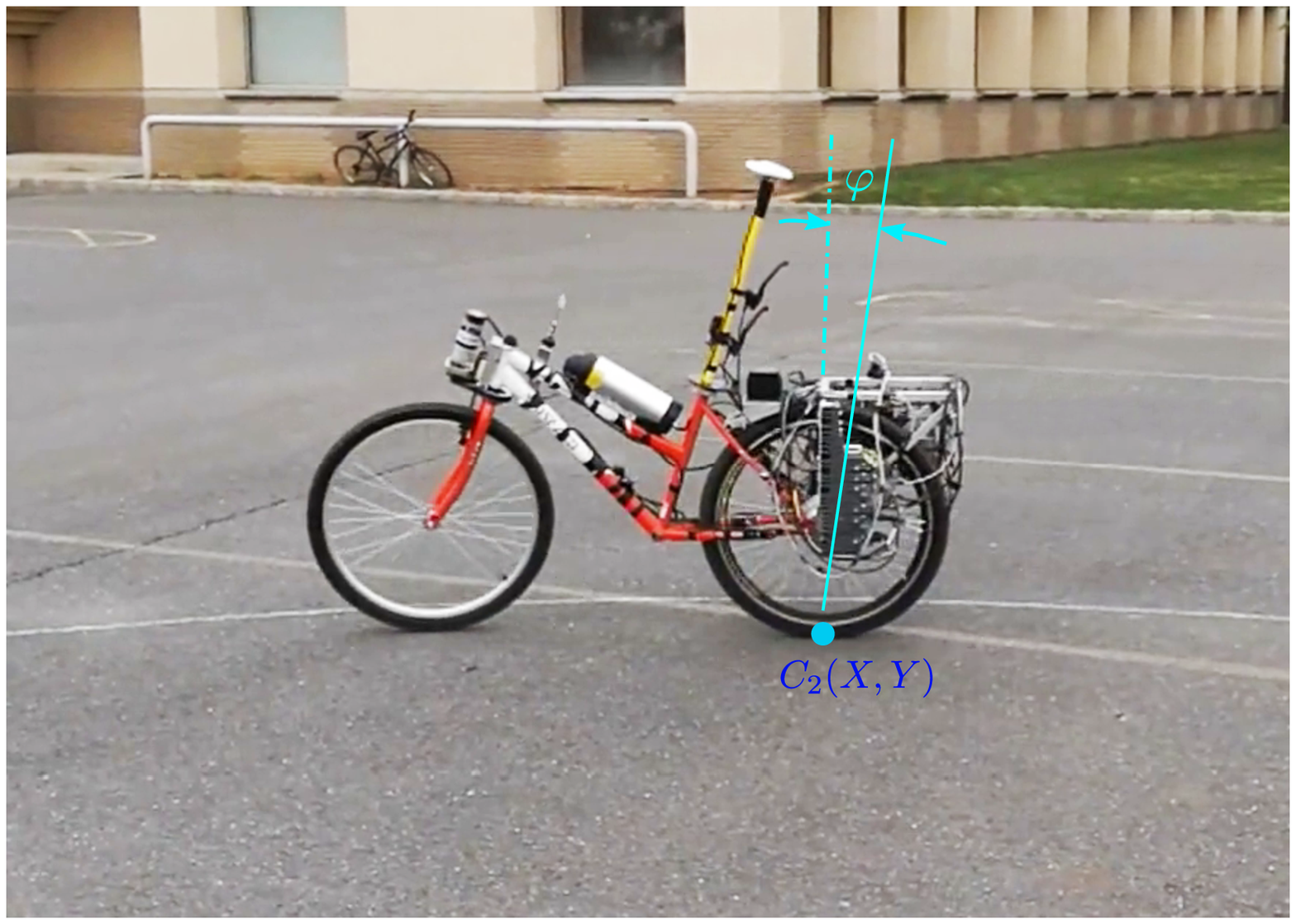}}
	\hspace{0mm}
	\subfigure[]{
	\label{examples:c}
		\includegraphics[width=1.4in]{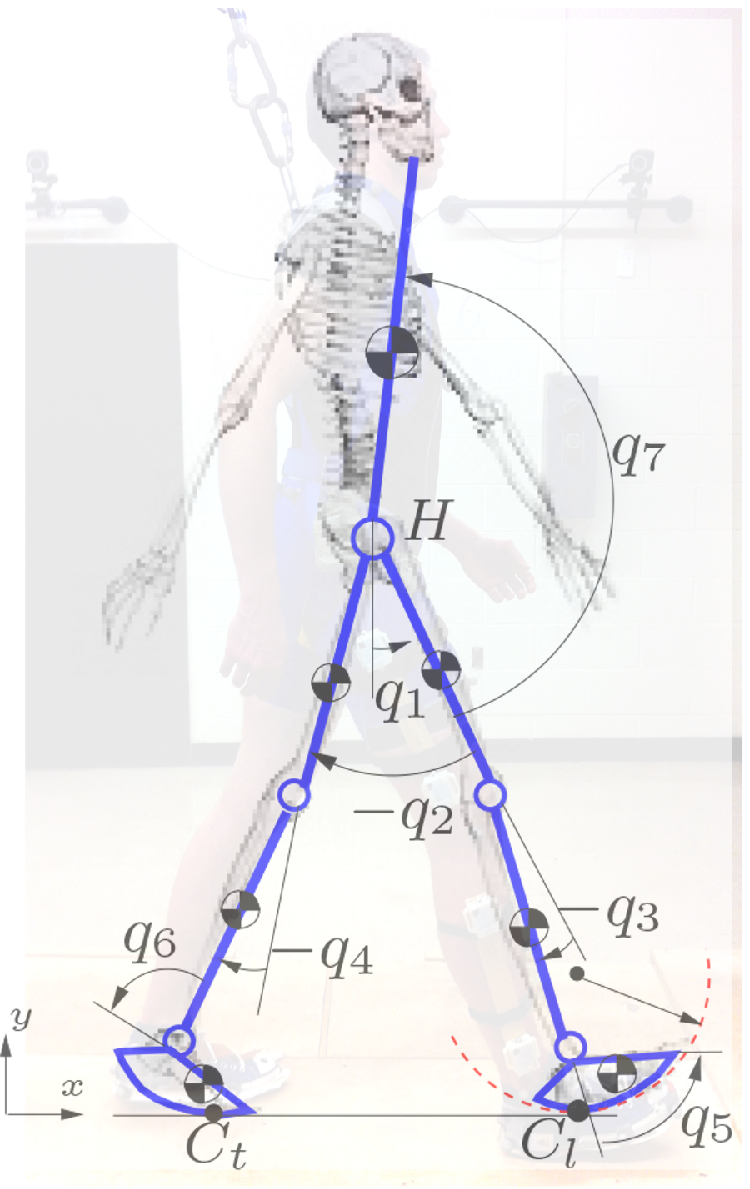}}
	\caption{A few examples of underactuated balance robotic systems. (a) Rotary inverted pendulum. Angular joint $\theta$ is the actuated DOF and joint angle $\alpha$ is the unactuated DOF. (b) The bikebot system. The robot has three DOFs (i.e., the rear wheel contact position $C_2(X,Y)$ and platform roll angle $\varphi$) and only two actuation inputs, that is, steering angle and velocity control. (c) A robotic bipedal walker. The robotic walker has seven DOFs ($q_1$-$q_7$) and six actuation inputs (i.e., double actuation at hip, knee and ankle joints) during single-stance gait.}
	\label{examples}
\end{figure*}

Underactuated systems commonly have fewer number of control inputs than the number of degree of freedom (DOF)~\cite{Choukchou2014}. Underactuated balance robots, first introduced in \cite{GetzPhD}, is a class of underactuated systems with control task of trajectory tracking for actuated subsystem, while balancing around unstable equilibra for unactuated subsystem. Cart-pole systems \cite{LeeAuto2015}, Furuta pendulums~\cite{Shiriaev2007,Park2009b,Freid2009} and autonomous bicycles~\cite{YiICRA2006,WangICRA2017} are a few examples of underactuated balance robots with the goal to balance the inverted pendulum or bikebot while the base platforms to follow desired trajectories (see Figs.~\ref{pendulum} and~\ref{bikebot}). Bipedal walkers (e.g., Fig.~\ref{examples:c}) are also a type of underactuated balance robots because the actuated joint angles are commanded to follow the desired trajectories to form certain gaits while the unactuated floating base is kept stable across steps~\cite{Wester2007,ChenACC2017,TrkovJCND2019}.

Control of underactuated balance robots faces challenges because no analytical casual compensator can achieve exactly trajectory tracking for the non-minimum phase systems~\cite{Grizzle1994}. The dynamics of the underactuated balance robotic systems can be naturally partitioned into an actuated (external) subsystem and an unactuated (internal) subsystem. An innovative control design of underactuated balance robots is to take advantages of the interaction between the external and internal subsystems. In~\cite{GetzPhD}, by observing the dependency of the balanced equilibra on trajectory tracking performance, a balance equilibrium manifold (BEM) concept is proposed to map and encode the external subsystem trajectory tracking into the desired internal subsystem profiles. A controller is then designed to stabilize the system state onto the BEM in order to achieve both tracking and balancing tasks. Despite of the mathematical elegance and guaranteed stability property, the design in~\cite{GetzPhD} requires accurate dynamics model and control robustness is not ensured to allow the robots to perform well in complex, dynamic environments. 

In recent years, using machine learning techniques, data-driven model-based controller design showed promising potentials to capture complex, high-dimensional systems dynamics and achieve superior performance over physical principle model-based controllers. Gaussian processes (GPs) are used as non-parametric machine learning models and have been widely applied to robot modeling and control~\cite{RasmussenGPML}. When they are applied to capture and model robotic system dynamics, GPs take the current robot states and control actuation and their derivatives as the learning model input and output, respectively. GP models provide differentiable and closed-form mean and covariance distributions and this property is attractive for optimization-based control designs such as model predictive control (MPC) or reinforcement learning~\cite{ko2007blimp,GPMPC_Rasmussen,PILCO,Caoquadrotor,RobustNMPC,predictcontrolGP,KollerCDC2018,McKinnonECC2019}. Compared to other dynamics learning methods, such as artificial neural network or support vector machine, GPs provide predictive covariance that can be used as a quantitative metric of model uncertainty. The covariance has also been used to design robust controllers (e.g.,~\cite{PILCO, Caoquadrotor, RobustNMPC, predictcontrolGP}).

MPC is an optimization-based preview control method. At each control step, the MPC design solves the optimal input sequence that minimizes the objective function. Computational cost is expensive for high-dimensional robotic systems dynamics. In this work, we adopt a singular perturbation method to reduce the dimensionality of the dynamic models of underacuated balance robots such that MPC is applied to the model effectively and efficiently. By transforming the underactuated balance robot dynamics into an external/internal convertible (EIC) form~\cite{GetzPhD}, the internal subsystem is feedback linearizable and the convergence rate of the error dynamics is designed to be much higher than that of the external subsystem dynamics. The internal states are then treated as the control input to the external subsystems. Taking the cart-pole system as an example, through feedback linearization, the pendulum angle is directly controlled with a desired balance angle profile that is treated as an input to the cart position dynamics. We adopt the MPC as an online planner to achieve the desired pendulum balance angle and the cart position tracking simultaneously. Both the external and internal subsystem dynamics are learned from experimental data with GPs models and the MPC trajectory planner takes the model uncertainties into the design to enhance the control robustness. We demonstrate the proposed planning and control design on the Furuta pendulum and the bikebot platforms.

The contribution of this work lies in three aspects. First, the control design is based on learning models without need of obtaining physical dynamics model and therefore, it has attractive for many complex, high-dimensional underactuated balance robotic systems. It is difficult, if not impossible, to obtain dynamic models of many of these robotic systems by physical principles. The proposed learning-based control design takes advantage of the EIC structure of systems dynamics of the underactuated balance robots~\cite{GetzPhD}. Second, the proposed control design is data efficient and effective. Most previous work relies on either the prior knowledge of the physical model or the successful demonstration from human expert or simple linear controller for efficiently training. The proposed approach takes random excitation data for model training, and then achieves successful balancing and tracking tasks. The system only needs to be excited under open-loop system control and the model is learned without any prior knowledge or successful balance demonstration. Finally, our proposed control demonstrates a novel design of explicitly incorporating the GPs model uncertainty to enhance control robustness. The design is also guaranteed stability and convergence and robustness performance.

The rest of the paper is organized as follows. Section~\ref{relatedwork} reviews relevant work. In Section~\ref{control}, we present the control systems design of the underactuated balance robots with physical models. Section~\ref{gp_p_c} extends the control design with GP models. We present the control performance analysis in Section~\ref{theory}. Experimental results are included in Section~\ref{experiment}. Finally, we summarize the concluding remarks and briefly discuss the future research directions.   

\section{Related Works}
\label{relatedwork}

We mainly review the most relevant work in research areas such as model-based control of underactuated balance robots, learned dynamics models and MPC learning schemes in robotic applications.   

\cite{GetzPhD} presented the EIC models of the underactuated balance robotic dynamics. The EIC form describes the coupling effect of the external and internal subsystem dynamics and the BEM is introduced to capture the dependency of the the internal subsystem equilibria on the external tracking performance. Dynamic inversion technique is used in~\cite{GetzPhD} to compute the BEM and the control system is proven to be asymptotically stable to a neighborhood around the desired trajectories. The work in~\cite{LeeAuto2015} formulates the EIC form in a multi-time-scale structure based on the singular perturbation theory and output feedback is achieved with extended high-gain observers. The work in~\cite{ChenCASE2017} extends the BEM approach to learning model-based control. GPs are adopted to identify the system dynamics but the dynamics structure was not successfully captured in spite of small prediction errors. The learned BEM approach demonstrates worse tracking performance than that with the physical model even though the learned model itself generates less prediction errors. The learning model-based BEM in~\cite{ChenCASE2017} is not accurately estimated due to the flexible structure of GPs and dynamic inversion does not accurately identify the BEM for the learned models. This observation motivates the work in this paper.

Learning inverse dynamics has been demonstrated in many robot control applications. A review of the model learning and robot control can be found in~\cite{NguyenCP2011}. The work in~\cite{JanInverse,gpLWR} adopt an inverse dynamics controller using global and local GPs regression models, respectively. The learned model predicts control inputs based on the robot current states and the desired derivative of robot states. Although GPs provide predictive distribution, only the mean value of the Gaussian distribution is used as the control input. The work in~\cite{Beckers} proposes a GP-based inverse dynamics control law and the feedback gain is adapted to the variance of the predictive distribution, that is, using low gains if the learned model is precise and otherwise high gains. The work in~\cite{Beckers,helwa2018provably,KollerCDC2018} give theoretically guaranteed stability or safety regions of GPs-based inverse dynamics control. Besides GPs, polynomial kernal functions are also used to predict the inverse dynamics of robotic systems (e.g.,~\cite{LiberaRAL2020}). In~\cite{DNNinverse}, deep neural network (DNN) is used to learn inverse dynamics to achieve impromptu trajectory tracking. The work in~\cite{DNNnonminimum} achieve robotic impromptu trajectory tracking for a cart-pole system and quadrotor system by learning a stable, approximate inverse of a non-minimum phase baseline system. The proposed algorithm first runs a baseline controller, usually a linear controller, to achieve the stabilization task and collect input-state data for DNN training. In training phase, the inverse model of the stabilized baseline system is learned, while in testing phase, given the desired trajectory, the learned DNN model computes a reference trajectory for the baseline system.  Under this learning-based inversion controller, the tracking performance is enhanced comparing with the baseline system. The algorithm however requires a baseline controller to stabilize the system for data collection.

Optimization-based controllers such as MPC and reinforcement learning have been applied to underactuated robot system such as cart-pole system, blimps and helicopters. In~\cite{ko2007blimp}, a learning model captures the difference between the collected acceleration data of the blimp and the prediction from the physical model so that the learning-based design does not have to build the blimp model from scratch. In~\cite{abbeel2010autonomous}, a helicopter model is learned with maneuvers and trajectories that are successfully demonstrated by human expert. By either adding prior knowledge of the robot model or learning from expert demonstration, the learned models are efficiently trained. The work in~\cite{PILCO} do not assume task-specific prior knowledge but take advantage of the probabilistic nature of Gaussian processes to achieve efficient learning. Many GP-based designs take advantage of the predicted Gaussian distribution to achieve robust control performance. For example, in~\cite{Caoquadrotor,PILCO,predictcontrolGP,KollerCDC2018,McKinnonECC2019}, the objective function is designed to include tracking errors over the prediction horizon with the variance of the predictive distribution. In~\cite{RobustNMPC}, the predictive variance is used to help reduce the feasible region for the predictive trajectory mean value. Learning-based inverse dynamics control and MPC have been demonstrated in many applications in~\cite{KoberIJRR2013,LenzRSS2015,TamarICRA2017,WilliamsICRA2017}. The works in~\cite{SchaalRAM2010,ChowNNLS2015,UmlauftCSL2018} adopt inverse dynamics controller with the global and local GPs regression models. These inverse dynamics controllers however cannot be directly applied to underactuated non-minimum phase balance robots due to the unstable internal dynamics. In this paper, we take advantages of the physical model structure of the underactuated balance robotic dynamics and use reduced-dimensional learning models to develop an computationally efficient control system. Moreover, we demonstrate the guaranteed stability and robust control performance with the GP-based design analysis.  

\section{Balance Robots Control}
\label{control}

\subsection{Notations}

Vectors $\bs{\alpha}$ and matrices $\bs{A}$ are denoted with bold lower-case and capital characters, respectively. An $n \times n$ identity matrix is denoted as $\bs{I}_n$. Estimated values of variables are denoted by symbols with hat (e.g.,  $\hat{\bs{\alpha}}$). Natural and real number sets are denoted as $\mathbb{N}$ and $\mathbb{R}$, respectively. Positive real value set and $n$-dimensional real valued vector space are denoted as $\mathbb{R}^+$ and $\mathbb{R}^n$, respectively. The smallest and largest eigenvalues of matrix $\bs{A}$ are denoted by $\lambda_{\min}(\bs{A})$ and $\lambda_{\max}(\bs{A})$, respectively. The matrix and vector norms are  defined respectively as $\|\bs{A} \|=[\lambda_{\max}(\bs{A}^T\bs{A})]^{\frac{1}{2}}$ and $\|\bs{\alpha}\|=\sqrt{\bs{\alpha}^T\bs{\alpha}}$. The metric $\|\bs{\alpha}\|_{\bs{P}}^2= \bs{\alpha}^T \bs{P} \bs{\alpha}$ is used for positive definition matrix $\bs{P}$. $\tr(\bs{A})$ and $\det(\bs{A})$ denote the trace and determinant of matrix $\bs{A}$, respectively.   

The expression $\bs{x} \sim \mathcal{N}(\bs{\mu},\bs{\Sigma})$ represents that $\bs{x}$ is a random variable satisfying Gaussian distribution with mean value $\bs{\mu}$ and covariance $\bs{\Sigma}$. The expression $\dot{\bs{x}}\sim \bs{f}(\bs{x},\bs{u})$ represents that $\dot{\bs{x}}$ is a random variable satisfying a distribution because either $(\bs{x},\bs{u})$ are random variables, $\bs{f}$ is a Gaussian process-based random function, or both. The expectation operator is denoted as $\mathbb{E}$, variable $\mathbf{\Pi}$ denotes a probabilistic event and its probability is written as $\Pr\{\mathbf{\Pi}\}$. For discrete-time MPC presentation, $k \in \mathbb{N}$ is used to denote the current time step, and $k+i$ with $i \in \mathbb{N}$ is used to denote the $i$-step forward time moment. A variable $\bs{\alpha}^*$ with a ``$*$"  superscript denotes the optimal value of the design parameter  $\bs{\alpha}$.

\subsection{Underactuated balance system control}

In this section, we present a physical model-based control system design for underactuated balance robots. The presented work will serve as a basic description of the approach that is used for GP-based design in later sections. 

An underactuated balance robotic system is described by the following dynamic model 
\begin{equation}
\bs{D}(\bs{q})\ddot{\bs{q}}+\bs{H}(\bs{q},\dot{\bs{q}})=\bs{B}(\bs{q})\bs{u},
\label{lagrangian_model}
\end{equation}
where $\bs{q}\in \mathbb{R}^{m+n}$ is the generalized coordinate of the system, $\bs{u} \in \mathbb{R}^m$  is the control input, $\bs{D}(\bs{q})$ is the inertia matrix,  $\bs{H}(\bs{q},\dot{\bs{q}})$ contains the centripetal, Coriolis and gravitational terms and $\bs{B}(\bs{q})$ is the input mapping matrix~\cite{Spong2006}. A few examples that share the above dynamic models include cart-pole systems~\cite{LeeAuto2015}, Furuta pendulums~\cite{Shiriaev2007}, bicycles and bikebots~\cite{YiICRA2006,WangICRA2017}, and bipedal walkers~\cite{Wester2007,ChenACC2017}, etc. 

Without loss of generality, coordinate $\bs{q}=[\bs{\theta}_1^T\, \bs{\alpha}_1^T]^T$ is considered to be decomposed into  generalized positions $\bs{\theta}_1 \in \mathbb{R}^m$ of the actuated subsystem and $\bs{\alpha}_1 \in \mathbb{R}^n$ of the unactuated subsystem. We assume that $m \geq n$, that is, the actuated DOF is not less than the unactuated DOF. We define generalized velocities $\bs{\theta}_2=\dot{\bs{\theta}}_1$ and  $\bs{\alpha}_2=\dot{\bs{\alpha}}_1$ such that $\dot{\bs{q}}=[\bs{\theta}_2^T \, \bs{\alpha}_2^T]^T$. Equation~(\ref{lagrangian_model}) is then partitioned into actuated and unactuated subsystems as 
\begin{equation}
\bs{D}\begin{bmatrix}
\dot{\bs{\theta}}_2 \\ \dot{\bs{\alpha}}_2
\end{bmatrix}+
\begin{bmatrix}
\bs{H}_1(\bs{q},\dot{\bs{q}}) \\ \bs{H}_2(\bs{q},\dot{\bs{q}})
\end{bmatrix}=
\begin{bmatrix}
\bs{B}_1(\bs{q}) \\
\bs{0}_{n\times m}
\end{bmatrix} \bs{u},
\label{partition}
\end{equation}
where $\bs{B}_1(\bs{q}) \in \mathbb{R}^{m\times m} $ is full rank. By inverting the mass matrix $\bs{D}(\bs{q})$ in (\ref{partition}), we obtain
\begin{equation}
\begin{bmatrix}
\dot{\bs{\theta}}_2 \\ \dot{\bs{\alpha}}_2
\end{bmatrix}
=\bs{D}^{-1}
\begin{bmatrix}
\bs{B}_1(\bs{q}) \bs{u}- \bs{H}_1(\bs{q},\dot{\bs{q}}) \\ -\bs{H}_2(\bs{q},\dot{\bs{q}})
\end{bmatrix}.
\label{invert_partition}
\end{equation}
A general state-space representation of~(\ref{invert_partition}) is formulated as
\begin{equation}
\begin{cases}
\Sigma_e: \, \dot{\bs{\theta}}_1=\bs{\theta}_2, \;
\dot{\bs{\theta}}_2 = \bs{f}_{\theta}(\bs{\theta},\bs{\alpha},\bs{u}), & \\
\Sigma_i: \, \dot{\bs{\alpha}}_1 =\bs{\alpha}_2, \;
\dot{\bs{\alpha}}_2=\bs{f}_{\alpha}(\bs{\theta},\bs{\alpha},\bs{u}), &
\end{cases}
\label{robot_dyn}
\end{equation}
where $\bs{\theta}=[\bs{\theta}_1^T\, \bs{\theta}_2^T]^T$, $\bs{\alpha}=[\bs{\alpha}_1^T\, \bs{\alpha}_2^T]^T$, and $\bs{f}_{\theta}(\cdot)$ and $\bs{f}_{\alpha}(\cdot)$ are nonlinear vector functions that represent state variables and velocity fields for external $\Sigma_e$ and internal $\Sigma_i$ subsystems, respectively. The goal of the control system is to force the external subsystem $\Sigma_e$ to track desired trajectory $\bs{\theta}_d=[\bs{\theta}_{d1}^T\, \bs{\theta}_{d2}^T]^T$, $\bs{\theta}_{d2}=\dot{\bs{\theta}}_{d1}$, while the internal subsystem $\Sigma_i$ to keep balancing around unstable equilibra. 

In~(\ref{robot_dyn}), the external subsystem $\Sigma_e$ and internal subsystem $\Sigma_i$ are coupled and considered dual relationship~\cite{GetzPhD}. For example, letting
\begin{equation}
\bs{v}=\bs{f}_{\alpha}(\bs{\theta},\bs{\alpha},\bs{u}),
\label{inverse_model}
\end{equation}
subsystem $\Sigma_i$ is feedback linearized as $\dot{\bs{\alpha}}_2=\bs{v}$. Because of $\bs{v}\in \mathbb{R}^n$ and $\bs{u}\in \mathbb{R}^m$, only a subspace of $\bs{u}$ is obtained by inverting (\ref{inverse_model}). Letting $\bs{u}=[\bs{u}_d^T\,\bs{u}_f^T]^T$, $\bs{u}_d\in \mathbb{R}^n$ and $\bs{u}_f\in \mathbb{R}^{m-n}$, $\bs{u}_d$ is obtained by an inverse dynamics method
\begin{equation}
\bs{u}_d=\bs{f}_{\alpha}^{-1}(\bs{\theta},\bs{\alpha},\bs{v},\bs{u}_f), 
\label{u_d}
\end{equation}
while $\bs{u}_f$ is freely designed. System~(\ref{robot_dyn}) under~(\ref{u_d}) becomes
\begin{equation}
\begin{cases}
\Sigma_e: \,\dot{\bs{\theta}}_1=\bs{\theta}_2, \;
\dot{\bs{\theta}}_2= \bs{f}_\theta\left(\bs{\theta},\bs{\alpha},\bs{u}(\bs{v}, \bs{u}_f)\right), & \\
\Sigma_i: \, \dot{\bs{\alpha}}_1 =\bs{\alpha}_2, \;\dot{\bs{\alpha}}_2=\bs{v}.
\end{cases}
\label{linearize_internal}
\end{equation}
In~(\ref{linearize_internal}), $\Sigma_i$ is directly controlled by $\bs{v}$ and not affected by $\Sigma_e$, while $\Sigma_e$ is affected by both inputs $\bs{u}_f$ and $\bs{v}$. 

Temporarily ignoring the tracking task of $\bs{\theta}$ for $\Sigma_e$, we design a proportional-differential (PD) controller to force $\bs{\alpha}$ to converge to desired trajectory $\bs{\alpha}_d=[\bs{\alpha}_{d1}^T \, \bs{\alpha}_{d2}^T]^T$,  $\bs{\alpha}_{d2}=\dot{\bs{\alpha}}_{d1}$, namely,  
\begin{equation}
\bs{v}_{pd}=  \dot{\bs{\alpha}}_{d2} - \frac{k_d}{\epsilon} \bs{e}_{\alpha2} - \frac{k_p}{\epsilon^2} \bs{e}_{\alpha1},   
\label{PDcontroller}
\end{equation}
where errors $\bs{e}_{\alpha1}=\bs{\alpha}_1-\bs{\alpha}_{d1}$, $\bs{e}_{\alpha2}=\bs{\alpha}_2-\bs{\alpha}_{d2}$, $\bs{e}_{\alpha}=[\bs{e}_{\alpha1}^T \,\bs{e}_{\alpha2}^T]^T$,  $\epsilon>0$ is a small positive constant called singular perturbation parameter, $k_p>0$ and $k_d>0$ are constant control gains. To enforce the tracking task for $\Sigma_e$, the desired trajectory $\bs{\alpha}_d(\bs{\theta}_d, \bs{\theta})$ is designed to be dependent on $(\bs{\theta}_d, \bs{\theta})$ such that $\bs{\theta} \rightarrow \bs{\theta}_d$ and BEM is used to capture such dependency. The BEM is defined as
\begin{equation}
\mathcal{E}=\{\bs{\alpha}_{d}=\bs{\alpha}^e_{d}:\bs{\alpha}^e_{d1}=\bs{\alpha}_{d1}(\bs{\theta}_d,\bs{\theta}), \bs{\alpha}^e_{d2}=\bs{0}\}
\end{equation}
and $\bs{\alpha}^e_{d1}$ is obtained by inverting an implicit function  
\begin{equation}
\bs{f}_{\theta}(\bs{\theta},\bs{\alpha}_{d})=\dot{\bs{\theta}}_{d2} -k_d\bs{e}_{\theta2} -k_p\bs{e}_{\theta1},  
\label{function_inversion}
\end{equation}
where errors $\bs{e}_{\theta1}=\bs{\theta}_1-\bs{\theta}_{d1}$, $\bs{e}_{\theta2}=\bs{\theta}_2-\bs{\theta}_{d2}$, and $\bs{e}_{\theta}=[\bs{e}_{\theta1}^T\,\bs{e}_{\theta2}^T]^T$. Under assumption of affine error structure, the controller in~(\ref{PDcontroller}) results in exponential convergence of $\bs{\alpha}$ and $\bs{\theta}$ to the respective neighborhoods of $\mathcal{E}$ and $\bs{\theta}_d$ simultaneously~\cite{GetzPhD}.

It is shown in~\cite{ChenCASE2017} that inverting~(\ref{function_inversion}) suffers accuracy issue for a learned model of $\bs{f}_{\theta}$.  We instead take an MPC approach to solve $\bs{\alpha}_d^0$ and obtain $\mathcal{E}$ under tracking design of $\bs{\theta}_d$. We do not directly apply MPC to~(\ref{linearize_internal}) to solve $\bs{v}$ because in that case the controlled $\Sigma_i$ might not be stable. We address the challenge of stabilizing the unstable internal subsystem $\Sigma_i$ and guarantee the stability performance through a singular perturbation design as described in the following subsection. 

\subsection{Model reduction through singular perturbation}

We apply controller~(\ref{PDcontroller}) to~(\ref{linearize_internal}) and the resulted error dynamics are
\begin{equation}
\begin{cases}
\dot{\bs{\theta}}_1=\bs{\theta}_2, \;
\dot{\bs{\theta}}_2 = \bs{f}_{\theta}(\bs{\theta},\bs{\alpha}_d+ \bs{e}_\alpha,\bs{u}(\bs{v}_{pd},\bs{u}_f)) & \\
\dot{\bs{e}}_{\alpha 1}=\bs{e}_{\alpha 2}, \; 
\dot{\bs{e}}_{\alpha 2} = -\frac{k_p}{\epsilon^2}\bs{e}_{\alpha1} -\frac{k_d}{\epsilon}\bs{e}_{\alpha2}. &
\end{cases}
\label{linearize_internal_err}
\end{equation}
As $\epsilon$ goes to zero, $\bs{e}_{\alpha1}$ and $\bs{e}_{\alpha2}$ converges to zero exponentially with a convergence rate of $-\frac{1}{\epsilon}$. The $\bs{\theta}$ dynamics are considered slow, while $\bs{e}_{\alpha}$ dynamics is referred as a fast one. By singular perturbation theory~\cite{Khalil}, it can be shown that $\|\bs{\theta}(t)-\bs{\hat{\theta}}(t)\|=O(\epsilon)$ or $\|\bs{\theta}(t)-\bs{\hat{\theta}}(t)\| \leq K \epsilon$ for a constant $K>0$, where $\hat{\bs{\theta}}(t)=[\bs{\hat{\theta}}_{1}(t)^T \,\bs{\hat{\theta}}_2(t)^T]^T$ is the solution of $\dot{\hat{\bs{\theta}}}_1 =\hat{\bs{\theta}}_2$, $\dot{\hat{\bs{\theta}}}_2  =\bs{f}_{{\theta}}(\hat{\bs{\theta}},\bs{\alpha}_d,\bs{u}(\dot{\bs{\alpha}}_{d2} ,\bs{u}_f))$.

Since estimating $\hat{\bs{\theta}}$ takes much less computational effort than obtaining $\bs{\theta}$ by~(\ref{linearize_internal_err}), we formulate the MPC state dynamic model to drive $\hat{\bs{\theta}}$ to follow $\bs{\theta}_d$, and similar to~(\ref{linearize_internal}), the estimated state dynamics are considered as 
\begin{equation}
\begin{cases}
\dot{\hat{\bs{\theta}}}_1=\hat{\bs{\theta}}_2, 
\dot{\hat{\bs{\theta}}}_2 = \bs{f}_{{\theta}}(\hat{\bs{\theta}},\hat{\bs{\alpha}},\bs{u}(\hat{\bs{w}},\bs{u}_f)), & \\
\dot{\hat{\bs{\alpha}}}_{1}=\hat{\bs{\alpha}}_{2}, 
\dot{\hat{\bs{\alpha}}}_{2}=\hat{\bs{w}} 
\end{cases}
\label{mpc_problem}
\end{equation} 
with $\hat{\bs{\alpha}}_{1}= \bs{\alpha}_{d1}$, $\hat{\bs{\alpha}}_{2}= {\bs{\alpha}}_{d2}$, and $\hat{\bs{w}}=\dot{\bs{\alpha}}_{d2}$. We define $\hat{\bs{x}}=[\hat{\bs{\theta}}^T \, \hat{\bs{\alpha}}^T]^T$ as the state variable of~(\ref{mpc_problem}). The design variable of the MPC problem is the input trajectory $\hat{\bs{w}}$, $\bs{u}_f$ and the initial values $\hat{\bs{\alpha}}_{1}(0)$ and $\hat{\bs{\alpha}}_{2}(0)$. Although the form of~(\ref{mpc_problem}) is the same as~(\ref{linearize_internal}), $\hat{\bs{\alpha}}(0)$ in~(\ref{mpc_problem}) is a design variable that needs to be determined, while $\bs{\alpha}(0)$ in~(\ref{linearize_internal}) is measured. We will present the MPC formally in Section~\ref{mpcdesign}.  

\section{GP-based Planning and Control}
\label{gp_p_c}

\subsection{GP-based inverse dynamics control for trajectory stabilization}

\renewcommand{\thefootnote}{\arabic{footnote}}

Controller~(\ref{u_d}) and dynamics~(\ref{linearize_internal}) require precise information about $\bs{f}_{\theta}$ and $\bs{f}_{\alpha}^{-1}$. We consider to use GP models to estimate them. In order to use a zero-mean Gaussian distribution in estimation, we re-write model~(\ref{linearize_internal}) as
\begin{equation}
\begin{cases}
\dot{\bs{\theta}}_1 =\bs{\theta}_2, \; \dot{\bs{\theta}}_2 = \bs{f}_{\theta}(\bs{\theta},\bs{\alpha},\bs{u}_d, \bs{u}_f ) & \\
\dot{\bs{\alpha}}_1=\bs{\alpha}_2, \; \dot{\bs{\alpha}}_2+ \bs{\kappa}_{\alpha}(\bs{\theta},\bs{\alpha},\dot{\bs{\alpha}_2},\bs{u}_f) =\bs{u}_d, &
\end{cases}
\label{robot_dyn_reshape}
\end{equation}
where $\bs{f}_{\theta}$ and $\bs{\kappa}_{\alpha}$ are unknown functions that need to be estimated. One benefit of representing the model in~(\ref{linearize_internal}) into~(\ref{robot_dyn_reshape}) is that the inverse dynamics controller becomes $\bs{u}_d=\bs{v} + \bs{\kappa}_{\alpha}(\bs{\theta},\bs{\alpha},\bs{v},\bs{u}_f) $ with $\bs{v}=\bs{v}_{pd}$ specified in~(\ref{PDcontroller}) and zero-mean GP for $\bs{\kappa}_{\alpha}$ estimation. Since $\bs{\kappa}_{\alpha}$ is estimated by a zero-mean GP model, when the testing input is far away from the training input, $\bs{\kappa}_{\alpha}$ will be close to zero and the inverse dynamics model degenerates to $\bs{u}_d=\bs{v}=\bs{v}_{pd}$. The inverse dynamics controller is stable by choosing high feedback gain in ~(\ref{PDcontroller}). By~(\ref{robot_dyn_reshape}), the learning model is formulated as
\begin{equation}
\begin{cases}
\dot{\bs{\theta}}_1 =\bs{\theta}_2, \; \dot{\bs{\theta}}_2 \sim \bs{gp}_{\theta}(\bs{\theta},\bs{\alpha},\dot{\bs{\alpha}}_2, \bs{u}_f), &\\
\dot{\bs{\alpha}}_1 =\bs{\alpha}_2, \; \bs{u}_d-\dot{\bs{\alpha}}_2 \sim \bs{gp}_{\alpha}(\bs{\theta},\bs{\alpha},\dot{\bs{\alpha}}_2,\bs{u}_f), &
\end{cases}
\label{GP}
\end{equation}
where $\bs{gp}_{\theta}$ and $\bs{gp}_{\alpha}$ are the GP distributions to estimate $\bs{f}_{\theta}$ and $\bs{\kappa}_{\alpha}$, respectively. To train these GP models, the inputs are tuple  $\{\bs{\theta}, \bs{\alpha}, \dot{\bs{\alpha}_2}, \bs{u}_f\}$ and the outputs are $\dot{\bs{\theta}}_2$ and $\bs{u}_d-\dot{\bs{\alpha}}_2$. For each output, an individual GP model is built and the GPs for different outputs are assumed independent.

With~(\ref{GP}), the control input $\bs{u}_d$ is obtained as 
\begin{equation}
\bs{u}_d \sim \bs{v}+\bs{gp}_{\alpha}(\bs{\theta},\bs{\alpha},\bs{v},\bs{u}_f),
\label{F_d}
\end{equation}
where $\bs{gp}_{\alpha}(\bs{\theta},\bs{\alpha},\bs{v},\bs{u}_f) \sim \mathcal{N}(\bs{\mu}_{\alpha},\bs{\Sigma}_{\alpha})$ is a predictive Gaussian distribution, $\bs{\mu}_{\alpha}$ and $\bs{\Sigma}_\alpha$~\footnote{We here drop dependency on $(\bs{\theta},\bs{\alpha},\bs{v},\bs{u}_f)$ for variables $\bs{\mu}_{\alpha}$ and $\bs{\Sigma}_\alpha$ for presentation convenience. For the same reason, in later presentation, we also drop dependency on $(\bs{\theta},\bs{\alpha},\bs{u}_f)$ for $\bs{\kappa}_{\alpha}$ and only leave the third argument $\bs{v}$ or $\dot{\bs{\alpha}}_2$.} are input dependent and computed from~(\ref{predictive_distribution}) in Appendix~\ref{app_GP}. Similar to~(\ref{PDcontroller}), $\bs{v}$ is designed as an inverse dynamics control for $\dot{\bs{\alpha}}_2$ as 
\begin{equation}
\bs{v} =  \hat{\bs{w}} - \frac{k_d}{\epsilon} \left[\bs{\alpha}_2 -\hat{\bs{\alpha}}_{2}(0)\right] -\frac{k_p}{\epsilon^2} \left[\bs{\alpha}_1-\hat{\bs{\alpha}}_{1}(0)\right]+\bs{r}(t),
\label{IDcontroller}
\end{equation}
where $\{ \hat{\bs{w}}, \hat{\bs{\alpha}}_{1}(0), \hat{\bs{\alpha}}_{2}(0)\} $ are solutions of the MPC design that will be given in the next section. $\bs{r}(t)$ is an auxiliary control input to be determined later in this section. By~(\ref{F_d}), $\bs{u}_d \sim \mathcal{N}(\bs{\mu}_{d},\bs{\Sigma}_{d})$ is a Gaussian distribution with $\bs{\mu}_d=\bs{v}+\bs{\mu}_{\alpha}$ and $\bs{\Sigma}_{d}=\bs{\Sigma}_{\alpha}$. The mean value $\bs{\mu}_d$ is used as the control input.

Under the inverse dynamics controllers~(\ref{F_d}) and~(\ref{IDcontroller}), we now show that the $\bs{\alpha}$ subdynamics is stabilized to $\hat{\bs{\alpha}}$. Plugging~(\ref{F_d}) into $\bs{\alpha}$ dynamics~(\ref{robot_dyn_reshape}), the closed-loop $\Sigma_i$ subsystem dynamics are
\begin{equation}
\text{\hspace{-3mm}}\begin{cases}
\dot{\bs{\alpha}}_1={\bs{\alpha}}_2, &\\
\dot{\bs{\alpha}}_2=\bs{v}+\bs{\mu}_{\alpha}-\bs{\kappa}_{\alpha}(\dot{\bs{\alpha}}_2) &
\end{cases}
\label{alpha_sub}
\end{equation}
and the error dynamics for $\Sigma_i$ are 
\begin{equation}
\dot{\bs{e}}_{\alpha}=\bs{A} \bs{e}_{\alpha} + \bs{B}[\bs{r}(t) + \bs{\mu}_{\alpha}-\bs{\kappa}_{\alpha}(\dot{\bs{\alpha}}_2) ],
\label{track_err}
\end{equation}
where 
\begin{equation}
\bs{A}=\begin{bmatrix} 0  & \bs{I}_n \\ -\frac{k_p}{\epsilon^2}\bs{I}_n & -\frac{k_d}{\epsilon}\bs{I}_n \end{bmatrix}, \bs{B}=\begin{bmatrix} 0 \\ \bs{I}_n \end{bmatrix}.
\label{matrixA}
\end{equation}
Note that $\bs{A}$ is Hurwitz when $k_p>0$ and $k_d>0$. To show convergence of $\bs{e}_{\alpha}$, it is required that the $n$-dimensional disturbance $\bs{\mu}_{\alpha}-\bs{\kappa}_{\alpha}(\dot{\bs{\alpha}}_2)$ is bounded. The disturbance terms $\bs{\mu}_{\alpha}$ and $\bs{\kappa}_{\alpha}(\dot{\bs{\alpha}}_2)$ have different inputs, that is, the latter has input $\dot{\bs{\alpha}}_2$ while the former has $\bs{v}$. {We first analyze the error $ \|\bs{\mu}_{\alpha} -\bs{\kappa}_{\alpha}(\bs{v}) \|  $. From Lemma~\ref{gp_lemma2}, the modeling error is bounded statistically, namely, for any $0<\delta<1$,
\begin{equation}
 	\Pr \{ \|\bs{\mu}_{\alpha}
 	-\bs{\kappa}_{\alpha}(\bs{v}) \| 
 	\leq \| \bs{\beta}_{\alpha}^T \bs{\Sigma}_{\alpha}^{\frac{1}{2}} \| \} \geq (1-\delta)^n,
 	\label{beta_alpha}
\end{equation}
where $\bs{\beta}_{\alpha}$ is $n$-dimensional vector with the $i$th element $\beta_{\alpha,i}=\sqrt{2\|\kappa_{\alpha,i} \|_k^2+300\gamma_{\alpha,i} \ln^3(\frac{N+1}{\delta}) }$. The following assumption is made in order to achieve deterministic statement on the convergence property. }
\begin{assum}
\label{assum1}
The modeling error of $\bs{\kappa}_{\alpha}$ is bounded for all testing inputs, i.e., 
 	\begin{equation}
 	\|\bs{\mu}_{\alpha}
 	-\bs{\kappa}_{\alpha}(\bs{v}) \| 
 	\leq \| \bs{\beta}_{\alpha}^T \bs{\Sigma}_{\alpha}^{\frac{1}{2}} \|.
 	\end{equation}
\end{assum}
Under Assumption~\ref{assum1}, the following lemma gives the bound for the disturbance term in~(\ref{track_err}), namely, $\bs{\mu}_{\alpha}-\bs{\kappa}_{\alpha}(\dot{\bs{\alpha}}_2)$. 

\begin{lem}
\label{lemma1}
Under Assumption~\ref{assum1}, the $n$-dimensional disturbance $\bs{\mu}_{\alpha}-\bs{\kappa}_{\alpha}(\dot{\bs{\alpha}}_2)$ is upper-bounded, namely, 
\begin{align}
\|\bs{\mu}_{\alpha}-\bs{\kappa}_{\alpha}(\dot{\bs{\alpha}}_2) \| \leq \rho(\bs{e}_{\alpha},\bs{\theta}), 
\end{align}
where $\rho(\bs{e}_{\alpha},\bs{\theta})=\lambda_{\min}^{-1}(\bs{A}_\kappa) \Bigl(\sum_{i=0}^2 c_i\|\bs{e}_{\alpha}\|^i+\|\bs{\beta}_{\alpha}^T \bs{\Sigma}_{\alpha}^{\frac{1}{2}}\|\Bigr)$ with $\bs{A}_\kappa=\bs{I}+\frac{\partial \bs{ \kappa}_{\alpha}}{\partial \bs{v}}$ and constants $c_i, i=0,1,2$, are defined in Appendix~\ref{proof_lemma1}. 
\end{lem}
The proof of Lemma~\ref{lemma1} is included in Appendix~\ref{proof_lemma1}. Since the disturbance term in~(\ref{track_err}) is upper-bounded, the auxiliary control term $\bs{r}(t)$ is designed according to~\cite{Spong2006} (Theorem 1 in Chapter 8.4) such that~(\ref{track_err}) is robustly stable. The following lemma gives the choice of $\bs{r}(t)$ and the convergence property of $\bs{e}_{\alpha}$.
\begin{lem}
\label{e_lemma}
Supposing $k_d^2>4k_p>0$ such that matrix $\bs{A}$ in~(\ref{matrixA}) has real eigenvalues. $\bs{A}$ is diagonalizable with $\bs{A}=\bs{M}\bs{\Lambda}\bs{M}^{-1}$, where $\bs{\Lambda}$ is the diagonal matrix and $\bs{M}$ is a non-singular matrix. The auxiliary control $\bs{r}(t)$ is designed as 
\begin{equation}
\bs{r}(t)=\begin{cases}
-\rho(\bs{e}_{\alpha},\bs{\theta}) \frac{\bs{B}^T \bs{P} \bs{e}_{\alpha}}{||\bs{B}^T \bs{P} \bs{e}_{\alpha}||}, & \text{if $||\bs{B}^T \bs{P} \bs{e}_{\alpha}||> \xi$}\\
-\frac{\rho(\bs{e}_{\alpha},\bs{\theta})}{\xi} {\bs{B}^T \bs{P} \bs{e}_{\alpha}}, & \text{if $||\bs{B}^T \bs{P} \bs{e}_{\alpha}|| \leq \xi$}
\end{cases}
\label{cond01}
\end{equation}
with constant $\xi >0$ and positive definite matrix $\bs{P}$ is the solution of the Lyapunov equation $\bs{A}^T\bs{P}+\bs{P}\bs{A}=-\bs{Q}=\bs{M}^{-T}\bs{M}^{-1}$. Under control~(\ref{cond01}) , the values of error $\|\bs{e}_\alpha(t)\|$ satisfy 
\begin{equation}
\|\bs{e}_\alpha(t)\|  \leq d_1 \| \bs{e}_{\alpha}(0)\|e^{\frac{\lambda_1}{4\epsilon}t}+d_2 , 
\label{vealpha}
\end{equation}
where $\lambda_1=\frac{-k_d+\sqrt{k_d^2-4k_p}}{2}$, $d_1=\sqrt{\frac{\lambda_{\max}(\bs{P})}{\lambda_{\min}(\bs{P})}}$, $d_2=\sqrt{-\frac{2\epsilon c_3}{\lambda_1\lambda_{\min}(\bs{P})}}$ and constant $c_3>0$ is defined in~(\ref{cond11}).
\end{lem}
The proof of Lemma~\ref{e_lemma} is given in Appendix~\ref{proof_e_lemma}. Since $\lambda_1<0$,  as positive parameter $\epsilon$ approaches to zero, term $e^{\frac{\lambda_1}{2\epsilon} t}$ converges to zero rapidly. The GP-based inverse dynamics controller derived above only uses the mean value $\bs{\mu}_d$ of the predictive distribution ($\ref{F_d}$). From Lemma~\ref{lemma1}, covariance $\bs{\Sigma}_{d}=\bs{\Sigma}_{\alpha}$ of the predictive distribution determines the disturbance error bound $\rho(\bs{e}_\alpha, \bs{\theta})$ and from Lemma~\ref{e_lemma}, $\bs{\Sigma}_{d}$ also determines the control performance of $\bs{e}_{\alpha}$. We will incorporate $\bs{\Sigma}_{d}$  information into the MPC design to enhance the control performance of $\bs{e}_{\alpha}$.

\subsection{MPC-based planning and control}
\label{mpcdesign}

Applying controllers~(\ref{F_d}) and~(\ref{IDcontroller}) to the robot dynamics model~(\ref{robot_dyn_reshape}), the closed-loop dynamics becomes
\begin{equation}
\begin{cases}
\dot{\bs{\theta}}_1=\bs{\theta}_2, \, \dot{\bs{\theta}}_2=\bs{f}_{\theta}(\bs{\theta},\hat{\bs{\alpha}}+\bs{e}_{\alpha},\bs{u}_d(\hat{\bs{w}}+\dot{\bs{e}}_{\alpha_2},\bs{u}_f),\bs{u}_f ), & \\
\dot{\bs{e}}_{\alpha}=\bs{A} \bs{e}_{\alpha} + \bs{B}[\bs{r}(t) + \bs{\mu}_{\alpha}
-\bs{\kappa}_{\alpha}(\dot{\bs{\alpha}}_2) ]. &
\end{cases}
\label{closedloop}
\end{equation}
We use $\bs{\alpha}=\hat{\bs{\alpha}}+\bs{e}_{\alpha}$ and $\dot{\bs{\alpha}}_2=\hat{\bs{w}}+ \dot{\bs{e}}_{\alpha2}$ in argument of $\bs{f}_{\theta}(\cdot)$. In the previous section, we have shown the convergence property of $\bs{e}_{\alpha}$. In this section, we discuss how to use MPC to obtain the desired internal subsystem profiles $\bs{u}_f$, $\hat{\bs{w}}$ and $\hat{\bs{\alpha}}(0)$. A learned GP model $\bs{gp}_{\theta}$ is used to predict unknown function $\bs{f}_{\theta}$ in~(\ref{closedloop}). By singular perturbation theory, assuming $\bs{e}_\alpha$ converges to zero rapidly, similar to~(\ref{mpc_problem}), we obtain the reduced system dynamics as
\begin{equation}
\begin{cases}
\dot{\hat{\bs{\theta}}}_1 =\hat{\bs{\theta}}_2, \,
\dot{\hat{\bs{\theta}}}_2 \sim \bs{gp}_{\theta}(\hat{\bs{\theta}},\hat{\bs{\alpha}},\hat{\bs{w}}, \bs{u}_f), & \\
\dot{\hat{\bs{\alpha}}}_1 =\bs{\hat{\alpha}}_{2},\,
\dot{\hat{\bs{\alpha}}}_2 =\hat{\bs{w}}. 
\end{cases}
\label{whole_gp_system}
\end{equation}

For presentation convenience, we use discrete-time representation of the above dynamics for MPC design as follows~\footnote{For notation clarity, we drop all arguments for the GP model and use $\bs{gp}_{\hat{\theta}}(k)$ to represent $\bs{gp}_{\theta}(\hat{\bs{\theta}}(k),\hat{\bs{\alpha}}(k),\hat{\bs{w}}(k), \bs{u}_f(k))$.}.
\begin{equation}
\text{\hspace{-2mm}}\begin{cases}
\Delta \hat{\bs{\theta}}_1(k) = \hat{\bs{\theta}}_2(k) \Delta t, \,
\Delta \hat{\bs{\theta}}_2(k)  \sim \bs{gp}_{\hat{\theta}}(k)\Delta t, & \\
\Delta \hat{\bs{\alpha}}_{1}(k) = \hat{\bs{\alpha}}_{2}(k) \Delta t, \, \Delta \hat{\bs{\alpha}}_{2}(k) = \hat{\bs{w}}(k) \Delta t, &
\end{cases}
\label{gp_mpc_problem}
\end{equation}
where  $\Delta t$ is the sampling period, $\Delta \hat{\bs{\theta}}_i(k)=\hat{\bs{\theta}}_i(k+1)-\hat{\bs{\theta}}_i(k)$, $\Delta \hat{\bs{\alpha}}_i(k)=\hat{\bs{\alpha}}_i(k+1)-\hat{\bs{\alpha}}_i(k)$, $i=1,2$. We use $\hat{\bs{\theta}}(k+i|k)$, $i=0,\ldots,H+1$, to denote the predicted state variable at the $(k+i)$th step given the $k$th observation $\bs{\theta}(k)$. $H$ is the prediction horizon and $\hat{\bs{\theta}}(k|k)=\bs{\theta}(k)$. We rewrite~(\ref{gp_mpc_problem}) as
\begin{equation}
\hat{\bs{\theta}}(k+i+1|k) \sim  \bs{F}\hat{\bs{\theta}}(k+i|k)+ \bs{G} \bs{gp}_{\hat\theta}(k+i),
\label{discrete_reduced}
\end{equation}
where 
\begin{equation}
\bs{F}= \begin{bmatrix}  \bs{I}_{m}  & \Delta t \bs{I}_m \\  \bs{0}_{m} & \bs{I}_m   \end{bmatrix}, \bs{G}=\begin{bmatrix}  \bs{0}_m \\ \Delta t \bs{I}_m    \end{bmatrix}.
\label{FG}
\end{equation}
$\hat{\bs{\theta}}(k+i+1|k)$ generally does not satisfy Gaussian distribution even if $\hat{\bs{\theta}}(k+i|k)$ is a Gaussian process. To make this prediction manageable, we adopt a linearization method in~\cite{PILCO} and the approximation of $\hat{\bs{\theta}}(k+i+1|k)$ is a Gaussian distribution with the mean and covariance respectively as
\begin{subequations}
\begin{align}
\text{\hspace{-4mm}}\bs{\mu}_{\hat{\theta}}(k+i+1|k)=&
\bs{F}\bs{\mu}_{\hat{\theta}}(k+i|k) 
+ \bs{G} \bs{\mu}_{gp_{\hat\theta}}(k+i),  \label{mean_var:a} \\
\text{\hspace{-4mm}}\bs{\Sigma}_{\hat{\theta}}(k+i+1|k)=&
\bs{F}\bs{\Sigma}_{\hat{\theta}}(k+i|k) \bs{F}^T 
+  \bs{G} \partial \bs{\Sigma}_{\hat{\theta}}(k+i) \bs{G}^T,  \label{mean_var:b} 
\end{align}
\label{mean_var}
\end{subequations}
\hspace{-1.3mm}where $\bs{\mu}_{gp_{\hat\theta}}$ and $\bs{\Sigma}_{gp_{\hat\theta}}$ are the mean and covariance functions of the Gaussian process $\bs{gp}_{\hat\theta}$, respectively, $\partial \bs{\Sigma}_{\hat{\theta}}(k+i)=\frac{\partial \bs{\mu}_{gp_{\theta}} }{\partial \bs{\theta} } \bs{\Sigma}_{\hat{\theta}}(k+i|k) \frac{\partial \bs{\mu}^T_{gp_{\theta}} }{\partial \bs{\theta} } +\bs{\Sigma}_{gp_{\hat{\theta}}}(k+1)$. Note that $\bs{\Sigma}_{\hat{\theta}}(k+i)$ is input dependent and $\bs{\mu}_{\hat{\theta}}(k|k)=\hat{\bs{\theta}}(k)=\bs{\theta}(k)$. 

By Lemma~\ref{bound_sigma}, it is straightforward to have $\|\bs{\Sigma}_{gp_{\hat\theta}}\| \leq \sigma^{2}_{\bs{f}\max}:=\max_{1\leq j \leq m} (\sigma_{f_{\theta_j}}^2 + \sigma_j^2)$, where $j$ is the index of the dimension of $\bs{f}_{\theta}$. The following lemma gives a bound of the state covariance $\bs{\Sigma}_{\hat{\theta}}(k+i|k)$.
\begin{lem}
\label{Sigma_theta_bound}
Assuming that $\bs{\mu}_{gp_{\hat\theta}}$ has a bounded gradient, with a small $\Delta t$, we have
\begin{equation*}
\| \bs{\Sigma}_{\hat{\theta}}(k+i|k)\| \leq i (\Delta t)^2 \|\bs{\Sigma}_{gp_{\hat\theta}}\| \leq i (\Delta t)^2\sigma^{2}_{\bs{f}\max}.
\end{equation*}
\end{lem}
The proof of Lemma~\ref{Sigma_theta_bound} is given in Appendix~\ref{proof_Sigma_theta_bound}. For the reduced system dynamics~(\ref{whole_gp_system}), the objective function of the MPC is first considered as 
\begin{align}
\bar{J}^k_{\hat{\theta},\hat{W}_H}=&\sum_{i=0}^{H} \Bigl[ \mathbb{E}\|\bs{e}_{\hat{\theta}}(k+i)\|^2_{Q_1}+\|\hat{\bs{w}}(k+i)\|^2_{R}+\|\hat{\bs{\alpha}}(k)\|^2_{Q_2}\nonumber \\
& +\|\bs{u}_f(k+i)\|^2_{R}\Bigr]+\mathbb{E}\|\bs{e}_{\hat{\theta}}(k+H+1)\|^2_{Q_3} \nonumber \\
=&\sum_{i=0}^{H} l_s(k+i)+l_f(k+H+1)+\|\hat{\bs{\alpha}}(k)\|^2_{Q_2},
\label{e_obj}
\end{align}
where $\bs{e}_{\hat{\theta}}(k+i)=\hat{\bs{\theta}}(k+i|k)-\bs{\theta}_d(k+i)$, matrices $\bs{Q}_i$, $i=1,2,3$, and $\bs{R}$ are positive definite. In~(\ref{e_obj}), the stage cost $l_s(j)$, $j=k+i$, is defined as
\begin{align}
\text{\hspace{-2mm}}l_s(j)=&\mathop{\mathbb{E}}[\|\bs{e}_{\hat{\theta}}(j)\|^2_{{Q}_1}]
+\|\hat{\bs{w}}(j)\|^2_{{R}} +\| \bs{u}_f(j)\|^2_{{R}}\nonumber\\
=&\|\bs{e}_{\bs{\mu}_{\hat{\theta}}}(j)\|^2_{{Q}_1} +\tr(\bs{Q}_1\bs{\Sigma}_{\hat{\theta}}(j|k)) 
+ \| \hat{\bs{w}}(j)\|^2_{{R}}+ \nonumber \\
& \| \bs{u}_f(j)\|^2_{{R}}, \label{stagecost}
\end{align}
where $\bs{e}_{\bs{\mu}_{\hat{\theta}}}(j):=\bs{\mu}_{\hat{\theta}}(j|k)-\bs{\theta}_d(j)$. Similarly, the terminal cost $l_f(k+H+1)$ is defined as
\begin{align}
&l_f(k+H+1)=\mathop{\mathbb{E}}[\|\bs{e}_{\hat{\theta}}(k+H+1)\|^2_{{Q}_3}] \nonumber\\
=&\|\bs{e}_{\bs{\mu}_{\hat{\theta}}} (k+H+1)\|^2_{{Q}_3}
+ \tr(\bs{Q}_3\bs{\Sigma}_{\hat{\theta}}(k+H+1|k)).
\label{finalcost}
\end{align}
The $k$th-step MPC input variable is
\begin{equation}
\hat{\bs{W}}(k)=\{ \hat{\bs{\alpha}}(k), \hat{\bs{w}}(k+i), \bs{u}_f(k+i),  i=0,\ldots,H\}.
\label{MPCinput}
\end{equation}
We take expectation operator in~(\ref{e_obj}) because $\bs{\theta}(k+i)$ is approximated by the probabilistic variable $\hat{\bs{\theta}}(k+i|k)$ from~(\ref{mean_var}). 

The distribution dynamics~(\ref{mean_var}) is used to predict the future trajectory and this gives computational benefit. The objective function~(\ref{e_obj}) however does not penalize  $\bs{\alpha}$ convergence. In fact, the convergence of $\bs{e}_{\alpha}$ affects $\bs{\theta}$ tracking performance as shown in~(\ref{closedloop}). To include the penalty on the internal subsystem tracking performance, we modify the MPC objective function as
\begin{equation}
J^k_{\hat{\theta},\hat{W}_H}=\bar{J}^k_{\hat{\theta},\hat{W}_H}+ \nu \|\bs{\Sigma}_d(k)\|, 
\label{MPC_cost}
\end{equation}
where $\bs{\Sigma}_d(k) $ is the covariance of the predictive distribution~(\ref{F_d}) at the $k$th step and $\nu>0$ is a weighting factor. The rationale to include $\bs{\Sigma}_d$  in the cost function is to incorporate the inverse dynamics model uncertainty in the MPC design. As shown in Lemmas~\ref{lemma1} and~\ref{e_lemma}, the convergence property of $\bs{e}_{\alpha}$ depends on $\bs{\Sigma}_d$ values. For a small value of $\bs{\Sigma}_d$, the MPC picks up the desired trajectory that can be stabilized by the inverse dynamics controller with high confidence. The significance of adding term $\bs{\Sigma}_d$ into the objective function will be demonstrated in Section~\ref{experiment}. 

The control input by the MPC design is denoted as 
\begin{equation}
\hat{\bs{W}}^*(k)=\argmin_{\hat{\bs{W}}(k)} J^k_{\hat{\theta},\hat{W}}.
\label{optimal_input}
\end{equation}
The optimization is formulated as an unconstrained MPC and solved with gradient decent method. The control input $\hat{\bs{W}}^*(k)$ is used in the inverse dynamics controller~(\ref{IDcontroller}). Two remarks need to be clarified before the MPC convergence is shown rigorously. First, the approximated model~(\ref{mean_var}) is used instead of the inaccessible model~(\ref{closedloop}) to compute the state prediction. The impact of using this approximation on tracking stability will be discussed in Section~\ref{theory}. Second, although prediction $\bs{\mu}_{\hat{\theta}}$ from~(\ref{mean_var}) is an accurate approximation of $\bs{\theta}$ for~(\ref{closedloop}), the convergence of  $\bs{\mu}_{\hat{\theta}}$ to the desired $\bs{\theta}_d$ under controller~(\ref{optimal_input}) is not straightforward and needs to be further clarified. 

The rest of this subsection is devoted to address the second item above. It should be noted that since the prediction model~(\ref{mean_var:a}) for $\bs{\mu}_{\hat{\theta}}$ is exact, no difference exists between $\bs{\mu}_{\hat{\theta}}(k+i|k)$ and $\bs{\mu}_{\hat{\theta}}(k+i|k+j)$, $j\leq i$, in the discussion of the convergence of $\bs{\mu}_{\hat{\theta}}$ to $\bs{\theta}_d$. The input given by~(\ref{optimal_input}) does not automatically guarantee the convergence of $\bs{\mu}_{\hat{\theta}}$ to $\bs{\theta}_d$ because of the finite prediction horizon. As shown in~\cite{Chen1998}, the stability is instead ensured with the appropriate choice of the terminal cost $l_f(k+H+1)$ and the terminal constraint. We here briefly describe the terminal cost design to ensure this convergence. 

Suppose that for the desired trajectory $\bs{\theta}_d$, there exists a corresponding inputs $\{\bs{\alpha}_d,\bs{w}_d, \bs{u}_{f,d}  \} $ satisfying the mean propagation dynamics~(\ref{mean_var:a}), that is,
\begin{equation}
\bs{\theta}_d(k+i+1)=\bs{F}\bs{\theta}_d(k+i)+\bs{G}\bs{\mu}_{gp_{\hat{\theta}}}(\bs{\theta}_d,\bs{\alpha}_d ,\bs{w}_d,\bs{u}_{f,d}).
\label{desire_dynamics}
\end{equation}
To show the stability of tracking error $\bs{e}_{\bs{\mu}_{\hat{\theta}}}= \bs{\mu}_{\hat{\theta}} -\bs{\theta}_d $ under~(\ref{optimal_input}), we assess $\bs{e}_{\bs{\mu}_{\hat{\theta}}}$ dynamics by taking the difference between~(\ref{desire_dynamics}) and~(\ref{mean_var:a}), namely, 
\begin{align}
\bs{e}_{\bs{\mu}_{\hat{\theta}}}(k+i+1)=&\bs{F}\bs{e}_{\bs{\mu}_{\hat{\theta}}}(k+i)
+\bs{G}[\bs{\mu}_{gp_{\hat{\theta}}}(\bs{\mu}_{\hat{\bs{\theta}}}, \hat{\bs{\alpha}} ,\hat{\bs{w}},\bs{u}_{f}) \nonumber \\
&-\bs{\mu}_{gp_{\hat{\theta}}}(\bs{\theta}_d,\bs{\alpha}_d,\bs{w}_d,\bs{u}_{f,d})].
\label{mpc_error}
\end{align}
Defining the input $\bs{u}_e=[\hat{\bs{\alpha}}^T - \bs{\alpha}_d^T \; \hat{\bs{w}}^T-\bs{w}_d^T \; \bs{u}_f^T-\bs{u}_{f,d}^T]^T$, (\ref{mpc_error}) is then linearized around its equilibrium point at the origin and we obtain 
\begin{equation}
\bs{e}_{\bs{\mu}_{\hat{\theta}}}(k+i+1)=\bs{A}_e  \bs{e}_{\bs{\mu}_{\hat{\theta}}}(k+i) + \bs{B}_e \bs{u}_e(k+i),
\label{cond44}
\end{equation}
$\bs{A}_e=\bs{F}+\bs{G}\frac{\partial \bs{\mu}_{gp_\theta}}{\partial \bs{\theta}_d}$, and 
$\bs{B}_e=\bs{G}\bigl[\frac{\partial \bs{\mu}^T_{gp_\theta} }{\partial \bs{\alpha}_d} \frac{\partial \bs{\mu}^T_{gp_\theta} }{\partial \bs{w}_d}  \frac{\partial \bs{\mu}^T_{gp_\theta} }{\partial \bs{u}_{f,d}} \bigr]^T$.

By~\cite{Chen1998}, stability of the error dynamics~(\ref{mpc_error}) is guaranteed by the solution $\hat{\bs{W}}^{\circledast}(k)$ of the following MPC problem
\begin{equation}
\hat{\bs{W}}^{\circledast}(k)=\argmin_{\hat{\bs{W}}(k)}{J}^{k*}_{\hat{\theta},\hat{W}},
\label{mpc_origin_objective}
\end{equation}
where ${J}^{k*}_{\hat{\theta},\hat{W}}= \sum_{i=0}^{H} l_s^*(k+i)+l_f^*(k+H+1)$, 
\begin{subequations}
\begin{align}
l_s^*(k+i)=&\|\bs{e}_{\bs{\mu}_{\hat{\theta}}}(k+i)\|^2_{{Q}_1^*} + \| \bs{e}_{\hat{\alpha}}(k+i)\|^2_{{Q}_2^*}+\nonumber \\
&\|\Delta \hat{\bs{w}}(k+i)\|^2_{{R}^*}+\|\Delta \bs{u}_d\|^2_{{R}^*}, \label{def31:a} \\
l_f^*(k+H+&1)= \|\bs{e}_{\bs{\mu}_{\hat{\theta}}}(k+H+1)\|^2_{{Q}_3^*}, \label{def31:b} 
\end{align}
\label{def31}
\end{subequations}
\hspace{-1.3mm}$\bs{e}_{\hat{\alpha}}(k+i)=\hat{\bs{\alpha}}(k+i)-\bs{\alpha}_d(k+i)$, $\Delta \hat{\bs{w}}(k+i)=\hat{\bs{w}}(k+i)-\bs{w}_d(k+i)$, and $\Delta \bs{u}_d=\bs{u}_f(k+i)-\bs{u}_{f,d}(k+i)$. Positive definite matrices $\bs{Q}_i^*$, $i=1,2,3$, and $\bs{R}^*$ are chosen for design specification. \cite{Chen1998} proposed a systematic approach to design the terminal cost matrix $\bs{Q}^*_3$ and the corresponding terminal region $\bs{\Omega}_e$.  Within $\bs{\Omega}_e$, a linear state feedback controller $\bs{u}_e = -\bs{K}_e \bs{e}_{\bs{\mu}_{\hat{\theta}}}$ (with gain $\bs{K}_e$) for~(\ref{cond44}) ensures the stability of the original dynamics~(\ref{mpc_error}) with the decreasing terminal cost, that is, if $\bs{e}_{\bs{\mu}_{\hat{\theta}}}(k+H+1) \in \bs{\Omega}_e$, controller  $\bs{u}_e$ results in $\bs{e}_{\bs{\mu}_{\hat{\theta}}}(k+H+2) \in\bs{\Omega}_e$ with $l_f^*(k+H+2)\leq l_f^*(k+H+1)-l_s^*(k+H+1)$.

Taking the MPC objective function~(\ref{mpc_origin_objective}) under the optimal input, we have
\begin{align*}
{J}^{(k+1)*}_{\hat{\theta}^*,\hat{W}^{\circledast}}&-{J}^{k*}_{\hat{\theta}^*,\hat{W}^{\circledast}}
=-l_s^*(k)+l_f^*(k+H+2)-\\
&l_f^*(k+H+1)+l_s^*(k+H+1) \leq -l_s^*(k).
\end{align*}
From the monotonicity of ${J}^{k*}_{\hat{\theta}^*,\hat{W}^{\circledast}}$, namely, $\|\bs{e}_{\bs{\mu}_{\hat{\theta}}}(k)\|^2_{\bs{Q}_1^*} \leq {J}^{k*}_{\hat{\theta},\hat{W}} \leq \| \bs{e}_{\bs{\mu}_{\hat{\theta}}}(k)\|^2_{\bs{Q}_3^*}$, we have 
\begin{equation*}
{J}^{(k+1)*}_{\hat{\theta}^*,\hat{W}^{\circledast}} \leq \left[1-\frac{\lambda_{\min}(\bs{Q}_1^*)}{\lambda_{\max}(\bs{Q}_3^*)}\right]
{J}^{k*}_{\hat{\theta}^*,\hat{W}^{\circledast}}. 
\end{equation*}
Comparing the MPC problem~(\ref{optimal_input}) with~(\ref{mpc_origin_objective}), we notice two major differences. The first one is that the former one includes the model uncertainty through the covariance term (i.e., $\nu \|\bs{\Sigma}_d(k)\|$). The second difference is that the former does not need the desired input trajectories $\bs{\alpha}_d$, $\bs{w}_d$ and $\bs{u}_{f,d}$, which are difficult to obtain. The MPC problem~(\ref{optimal_input}) only assumes that the desired trajectories exist but no need to be known. This is one of the attractive properties of the proposed control design. 

To apply the result of~(\ref{mpc_origin_objective}) to show the stability of the MPC design in~(\ref{optimal_input}), the following lemma is needed. 
\begin{lem}
\label{lemma_terminal}
For terminal cost $l^*_f(k+H+1)$ and the stage cost $l^*_s(k+i)$ defined in~(\ref{def31}), let matrices $\bs{Q}_1$ and $\bs{R}$ in~(\ref{stagecost}) and $\bs{Q}_3$ in~(\ref{finalcost}) satisfy $\bs{Q}_1=\bs{Q}_1^*$, $\lambda_{\max}(\bs{R})<\lambda_{\min}(\bs{R}^*)$ and $\bs{Q}_3=\bs{Q}_3^*$, then
\begin{align*}
l_f(k+H+2) &\leq l_f(k+H+1)+\tr(\bs{Q}_3 \bs{\Sigma}_{\hat{\theta}}(k+H+2))\\
&-l_s(k+H+1)+\tr(\bs{Q}_1 \bs{\Sigma}_{\hat{\theta}}(k+H+1))
\end{align*}
if the following conditions are satisfied 
\begin{subequations}
	\begin{align}
	& \|\hat{\bs{w}}(k+H+1)\|\geq \lambda_R \|\bs{w}_d(k+H+1)\|, \\
	& \|\bs{u}_f(k+H+1)\|\geq \lambda_R \|\bs{u}_{f,d}(k+H+1)\|,
	\end{align}
	\label{input_domain}
\end{subequations}
\hspace{-1.5mm} where $\lambda_R=\frac{2\lambda_{\min}(\bs{R}^*)}{\lambda_{\min}(\bs{R}^*)-\lambda_{\max}(\bs{R})}$.
\end{lem}
The proof of this lemma is included in Appendix~\ref{proof_lemma_terminal}. With the result in Lemma~\ref{lemma_terminal}, we obtain the bound of tracking error $\bs{e}_{\bs{\mu}_{\hat{\theta}}}(k+i):=\bs{\mu}_{\hat{ \theta} }(k+i|k)-\bs{\theta}_d(k+i)$, $i=0,\ldots,H+1$, through the following lemma.
\begin{lem}
\label{mpc_lemma}
Using $J^{k}_{\hat{\theta},\hat{W}}$ {under} the optimal input~(\ref{optimal_input}) {as the Lyapunov function candidate}, the tracking error satisfies 
{$\|\bs{e}_{\bs{\mu}_{\hat{\theta}}}(k+i)\|\leq
	a_4(i)\|\bs{e}_{\theta}(k)\|+a_5(i)$
	where 
$a_4(i)=d_3^{\frac{i}{2}}\sqrt{\frac{\lambda_{\max}(\bs{Q}_3)}{\lambda_{\min}(\bs{Q}_1)} }$,
$a_5(i)=\sqrt{\frac{d_3^i(\alpha_{\max}^2+ \nu \sigma_{\kappa \max}^2)+d_4 \frac{1-d_3^i}{1-d_3}}{\lambda_{\min}(\bs{Q}_1)  }}$,
$0<d_3=1-\frac{\lambda_{\min}(\bs{Q}_1)}{\lambda_{\max}(\bs{Q}_3)} <1$, $d_4=m \lambda_m(H+2)(\Delta t)^2 \sigma^2_{\bs{f}\max}+(1+\frac{\lambda_{\min}(\bs{Q}_1)}{\lambda_{\max}(\bs{Q}_3)})(\nu \sigma^2_{{\kappa}\max}+\alpha_{\max}^2)$, and  
$\lambda_m=\lambda_{\max}(\bs{Q}_1)+\lambda_{\max}(\bs{Q}_3)$.}
\end{lem}
The proof of Lemma~\ref{mpc_lemma} is included in Appendix~\ref{proof_mpc_lemma}. Because $0<d_3<1$, $a_4(i)$ converges to zero and $a_5(i)$ converges to $\sqrt{ \frac{d_4}{(1-d_3)\lambda_{\min}(\bs{Q}_1)} }$  exponentially as $i$ goes to infinity. Since $J^{k}_{\hat{\theta},\hat{W}}$ is positive definite, Lemma~\ref{mpc_lemma} confirms that if we choose $J^{k}_{\hat{\theta},\hat{W}}$ as the Lyapunov function candidate, the values of $J^{k}_{\hat{\theta},\hat{W}}$ decrease along the trajectory predicted from model~(\ref{mean_var}) as long as~(\ref{input_domain}) holds. This implies that by solving the MPC problem~(\ref{optimal_input}), the  mean value variable $\bs{\mu}_{\hat{\theta}}$ predicted by~(\ref{mean_var}) is stabilized to track $\bs{\theta}_d$ exponentially.

\begin{figure}[htb!]
	\centering
	\includegraphics[width=3in]{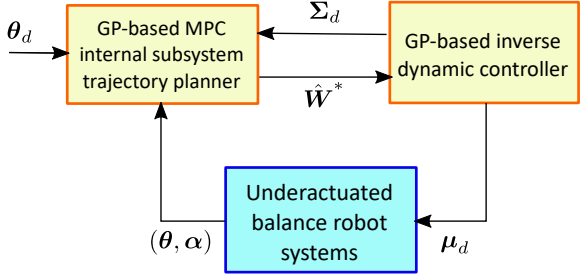}
	\vspace{-0mm}
	\caption{Schematic flow of the GP-based control framework.}
	\label{fig_gp_control_framework}
	\vspace{-3mm}
\end{figure}

In summary, Fig.~\ref{fig_gp_control_framework} illustrates the framework of GP-based control design. In each control step, the trajectory planner solves the MPC problem~(\ref{optimal_input}) with model~(\ref{mean_var}) and generates the planned internal subsystem trajectory $\hat{\bs{W}}^*$. The MPC also incorporates the predictive variance $\bs{\Sigma}_d$ from  the inverse dynamics model. The inverse dynamics controller takes the $\hat{\bs{W}}^*$ profile and uses~(\ref{F_d}) and~(\ref{IDcontroller}) to compute the control input $\bs{\mu}_d$. In the framework, the prediction uncertainty (i.e., $\bs{\Sigma}_d$) is used in both the MPC-based trajectory planner and the inverse dynamics stabilization. 

\section{Control Performance Analysis}
\label{theory}

In this section, we show stability and performance analysis of the control design and also discuss the impact of learning-based model errors on controller performance.

Under the learning-based controller~(\ref{F_d}) and~(\ref{IDcontroller}), we have the closed-loop dynamics~(\ref{closedloop}). Assuming that both models~(\ref{robot_dyn_reshape}) and (\ref{closedloop}) are deterministic, we consider a Lyapunov function candidate 
\begin{equation}
V(k)=V_{\theta}(k)+\zeta V_{\alpha}(k),
\label{lyap0}
\end{equation}
where constant $\zeta >0$, $V_{\alpha}(k)=\bs{e}_{\alpha}^T(k) \bs{P} \bs{e}_{\alpha}(k)$, $\bs{P}$ is defined in Lemma~\ref{e_lemma}, and $V_{\theta}(k)$ is similar to the MPC cost function in~(\ref{e_obj}) without expectation operator and under the optimal control $\hat{\bs{W}}^*(k)$, namely, 
\begin{align}
V_{\theta}(k)=&\bar{J}_{{\theta},\hat{W}^*}^k=\sum_{i=0}^{H} \bigl[ \|\bs{e}_{\theta}(k+i)\|^2_{Q_1}+\|\hat{\bs{W}}^*(k+i)\|^2_{R}\nonumber \\
&+\|\bs{u}^0_f(k+i)\|^2_{R}\bigr]+\|\hat{\bs{\alpha}}^0(k)\|^2_{Q_2}\nonumber \\
&+\|\bs{e}_{\theta}(k+H+1)\|^2_{Q_3}.
\label{e_obj2}
\end{align}
Here $\bs{e}_{\theta}(k+i)= \bs{\theta}(k+i)-\bs{\theta}_d(k+i)$. Note that $\bar{J}^k_{\theta,\hat{W}^*}$ is a quadratic function of the actual state $\bs{\theta}(k+i)$ following the unknown deterministic model~(\ref{closedloop}) under $\hat{\bs{W}}^*$ given by~(\ref{optimal_input}). At the $k$th step, it is impossible to directly evaluate $\bar{J}_{\theta,\hat{W}^*}^k$ because inaccessible future states and the unknown model~(\ref{closedloop}), and instead its value is approximated by $\bar{J}_{\hat{\theta}^*,\hat{W}^*}^k$ given by~(\ref{e_obj}).

\begin{figure}[htb!]
	\centering
	\includegraphics[width=2.9in]{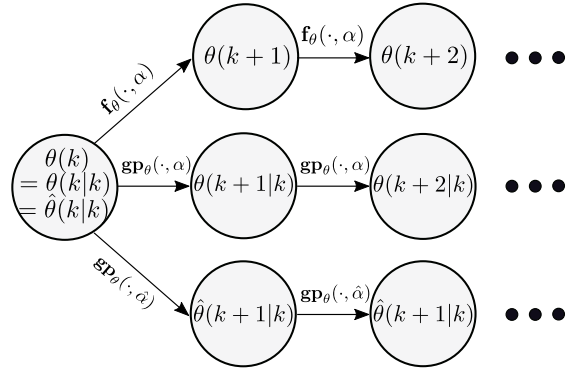}
	\vspace{-0mm}
	\caption{Flow chart of the state estimation by three predictive models.}
	\label{fig_prediction_trj}
	\vspace{-0mm}
\end{figure}

We assess the decreasing property of the proposed Lyapunov function candidate as
\begin{align}
\text{\hspace{-3mm}}\Delta V(k)&=\left(\bar{J}_{{\theta},\hat{W}^*}^{k+1}-\bar{J}_{\hat{\theta}^*,\hat{W}^*}^{k+1}\right)-\left(\bar{J}_{{\theta},\hat{W}^*}^{k}-\bar{J}_{\hat{\theta}^*,\hat{W}^*}^{k}\right) \nonumber \\
&+\left(J_{\hat{\theta}^*,\hat{W}^*}^{k+1}-J_{\hat{\theta}^*,\hat{W}^*}^{k} \right) +\zeta \left[V_\alpha(k+1)-V_\alpha(k)\right] \nonumber \\
&-\nu \left[\|\bs{\Sigma}_d({\hat{W}^*}(k+1))\| -\|\bs{\Sigma}_d({\hat{W}^*}(k))\| \right],
\label{lya_decrease}
\end{align}
where $\Delta V(k)=V(k+1)-V(k)$ and $J^k_{\hat{\theta}^*,\hat{W}^*}=\bar{J}^k_{\hat{\theta}^*,\hat{W}^*}+ \nu \|\bs{\Sigma}_d(\hat{W}^*(k))\|$ are used in the above expansion. In~(\ref{lya_decrease}), term $\bar{J}_{\theta,\hat{W}^*}^k-\bar{J}_{\hat{\theta}^*,\hat{W}^*}^{k}$ quantifies the difference between the approximated cost-to-go and the actual cost-to-go at the $k$th step. We use $\bs{\theta}(k+i|k)$ for $i\geq 0$ to denote the predicted value of $\bs{\theta}(k+i)$ given the measured $\bs{\theta}(k)$ with the initial condition $\bs{\theta}(k|k)=\bs{\theta}(k)$. Similar to~(\ref{discrete_reduced}), the evolution of $\bs{\theta}(k+i|k)$ follows discretized form of~(\ref{GP}), namely,
\begin{equation}
\bs{\theta}(k+i+1|k)\sim
\bs{F}\bs{\theta}(k+i|k)+\bs{G} \bs{gp}_{\theta}(k+i),
\label{discrete_full}
\end{equation}
with the mean value $\bs{\mu}_{\theta}(k+i+1|k)$ and variance $\bs{\Sigma}_{\theta}(k+i+1|k)$ calculations similar to~(\ref{mean_var}). The difference between models~(\ref{discrete_full}) and~(\ref{discrete_reduced}) is that the former depends on the actual internal state $\bs{\alpha}(k+i)$, while the latter uses the estimated internal state $\hat{\bs{\alpha}}(k+i|k)$. Model~(\ref{discrete_reduced}) is actually used for $\bs{\theta}$ trajectory prediction through the MPC formulation. Figure~\ref{fig_prediction_trj} further illustrates the relationships among the three different $\bs{\theta}$-prediction models~(\ref{robot_dyn_reshape}),~(\ref{discrete_full}) and~(\ref{discrete_reduced}).

We now quantify the difference between $\hat{\bs{\theta}}(k+i|k)$ and $\bs{\theta}(k+i|k)$. At $i=0$, $\hat{\bs{\theta}}(k|k)=\bs{\theta}(k|k)=\bs{\theta}(k)$. The difference between $\hat{\bs{\theta}}(k+i|k)$ and $\bs{\theta}(k+i|k)$ comes from the difference between the reduced model~(\ref{discrete_reduced}) and full model~(\ref{discrete_full}). We have the following results about their differences.
\begin{lem}
\label{err1_lemma}
Assuming the mean value of the predictive distribution~\footnote{For presentation convenience, we drop the third and four arguments and use notation $\bs{\mu}_{gp_{\theta}}(\bs{\mu}_{\theta},\bs{\alpha})$ to represent $\bs{\mu}_{gp_{\theta}}(\bs{\mu}_{\theta},\bs{\alpha},\dot{\bs{\alpha}}_2,\bs{u}_f)$.} $\bs{\mu}_{gp_{\theta}}(\bs{\mu}_{\theta},\bs{\alpha})$ is Lipshitz in $\bs{\mu}_{\theta}$ and $\bs{\alpha}$, namely,
\begin{align*}
&\|\bs{\mu}_{gp_{\theta}}(\cdot,\bs{\alpha})- \bs{\mu}_{gp_{\theta}}(\cdot,\hat{\bs{\alpha}})\|\leq L_2\|\bs{e}_{\alpha}  \|, \\
&\|\bs{\mu}_{gp_{\theta}}(\bs{\mu}_{\theta},\cdot)-\bs{\mu}_{gp_{\theta}}(\bs{\mu}_{\hat{\theta}},\cdot)\|\leq L_3 \|\bs{\mu}_{\theta}-\bs{\mu}_{\hat{\theta}}\|,
\end{align*}
with constants $L_2,L_3 >0$, $\tilde{\bs{\mu}}_\theta(k+i):=\bs{\mu}_{{\theta}}(k+i|k)-\bs{\mu}_{\hat{\theta}}(k+i|k)$ satisfies $\|\tilde{\bs{\mu}}_\theta(k+i)\| \leq \varrho_{\hat{\theta}}(i)\|\bs{e}_{\alpha}(k)\|+\varrho_2(i)$, 
where
\begin{equation*}
\varrho_{\hat{\theta}}(i)=d_1 L_2 \Delta t \left[\left(\frac{1-a_1^i}{1-a_1}-i\right)\left(1-\frac{L_3\Delta t}{1-a_1}\right)+i\right],
\end{equation*}
$a_1=e^{\frac{\lambda_1}{4\epsilon} \Delta t}$ and $\varrho_2(i)=d_2 L_2\Delta t  [i+ \frac{1}{2} L_3 \Delta t (i-1)i]$. $d_1$, $d_2$, $\lambda_1$ are defined in Lemma~\ref{e_lemma}.
\end{lem}
The proof of this lemma is included in Appendix~\ref{proof_err1_lemma}. We then inspect the difference between ${\bs{\theta}}(k+i|k)$ and $\bs{\theta}(k+i)$. The difference between $\bs{\theta}(k+i|k)$ and $\bs{\theta}(k+i)$ comes from the difference between the learning model~(\ref{discrete_full}) and the unknown actual model~(\ref{robot_dyn_reshape}) as shown in Fig.~\ref{fig_prediction_trj}. From Lemma~\ref{gp_lemma2}, we obtain the GP learned prediction guaranteed to be closed to the $m$-dimensional model $\bs{f}_{\theta}$ with high probability~\footnote{We here drop all arguments $(\bs{\theta},\bs{\alpha},\dot{\bs{\alpha}}_2,\bs{u}_f)$ of functions $\bs{f}_{\theta}$ and $\bs{\Sigma}_{gp_{\theta}}$ for presentation brevity.}, namely,
\begin{equation}
\text{\hspace{-3mm}}\Pr \{ \|\bs{\mu}_{gp_\theta}(\bs{\theta},\bs{\alpha})
-\bs{f}_{\theta} \|  \leq \| \bs{\beta}_{\theta}^T \bs{\Sigma}_{gp_{\theta}}^{\frac{1}{2}} \| \} \geq (1-\delta)^m,
\label{beta_theta}
\end{equation}
where $0< \delta <1$ and $\bs{\beta}_{\theta}$ is an $m$-dimensional vector with its $j$th element $\beta_{\theta,j}= \sqrt{ 2 \|f_{\theta,j}\|_k^2 +300 \gamma_{\theta,j} \ln^3(\frac{N+1}{\delta} ) }$. $\gamma_{\theta,j}$ is the maximum information gain for $f_{\theta,j}$ ($j$th element of $\bs{f}_{\theta}$). To conduct the performance analysis, the following assumption is considered.
\begin{assum}
\label{assum2}
The modeling error of $\bs{f}_{\theta}$ is bounded for all testing inputs, namely,
	\begin{equation}
	 \|\bs{\mu}_{gp_\theta} \left(\bs{\theta} ,\bs{\alpha}  \right)
	-\bs{f}_{\theta}  \|  
	 \leq \| \bs{\beta}_{\theta}^T \bs{\Sigma}_{gp_{\theta}}^{\frac{1}{2}}  \|.
	\label{assumption_thetamu}
	\end{equation}
\end{assum}

Under Assumption~\ref{assum2}, the following lemma gives upper-bound of $\bs{\theta}_\mu(k+i):=\bs{\mu}_{\theta}(k+i|k)-\bs{\theta}(k+i)$.
\begin{lem}
\label{err2_lemma}
Under Assumption~\ref{assum2}, we have $\| \bs{\theta}_\mu(k+i)\| \leq  \varrho_{\mu_\theta}(i)$,
where $\varrho_{\mu_\theta}(i)=\Delta t \sum_{j=0}^{i-1} \|\bs{\beta}_{\theta}^T \bs{\Sigma}_{gp_\theta}^{\frac{1}{2}}(k+j|k)\|$.
\end{lem}
The proof of this lemma is included in Appendix~\ref{proof_err2_lemma}. Lemmas~\ref{err1_lemma} and~\ref{err2_lemma} give the error bounds on $\tilde{\bs{\mu}}_{\theta}(k+i)=\bs{\mu}_{\theta}(k+i|k)-\bs{\mu}_{\hat{\theta}}(k+i|k)$ and $\bs{\theta}_\mu(k+1)=\bs{\mu}_{\theta}(k+i|k)-\bs{\theta}(k+i)$, respectively. Combining these results, we have the  error bound of $\tilde{\bs{\theta}}_\mu(k+i):=\bs{\mu}_{\hat{\theta}}(k+i|k)-\bs{\theta}(k+i)$ as 
\begin{align*}
\| \tilde{\bs{\theta}}_\mu(k+i) \| & \leq \|\tilde{\bs{\mu}}_{\theta}(k+i) \|+\| {\bs{\theta}}_\mu(k+i)   \| \\
& \leq \varrho_{\hat\theta}(i)\|\bs{e}_{\alpha}(k)\|+
\varrho_2(i)+\varrho_{\mu_\theta}(i).
\end{align*}
By defining $a_2(i)=\varrho_2(i)+\varrho_{\mu_\theta}>0$, we obtain that $ \| \tilde{\bs{\theta}}_\mu(k+i)\| \leq \varrho_{\hat\theta}(i)\|\bs{e}_{\alpha}(k)\|+a_2(i)$. We estimate the difference of $\bar{J}_{\hat{\theta}^*,\hat{W}^*}^{k}-\bar{J}_{\theta,\hat{W}^*}^k$ in the following lemma with proof given in Appendix~\ref{proof_Edifflemma}.
\begin{lem}
\label{Edifflemma}
Under Assumptions~\ref{assum1} and~\ref{assum2}, we obtain
\begin{equation*}
|\bar{J}_{\hat{\theta}^0,\hat{W}^0}^{k}-\bar{J}_{\theta,\hat{W}^0}^k|
\leq \rho_{J}(\bs{e}_{\alpha},\bs{e}_{\theta}),
\end{equation*}
where 
\begin{align}
\rho_{J}(\bs{e}_{\alpha},\bs{e}_{\theta})&=\lambda_{\max}(\bs{Q}_3) \sum_{i=0}^{H+1}\Bigl\{ \bar{\xi}_1(i) 
\|\bs{e}_{\alpha}(k)\|^2 + \nonumber \\
& \bar{\xi}_3(i)\|\bs{e}_{\alpha}(k)\|+\bar{\xi}_2(i) \|\bs{e}_{\alpha}(k)\|  \|\bs{e}_{\theta}(k)\|+ \nonumber \\
& \bar{\xi}_4(i)\|\bs{e}_{\theta}(k)\|+\bar{\xi}_5(i)\Bigr\},
\label{cond66}
\end{align}
  $\bar{\xi}_1(i)=\varrho^2_{\hat\theta}(i)$, $\bar{\xi}_2(i)=2\varrho_{\hat\theta}(i)a_4(i)$, $\bar{\xi}_3(i)=2\varrho_{\hat\theta}(i)[a_2(i)+a_5(i)]$, $\bar{\xi}_4(i)=2a_2(i)a_4(i)$, and $\bar{\xi}_5(i)=a_2(i)(a_2(i)+2a_5(i))+mi(\Delta t)^2\sigma^2_{\bs{f}\max}$.
$\varrho_{\hat\theta}(i)$ is defined in Lemmas~\ref{err1_lemma}, $a_4(i)$ and $a_5(i)$ are defined in Lemma~\ref{mpc_lemma}.
\end{lem}

The result in Lemma~\ref{Edifflemma} is used for the first two pairs of terms of $\Delta V(k)$ in~(\ref{lya_decrease}). Letting  $\bs{e}(k)=[\bs{e}_{\theta}^T(k) \; \bs{e}_{\alpha}^T(k)]^T$ denote the error vector, it is straightforward to obtain that the Lyapunov function candidate  $V(k)$ in~(\ref{lyap0}) satisfies $\underline{\lambda} \| \bs{e}(k) \|^2 \leq  V(k)\leq \overline{\lambda} \| \bs{e}(k) \|^2$, where $\underline{\lambda}= \min(\lambda_{\min}(\bs{Q}_1), \zeta \lambda_{\min}(\bs{Q}) )$ and $\overline{\lambda}= \max(\lambda_{\max}(\bs{Q}_1), \zeta \lambda_{\max}(\bs{Q}) )$, where matrices $\bs{Q}$ and $\bs{Q}_1$ are defined in Lemma~\ref{e_lemma} and~(\ref{e_obj}), respectively. We are now ready to give the following main result.
\begin{thm}
\label{stability}
For parameters $\bar{\xi}_j(i)$, $i=0,1,\ldots,H+2$, $j=1,\ldots,5$, given in Lemma~\ref{Edifflemma}, defining $\xi_j=\bar{\lambda}\left[\bar{\xi}_j(0)+2\sum_{i=1}^{H+1}\bar{\xi}_j(i)+\bar{\xi}_j(H+2)\right]$, $\gamma_1=\sqrt{\eta}$, $\gamma_2=\frac{\xi_3}{2\gamma_1}$, $\gamma_3=\sqrt{\lambda_{\min}(\bs{Q}_1)}$, $\gamma_4=\frac{\xi_4}{\gamma_3}$, and $\gamma_5 =\frac{\xi_4^2}{\gamma_3^2}+\frac{\xi_3^2}{4\gamma_1^2}+\xi_5+ \hat{\alpha}_{\max}^2 + \nu \sigma_{\bs{\kappa}\max}^2+\zeta c_3 \Delta t + m \lambda_{m}(H+2)(\Delta t)^2 \sigma^2_{\bs{f}\max}$, where $\lambda_m=\lambda_{\max}(\bs{Q}_1)+\lambda_{\max}(\bs{Q}_3)$ and 
\begin{equation}
\eta=\frac{1}{4}\zeta \lambda_{\min}(\bs{Q})\Delta t -\xi_1 - \frac{\xi_2^2}{2 \lambda_{\min}(\bs{Q}_1)} -\frac{\lambda_{\min}(\bs{Q}_1)}{4} >0,
\label{parameter_condition_theory}
\end{equation}
the following property is then held 
\begin{equation}
 V(k+1)\leq \gamma_\lambda V(k)+\gamma_5 
 \nonumber
\end{equation}
where $0< \gamma_\lambda=1-\frac{\gamma_3^2}{4\overline{\lambda}} < 1$.
\end{thm}
The proof of Theorem~\ref{stability} is given in Appendix~\ref{proof_stability}. If $ V(k+1)\leq \gamma_\lambda V(k)+\gamma_5$ holds for $i$ consecutive steps, we have 
\begin{equation*}
V(k+i)\leq \gamma_\lambda^i V(k) + \frac{4\gamma_5\overline{\lambda}(1-\gamma_\lambda^i)}{\gamma_3^2}.
\end{equation*}
Introducing the static state values $V_{ss}=\lim_{i\to\infty} V(k+i) $ and  $\|\bs{e}\|_{ss}=\lim_{i\to\infty} \|\bs{e}(k+i)\| $ for any fixed $k$, then $V_{ss} \leq \frac{4\overline{\lambda}}{\gamma_3^2} \gamma_5 $  and    $ \| \bs{e} \|_{ss} \leq \sqrt{\frac{4\overline{\lambda}}{\gamma_3^2 \underline{\lambda}} \gamma_5 }$. 

Theorem~\ref{stability} implies that the error magnitude $\|\bs{e} \|$ decreases exponentially until $ \| \bs{e} \|_{ss} \leq \sqrt{\frac{4\overline{\lambda}}{\gamma_3^2 \underline{\lambda}} \gamma_5 }$. Parameter condition~(\ref{parameter_condition_theory}) can be satisfied by choosing small enough value for singular perturbation parameter $\epsilon$. As $\epsilon$ value is small, $\lambda_{\min}(\bs{Q})$ becomes large according to Lemma~\ref{e_lemma} and $\varrho_{\hat{\theta}}$ goes small according to Lemma~\ref{err1_lemma} and henceforth both $\xi_1$ and $\xi_2$ values are small. Modeling errors are also important factors for control performance. As the error bound $\|\bs{\beta}_{\alpha}^T\bs{\Sigma}_{\alpha}^{\frac{1}{2}} \|$ for $\bs{\kappa}_\alpha$ increases, values of $d_2$ and $\varrho_2(i)$ increase, $a_2(i)$ increases, $\bar{\xi}_3,\bar{\xi}_4,\bar{\xi}_5$ increase, $\gamma_5$ increases, and finally the bound of $\|\bs{e}\|_{ss} $ increases. As the error bound $\|\bs{\beta}_{\theta}^T\bs{\Sigma}_{gp_\theta}^{\frac{1}{2}} \|$ for $\bs{f}_{\theta}$ increases, values of $\varrho_{\mu_\theta}(i)$ and $a_2(i)$ increase, therefore both $\gamma_5$ value and the bound of $\|\bs{e} \|_{ss}$ increase. The results in Theorem~\ref{stability} are obtained under Assumptions~\ref{assum1} and~\ref{assum2}. With enough training data for the learning model, $\delta$ defined in Lemma~\ref{gp_lemma} can be chosen small so that Assumptions~\ref{assum1} and~\ref{assum2} are satisfied practically.

\section{Experiments}
\label{experiment}

\begin{figure*}[htb!]
\hspace{-2mm}
	\subfigure[]{
		\label{bikebot3}
		\includegraphics[width=2.2in]{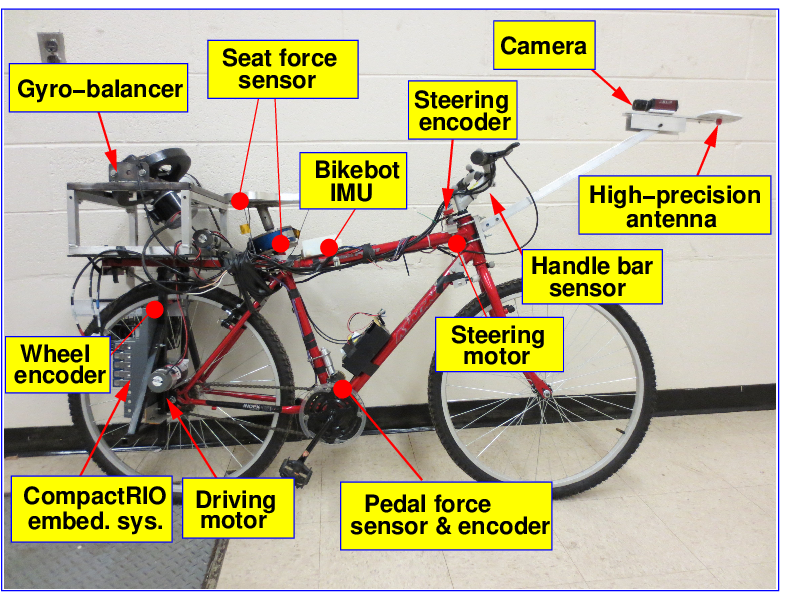}}
	\hspace{-1mm}
	\subfigure[]{
	\label{bikebot4}
	\includegraphics[width=2.47in]{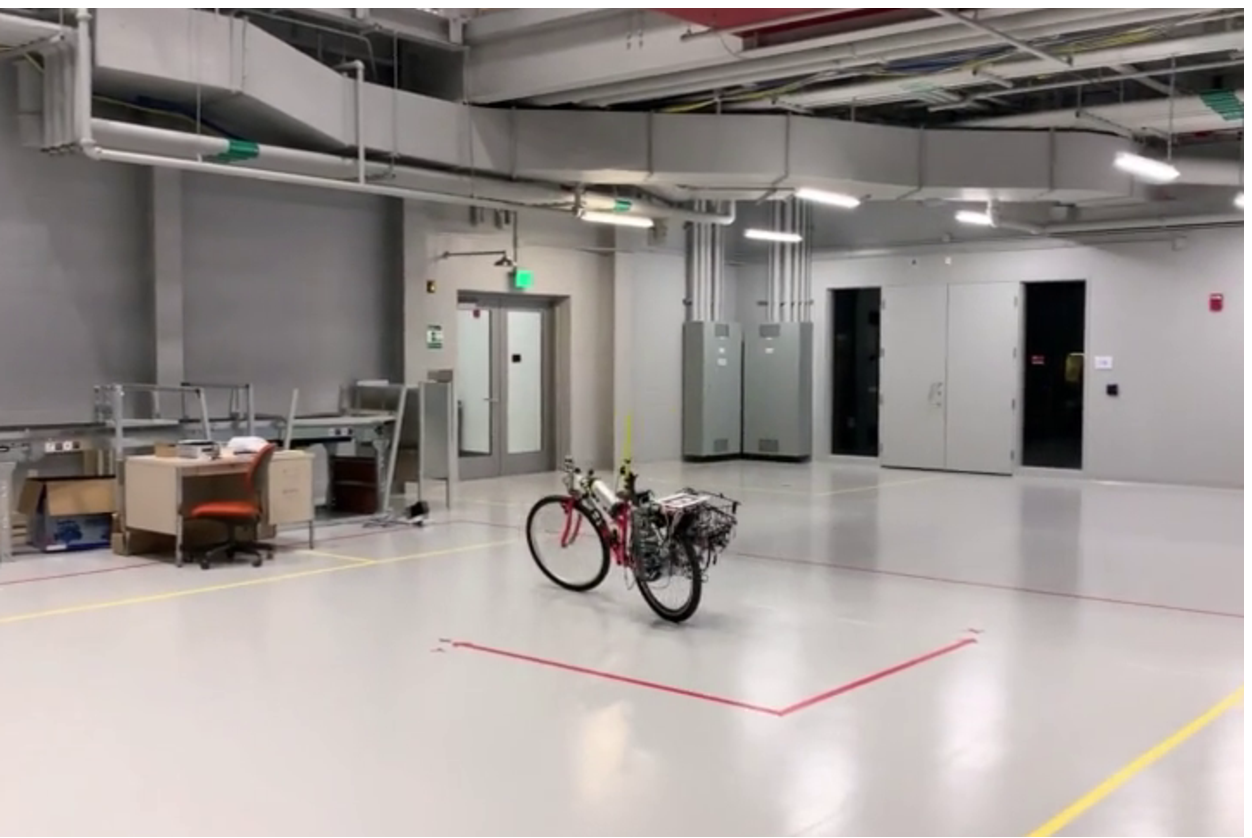}}
\hspace{-0mm}
	\subfigure[]{
		\label{bikebot_sketch}
		\psfrag{a1}[][]{\scriptsize $x$}
		\psfrag{a11}[][]{\scriptsize  $x$}
		\psfrag{a2}[][]{\scriptsize  }
		\psfrag{a3}[][]{\scriptsize  }
		\psfrag{b1}[][]{\scriptsize  }
		\psfrag{b2}[][]{\scriptsize  }
		\psfrag{b3}[][]{\scriptsize  }
		\psfrag{the}[][]{\scriptsize  $\varphi$}
		\psfrag{the3}[][]{\scriptsize  }
		\psfrag{phi2}[][]{\scriptsize  $\psi$}
		\psfrag{alf}[][]{\scriptsize  }
		\psfrag{a11}[][]{\scriptsize  $\phi$}
		\psfrag{b}[][]{\scriptsize  $l_b$}
		\psfrag{h}[][]{\scriptsize  $h_b$}
		\psfrag{bbb}[][]{\scriptsize  $l$}
		\psfrag{h5}[][]{\scriptsize  }
		\psfrag{ca}[][]{\scriptsize  $\xi$}
		\psfrag{c1}[][]{\scriptsize  }
		\psfrag{C1}[][]{\scriptsize  $C_1$}
		\psfrag{C2}[][]{\scriptsize  $C_2$}
		\psfrag{G}[][]{\scriptsize  $G$}
		\psfrag{R2}[][]{\scriptsize  $\mathcal{B}$}
		\psfrag{RR}[][]{\scriptsize  $\mathcal{R}$}
		\psfrag{R}[][]{\scriptsize  $\mathcal{N}$}
		\psfrag{d1}[][]{\scriptsize  $X$}
		\psfrag{d2}[][]{\scriptsize  $Y$}
		\psfrag{d3}[][]{\scriptsize  $Z$}
		\psfrag{d5}[][]{\small }
		\includegraphics[width=1.8in,height=1.75in]{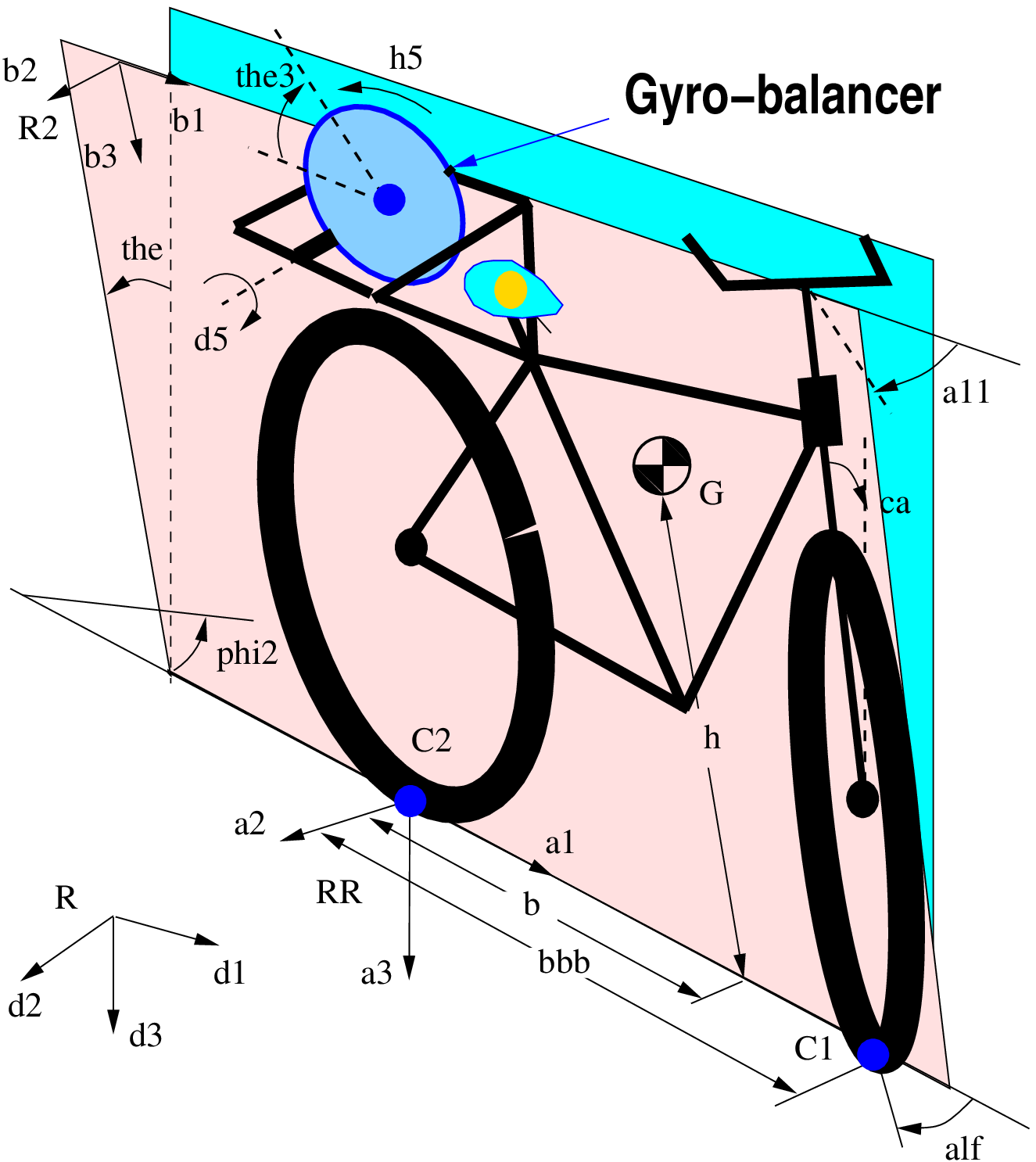}}
	\caption{(a) The autonomous bikebot system with various sensors and actuators developed at Rutgers University. (b) A snapshot of the indoor testing experiment setup. (c) The schematic of the bikebot modeling setup.}
	\label{twosystem2}
\end{figure*}

The learning-based control method is implemented and demonstrated independently on two underactuated balance robotic platforms: a rotary inverted pendulum and a bikebot. Figures~\ref{pendulum} and~\ref{bikebot} show these two robotic systems and we present the experimental results in this section.    

\subsection{Experimental testbeds}

The rotary inverted pendulum shown in Figure~\ref{pendulum} is a commercial robotic platform provided by Quanser Inc. In this system, the actuated joint is the base angle $\theta$ that is driven by a motor. The unactuated joint is the pendulum angle $\alpha$ and its value is defined to be zero when the pendulum is vertically upright. The voltage of the motor, denoted by $V_m$, is the control input to the system. The control goal is to balance the pendulum around upright position, while the rotary base tracks a desired trajectory $\theta_d$.

The motion of the external subsystem is captured by angular position $\theta_1=\theta$ and velocity $\theta_2=\dot{\theta}$, while the motion of the internal subsystem is modeled by position $\alpha_1=\alpha$  and velocity $\alpha_2=\dot{\alpha}$. The control input is $u_d=V_m$. Defining $\bs{\alpha}=[\alpha_1 \; \alpha_2]^T$, the dynamic model is
\begin{equation}
\begin{cases}
\dot{{\theta}}_1 =\theta_2, \; \dot{\theta}_2 = {f}_{\theta}(\theta_2,\bs{\alpha},u_d), \\
\dot{\alpha}_1 ={\alpha}_2,   \; \dot{\alpha}_2 + {\kappa}_{\alpha}(\theta_2,\bs{\alpha},\dot{\alpha}_2)=u_d
\end{cases}
\label{pend_dyn}
\end{equation}
with functions $f_{\theta}$ and $\kappa_{\alpha}$ are given in Appendix~\ref{invertedpendulum}. 

The bikebot shown in Fig.~\ref{bikebot3} is a single-tracked vehicle and is equipped with multiple sensors and actuators to study autonomous driving~\cite{WangICRA2017} and physical human-robot interactions~\cite{chenIJRR2015,zhangBME2013}. Figure~\ref{bikebot4} shows an indoor testing experiment setup that we use in this study. The bikebot position is obtained by a computer vision system with a camera that is mounted on the high ceiling of the lab. The bikebot roll and steering angles are obtained from onboard sensors.  The detailed description of the system hardware setup can be found in~\cite{WangICRA2017}.  

Figure~\ref{bikebot_sketch} illustrates the kinematic relationship and configuration of the bikebot system. The bikebot platform consists of a main body structure (with the rear wheel) and a front wheel that is connected with the frame through the steering joint. The rear wheel contact point is denoted as $C_2$ and its planar coordinate is denoted as $\bs{r}_{C_2}=[X\,\;Y]^T$ in the $XY$ plane of the inertial frame $\mathcal{N}$. The yaw (heading) and roll (with the vertical plane) angles of the bikebot platform are denoted as $\psi$ and $\varphi$, respectively. The steering angle is denoted as $\phi$ and the rear wheel velocity as ${v}_c$. Due to the nonholonomic constraint of point $C_2$, its velocity is obtained as $\boldsymbol{v}_{C_{2}}=[\dot{X}\,\; \dot{Y}]^T=[v_c\cos\psi  \;\, v_c\sin\psi]^T$. The external subsystem motion of the bikebot is captured by position $\bs{\theta}_1=[X \,\; Y]^T$ and velocity $\bs{\theta}_2=[\dot{X} \,\; \dot{Y}]^T$ and the internal subsystem motion is by position  ${\alpha}_1=\varphi$ and velocity ${\alpha}_2=\dot{\varphi}$. The control inputs are $\bs{u}=[u_d \,\; u_f]^T$ with $u_d=\phi$ and $u_f=\dot{v}_c$. The bikebot dynamic model is written as~\cite{WangICRA2017} 
\begin{equation}
\begin{cases}
\dot{\bs{\theta}}_1= \bs{\theta}_2, \; \dot{\bs{\theta}}_2=\bs{f}_\theta(\bs{\theta},\bs{\alpha},\bs{u}),   \\
\dot{\alpha}_1=\alpha_2,\; \dot{\alpha}_2+ \kappa_{\varphi}(\bs{\theta}, \bs{\alpha},\dot{\alpha}_2,u_f)=u_d 
\end{cases}
\label{bike}
\end{equation}
with $\bs{f}_\theta$ and $\kappa_{\varphi}$ are given in Appendix~\ref{invertedpendulum}. The desired trajectory for the external subsystem is denoted as $\bs{\theta}_{d}=[X_d \,\;Y_d]^T$.

\begin{figure*}[htb!]
	\hspace{-2mm}
	\subfigure[]{
		\label{random_excite_training}
		\psfrag{T}[][]{\scriptsize Time (s)}
		\psfrag{A}[][]{\scriptsize  $\theta/\alpha$ (rad)}
		\includegraphics[width=1.9in]{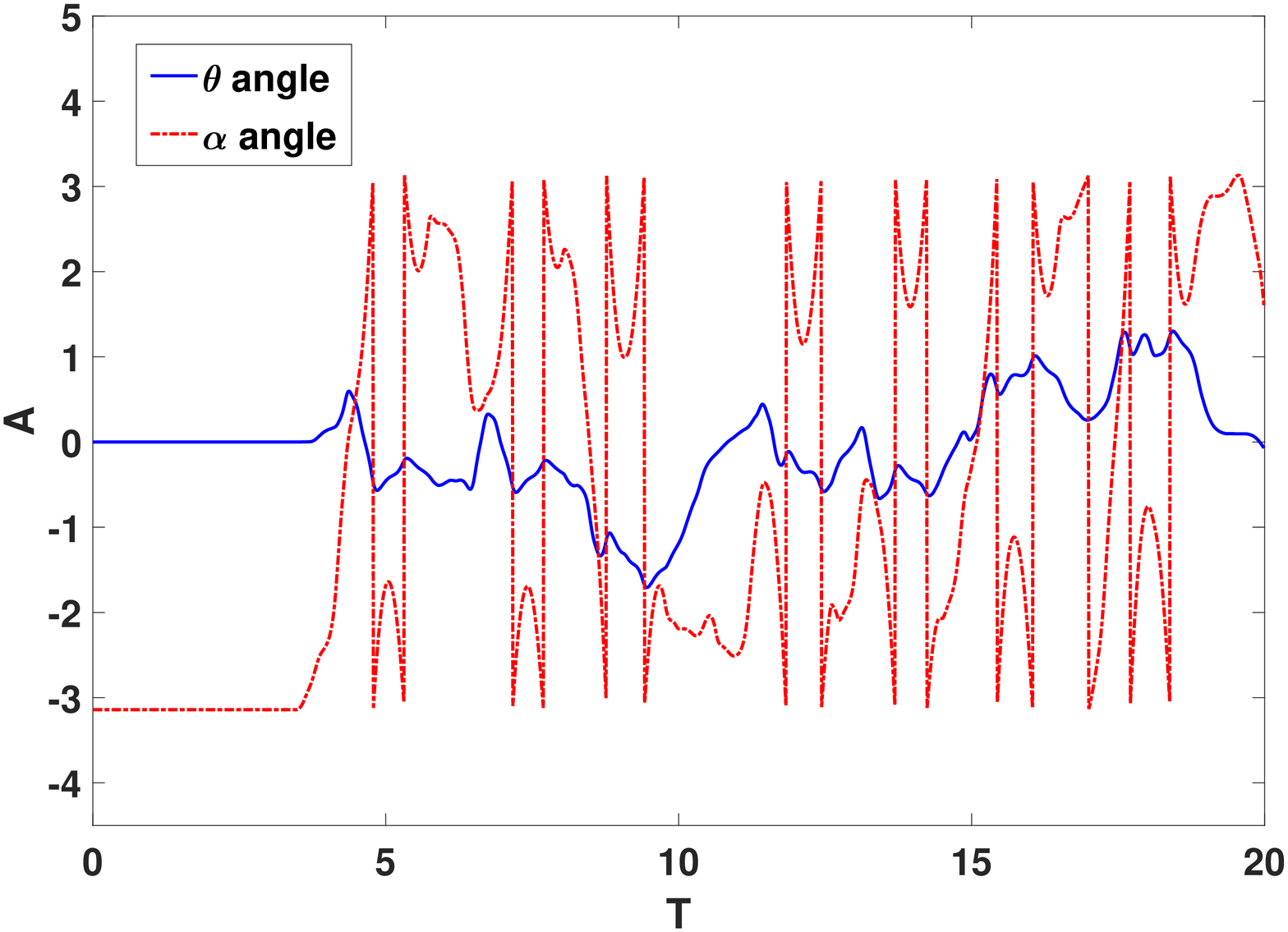}}
	\hspace{-1mm}
	\subfigure[]{
		\label{training_position}
		\psfrag{X}[][]{\small  $X$ (m)}
		\psfrag{Y}[][]{\small  $Y$ (m)}
		\includegraphics[width=2.38in]{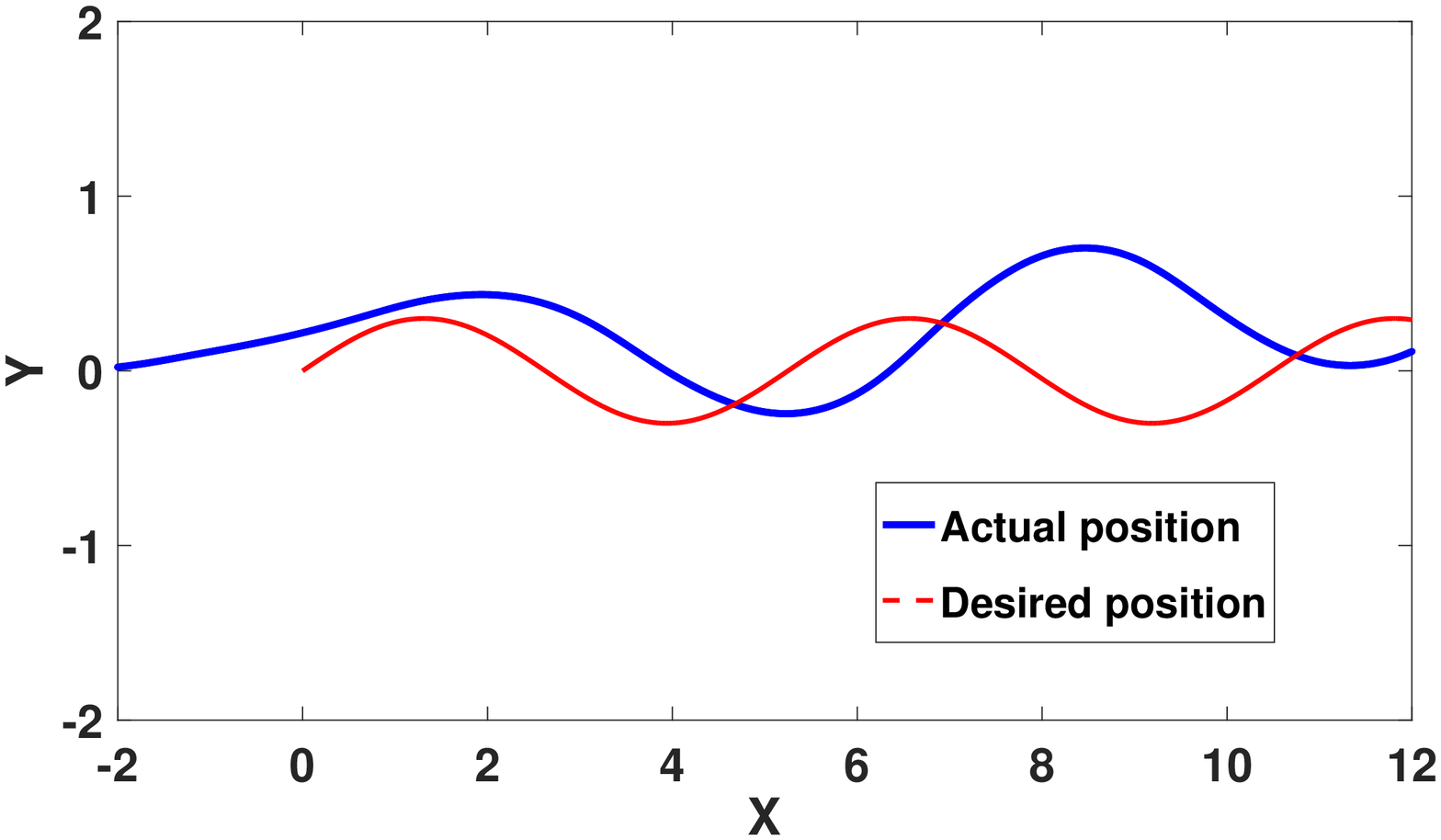}}
	\hspace{-2mm}
	\subfigure[]{
		\label{training_roll}
		\psfrag{T}[][]{\small  Time (s)}
		\psfrag{An}[][]{\small  $\varphi/\phi$ (deg)}
		\psfrag{ Roll angle}[][]{\tiny \hspace{2mm} Roll angle $\varphi$}
		\psfrag{ Steer angle}[][]{\tiny \hspace{2mm} Steer angle $\phi$}
		\includegraphics[width=2.33in]{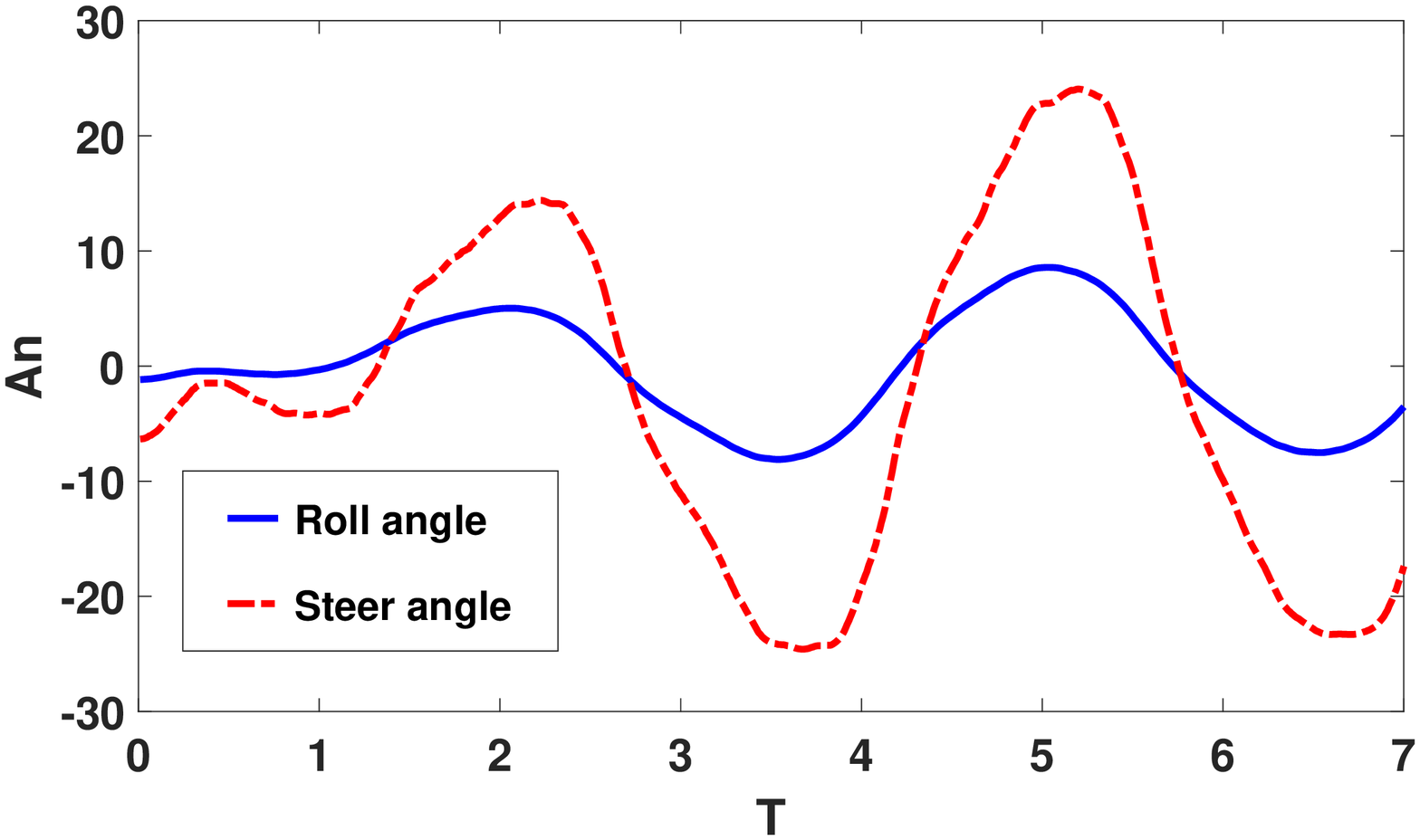}}
	\caption{Example profiles of the collected training data for (a) rotary pendulum experiments (under open-loop control) and for the bikebot experiments (under baseline EIC controller): (b) Bikebot position and (c) bikebot roll and steer angle.}
	\label{training_data}
	\vspace{-0mm}
\end{figure*}

\subsection{Experimental results}

\subsubsection{Rotary inverted pendulum experiments}

To obtain the learned model of the rotary inverted pendulum, we perturb the system and collect the motion data. An open-loop input is implemented as 
\begin{equation}
V_m= \begin{cases} a_1\sin(\omega_1 t)+a_2\sin(\omega_2 t) \; , |\alpha|\leq \frac{\pi}{3}, \\  0 \;, |\alpha|> \frac{\pi}{3},
\end{cases}
\label{open_loop}
\end{equation}
where $a_1$ and $a_2$ are chosen to satisfy the input bound $|V_m|\leq 5$ V, $\omega_1$ and $\omega_2$ are designed to excite the system by both low and high frequencies. In experiment, we choose $a_1=3$, $a_2=1.5$, $\omega_1=8$ rad/s and $\omega_2=40$ rad/s. Under this input, we swing up the pendulum manually by giving an initial velocity when angle $|\alpha|\geq \frac{\pi}{2}$. The above open-loop input $V_m$ cannot stabilize the pendulum to stay around the upright position. For each swing, the pendulum angle $\alpha$ might stay in the range of $|\alpha|\leq \frac{\pi}{3}$ for less than one second and then fall. We choose the input in~(\ref{open_loop}) as an example to collect training data and indeed, some other forms of input voltage are also used.  

\begin{figure*}[htb!]
	\centering
	\subfigure[]{
		\label{time_track:a}
		\psfrag{T}[][]{\small Time (s)}
		\psfrag{A}[][]{\small  $\theta$ (rad)}
		\includegraphics[width=2.5in]{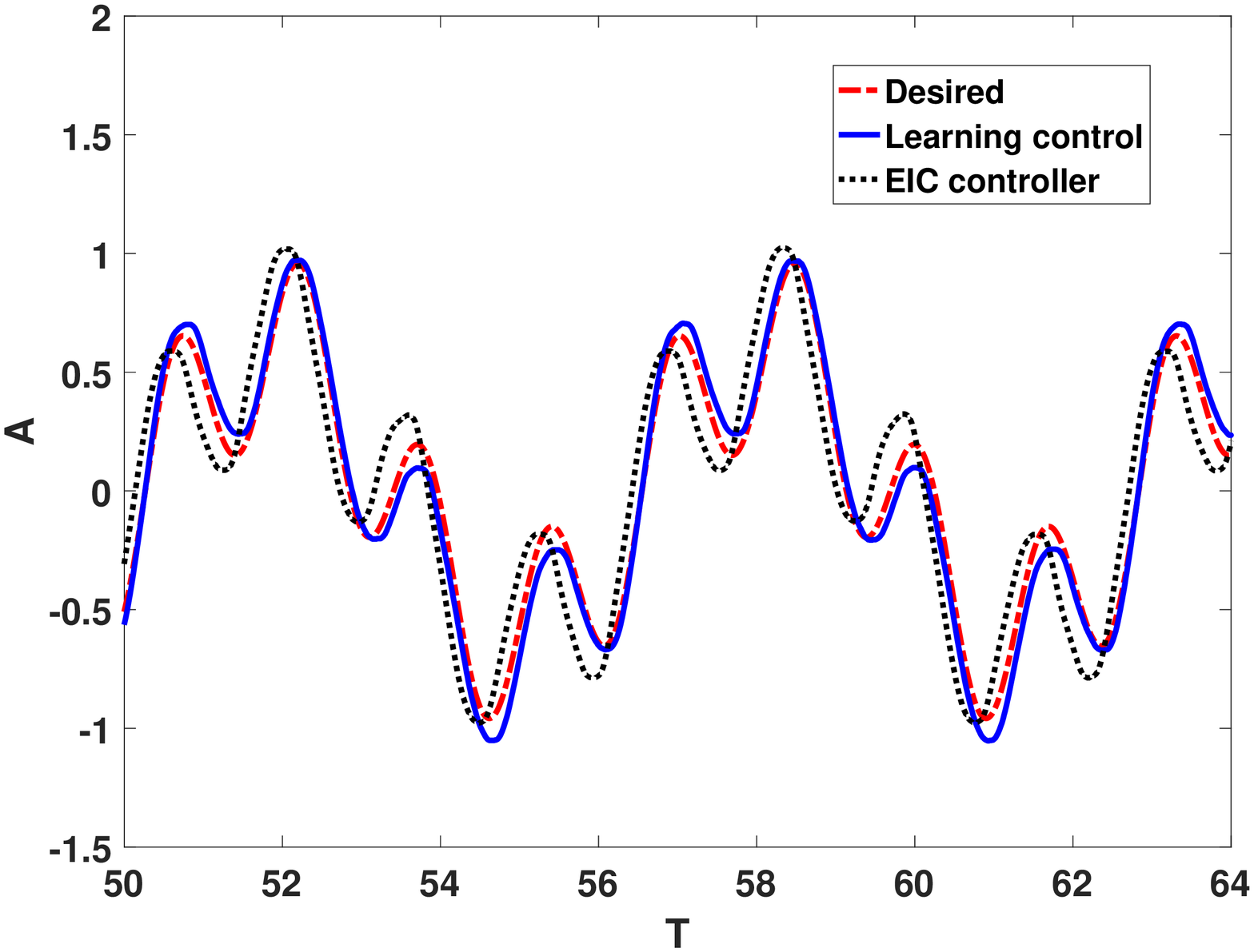}}
	\hspace{2mm}
	\subfigure[]{
		\label{time_track:b}
		\psfrag{T}[][]{\small  Time (s)}
		\psfrag{B}[][]{\small  $\alpha$ (rad)}
		\includegraphics[width=2.5in]{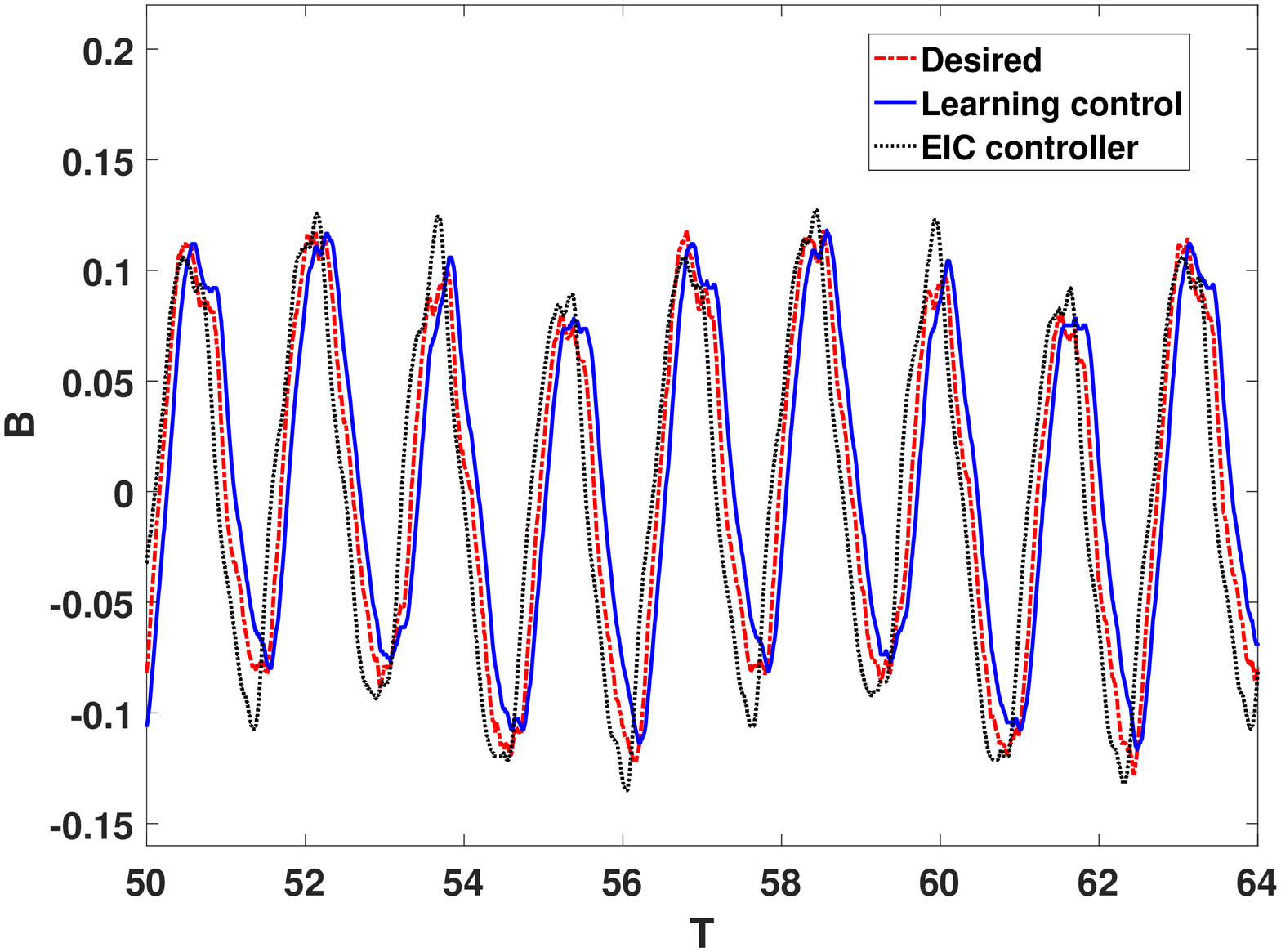}}
	\subfigure[]{
		\label{time_track:bb}
		\psfrag{T}[][]{\small  Time (s)}
		\psfrag{AE}[][]{\small  $e_\theta$ (rad)}
		\includegraphics[width=2.5in]{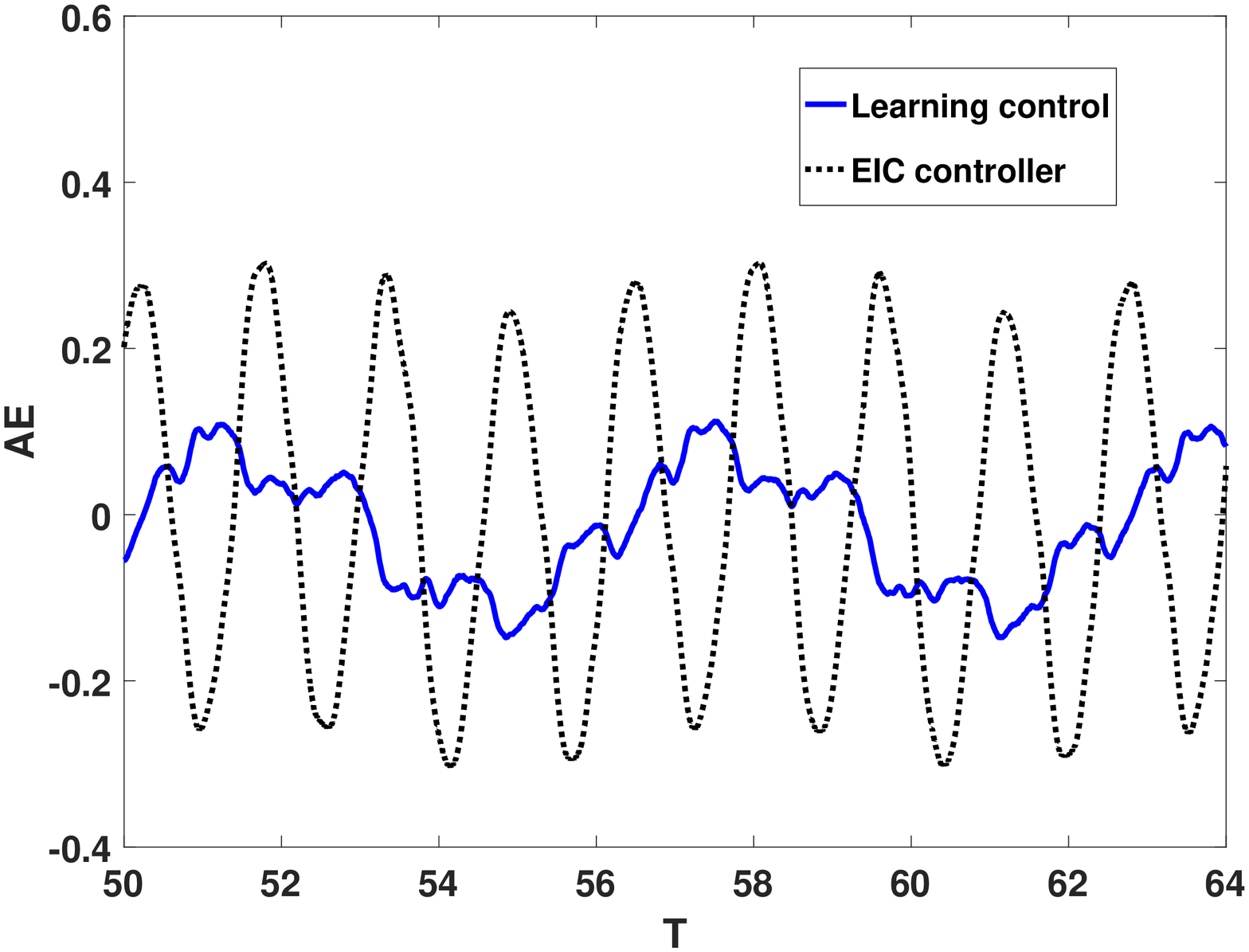}}
	\hspace{2mm}
	\subfigure[]{
		\label{time_track:c}
		\psfrag{T}[][]{\small  Time (s)}
		\psfrag{BE}[][]{\small  $e_\alpha$ (rad)}
		\includegraphics[width=2.5in]{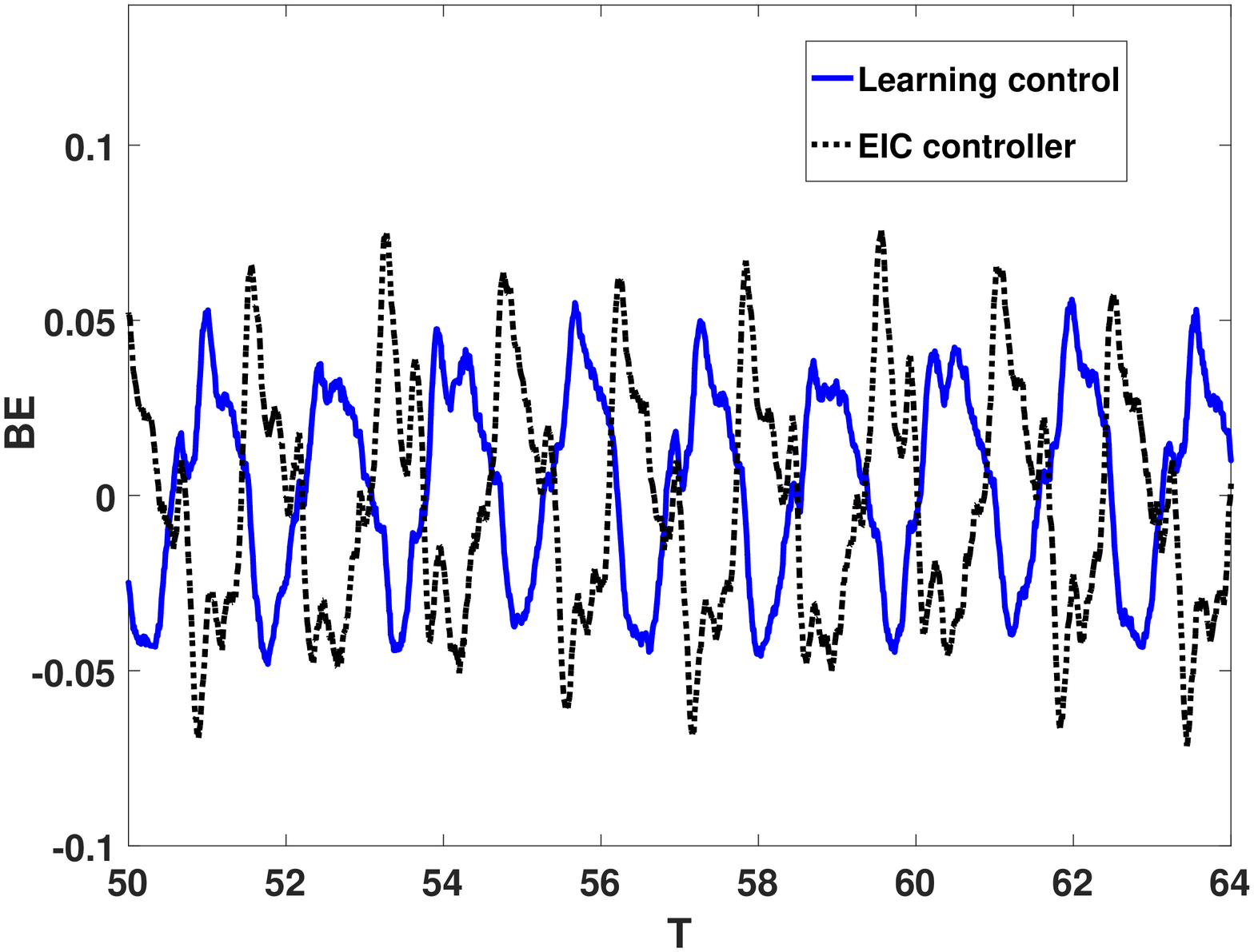}}
	\caption{Tracking errors for one experimental run under the learning-based and the EIC-based controllers for the rotary inverted pendulum. (a) External angle $\theta$ tracking profiles. (b) Internal angle $\alpha$ tracking profiles. (c) External angle tracking errors $e_\theta$. (d) Internal angle tracking errors $e_\alpha$.}
	\label{time_track}
	\vspace{-2mm}
\end{figure*}

Control input $V_m$ and motion data are recorded when $ |\alpha|\leq \frac{\pi}{3}$. The joint angles $\theta$ and $\alpha$ are measured with encoders. Their velocities and accelerations are obtained by numerically differentiation of the filtered joint angles. The open-loop controller and data collection are implemented at a frequency of 100 Hz. Figure~\ref{random_excite_training} shows an example of collected $\theta$ and $\alpha$ angles under the open-loop input~(\ref{open_loop}). It is clear that the pendulum does not achieve balance under~(\ref{open_loop}). Multiple trials of manual swing are applied to the pendulum to collect enough data for $|\alpha| \leq \frac{\pi}{3}$. The controller is implemented through Matlab Real-Time Workshop. The MPC is implemented with a period of $0.02$ s, that is, $\Delta t=0.02$ s, and preview horizon is $H=27$. In implementation, the weight matrices in~(\ref{e_obj}) are chosen as  $\bs{Q}_1=\bs{Q}_3=\diag\{1000,100\}$, $\bs{Q}_2=\diag\{100,100\}$, $\bs{R}=10 \bs{I}_2$ and $\nu=1$.

\begin{figure*}[htb!]
	\centering
	\subfigure[]{
		\label{time_track:d}
		\includegraphics[width=2.8in]{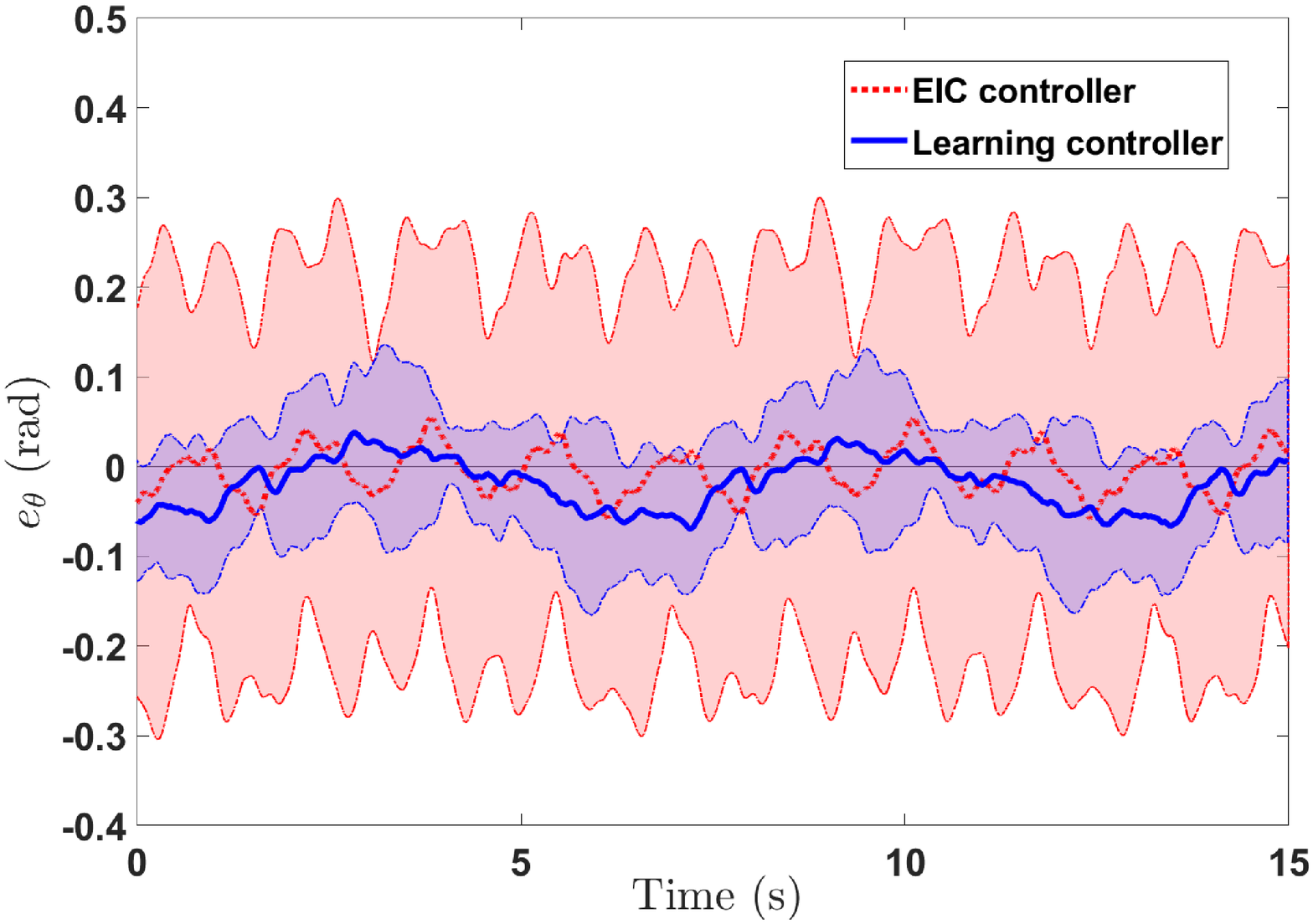}}
	\hspace{-2mm}
	\subfigure[]{
		\label{time_track:e}
		\includegraphics[width=2.8in]{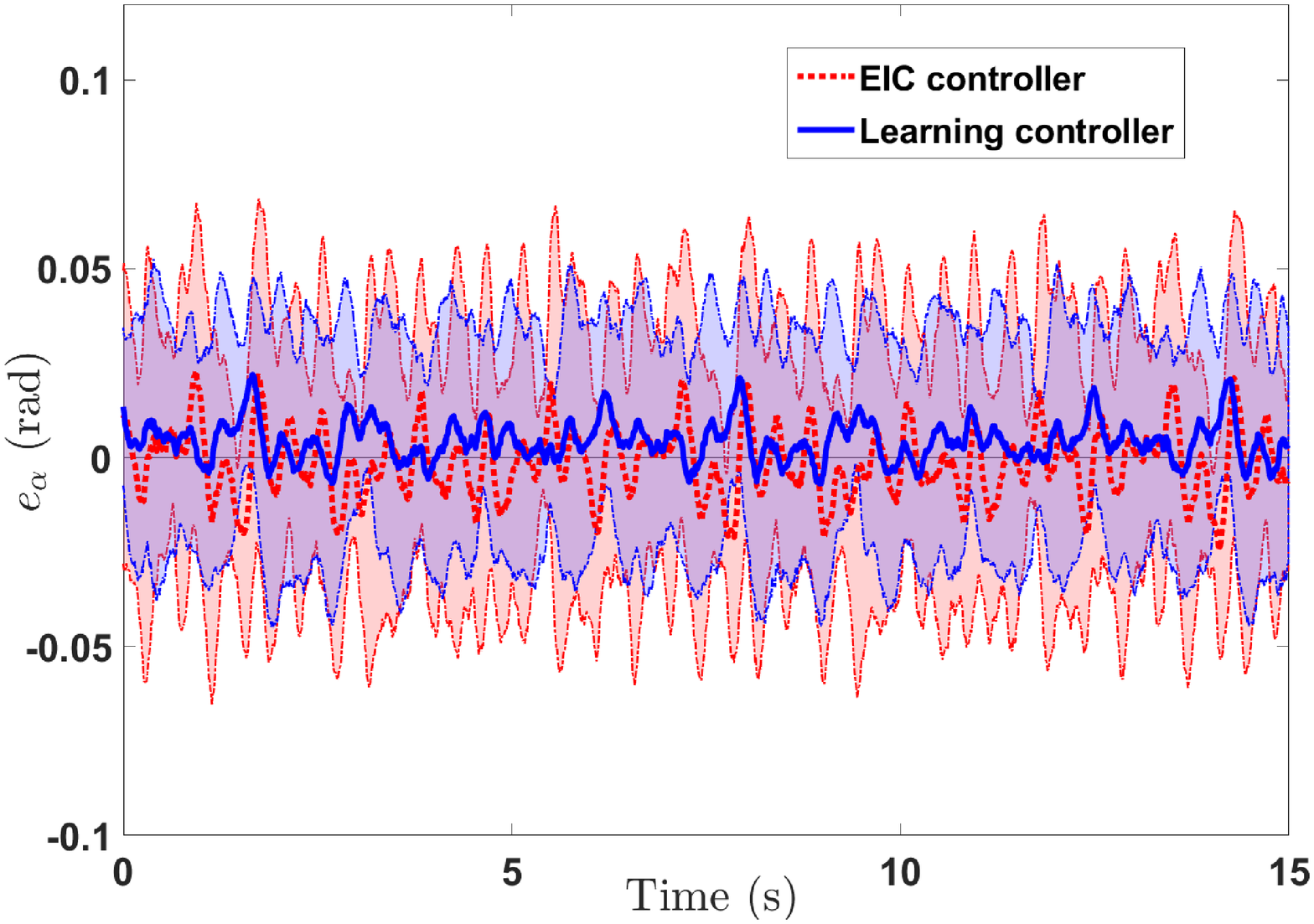}}
	\caption{Tracking errors $e_\theta$ and $e_\alpha$ by multiple experimental runs under the learning-based and EIC-based controllers for the rotary inverted pendulum. Mean error and standard deviation profiles for (a) $e_\theta$ and for (b) $e_\alpha$.}
	\label{time_track2}
	\vspace{-2mm}
\end{figure*}

\begin{figure*}[htb!]
	\hspace{-4mm}
	\subfigure[]{
		\label{straight_track0}
		\psfrag{X}[][]{\small  $X$ (m)}
		\psfrag{Y}[][]{\small  $Y$ (m)}
		\includegraphics[width=2.36in]{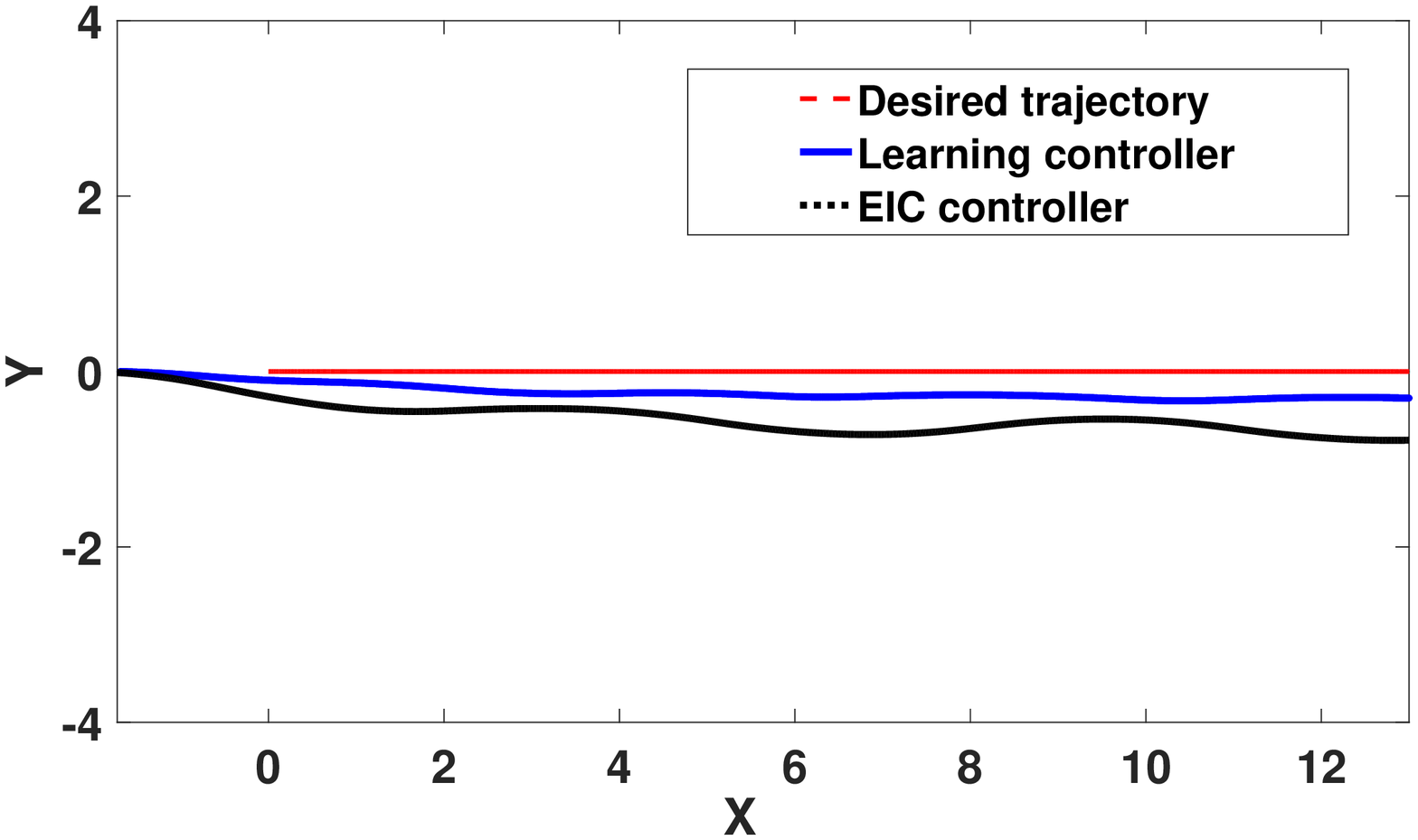}}
	\hspace{-5mm}
	\subfigure[]{
		\label{sine_track0}
		\psfrag{X}[][]{\small  $X$ (m)}
		\psfrag{Y}[][]{\small  $Y$ (m)}
		\includegraphics[width=2.36in]{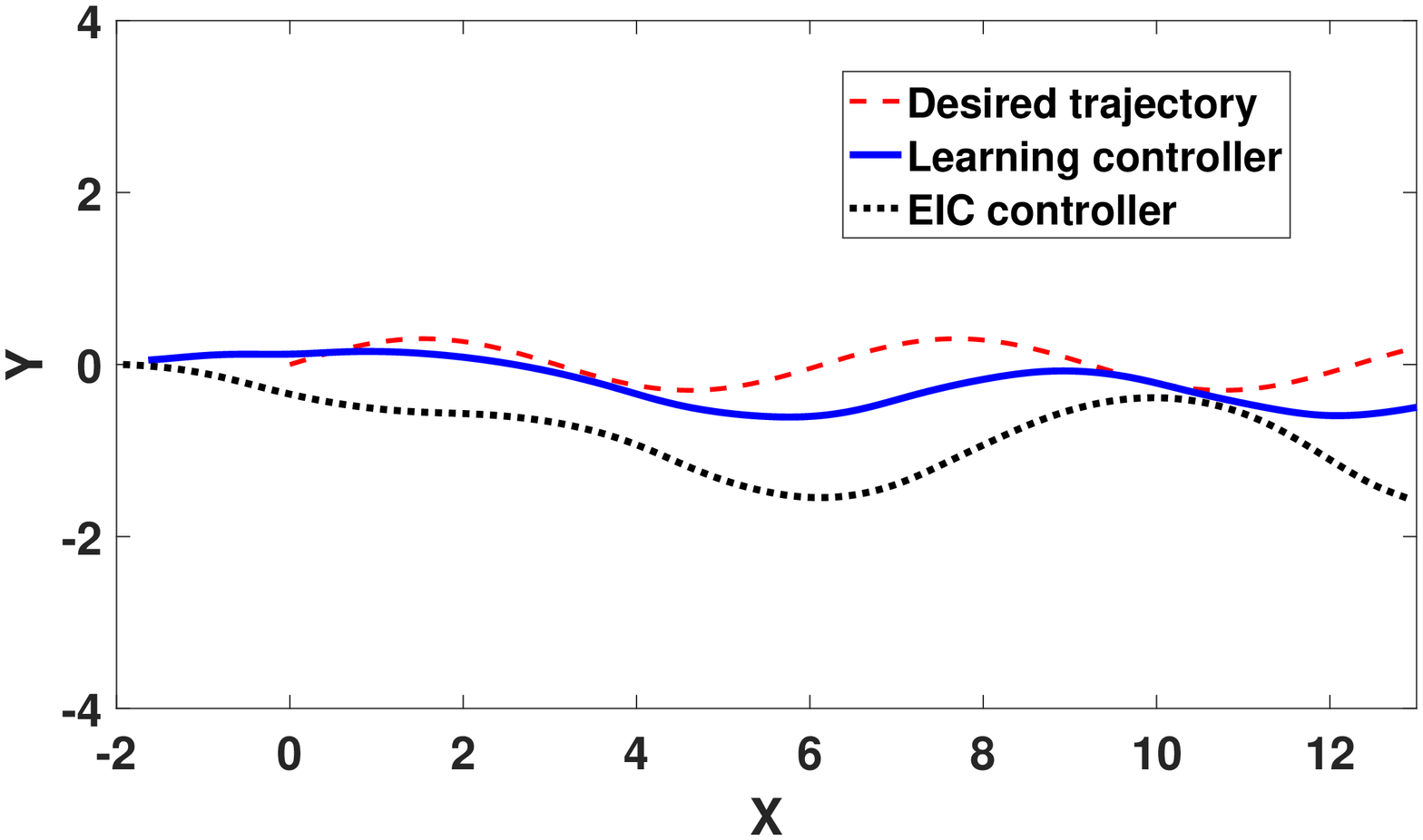}}
	\hspace{-5mm}
	\subfigure[]{
		\label{circle_track0}
		\psfrag{X}[][]{\small  $s$ (m)}
		\psfrag{Y}[][]{\small  $Y$ (m)}
		\includegraphics[width=1.8in]{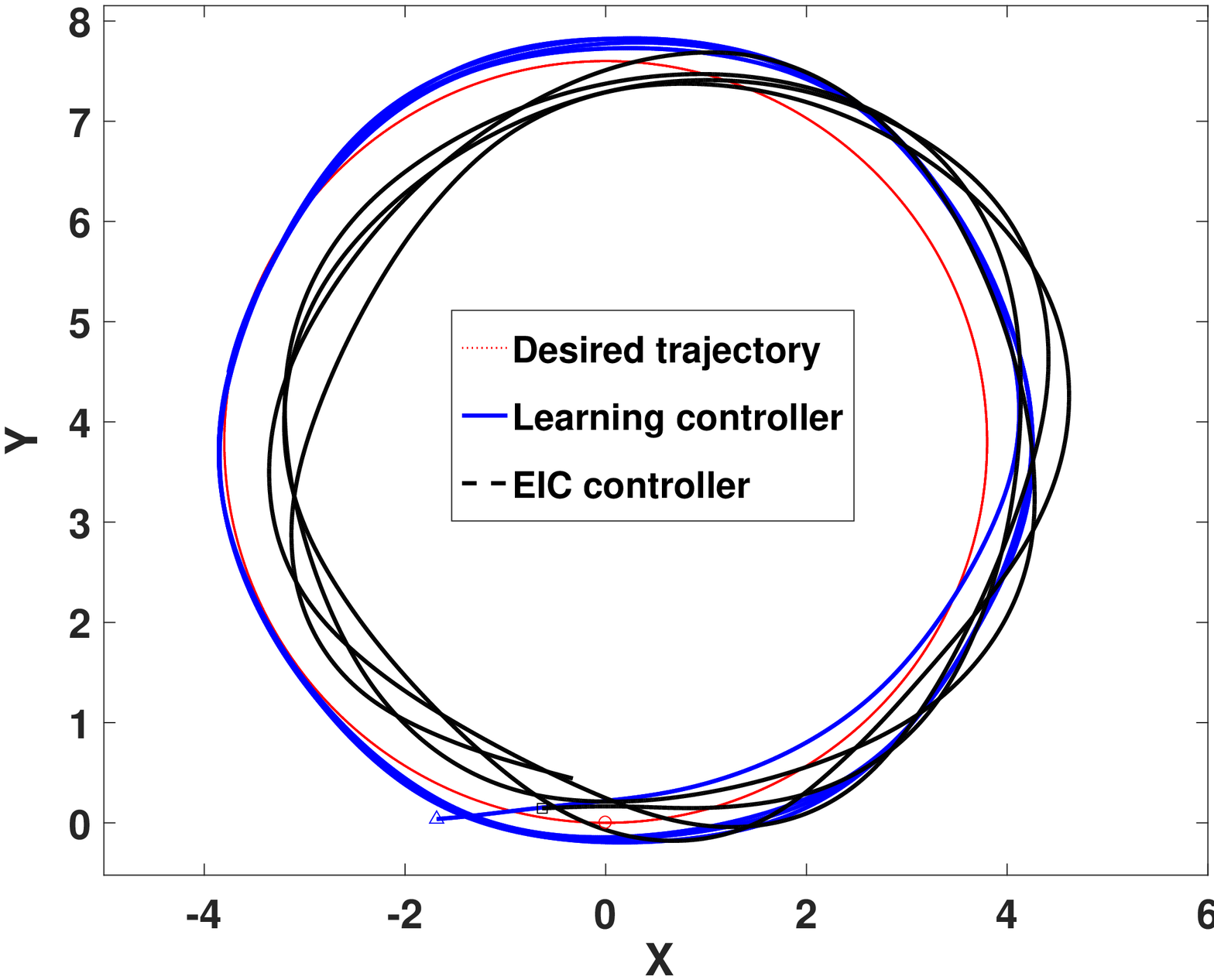}}
	\hspace{-4mm}
	\subfigure[]{
		\label{straight_roll0}
		\psfrag{T}[][]{\small  Time (s)}
		\psfrag{BR}[][]{\small  $\varphi$ (deg)}
		\includegraphics[width=2.15in]{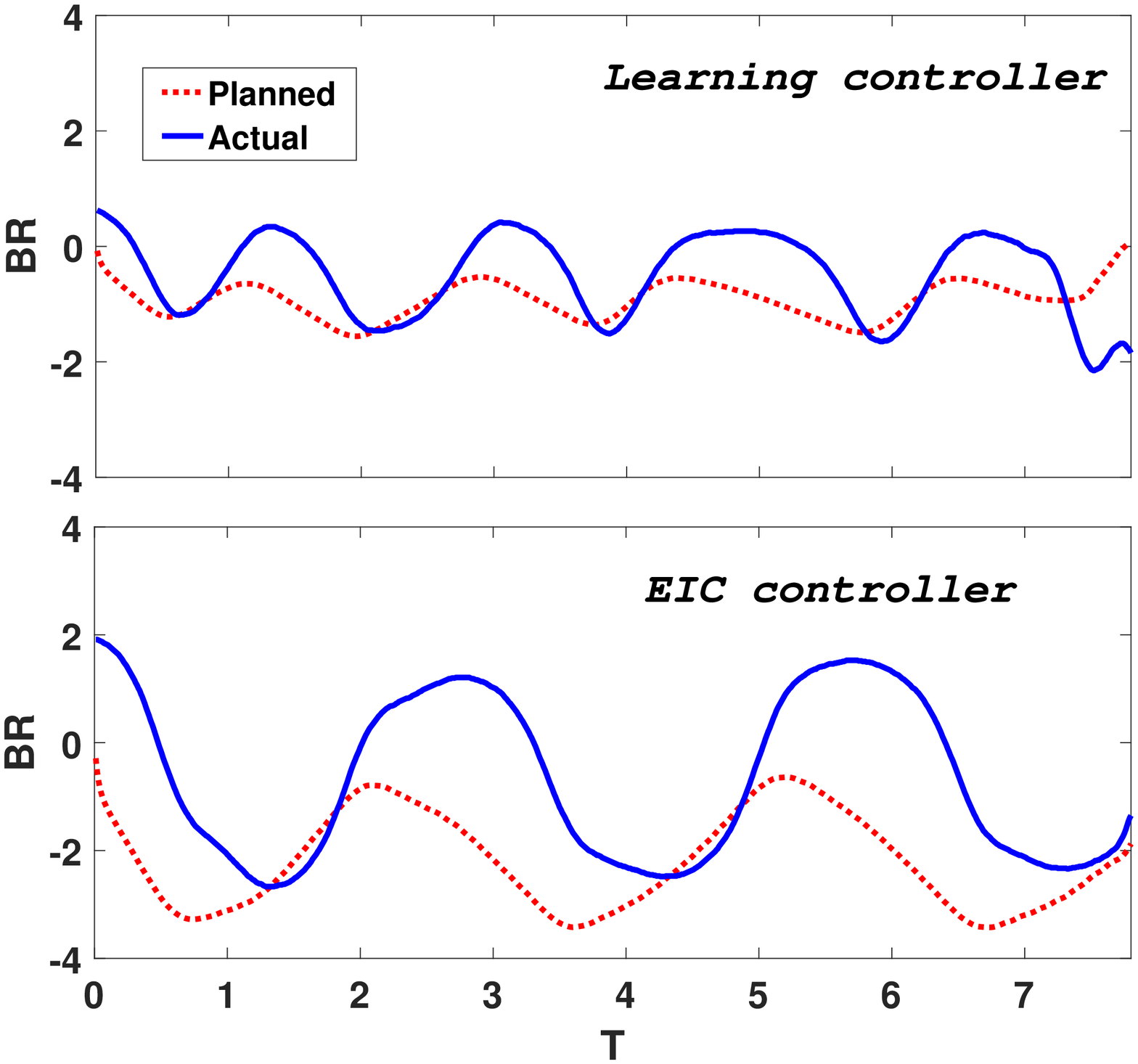}}
	\hspace{-1mm}
	\subfigure[]{
		\label{sine_roll0}
		\psfrag{T}[][]{\small  Time (s)}
		\psfrag{BR}[][]{\small  $\varphi$ (deg)}
		\includegraphics[width=2.23in]{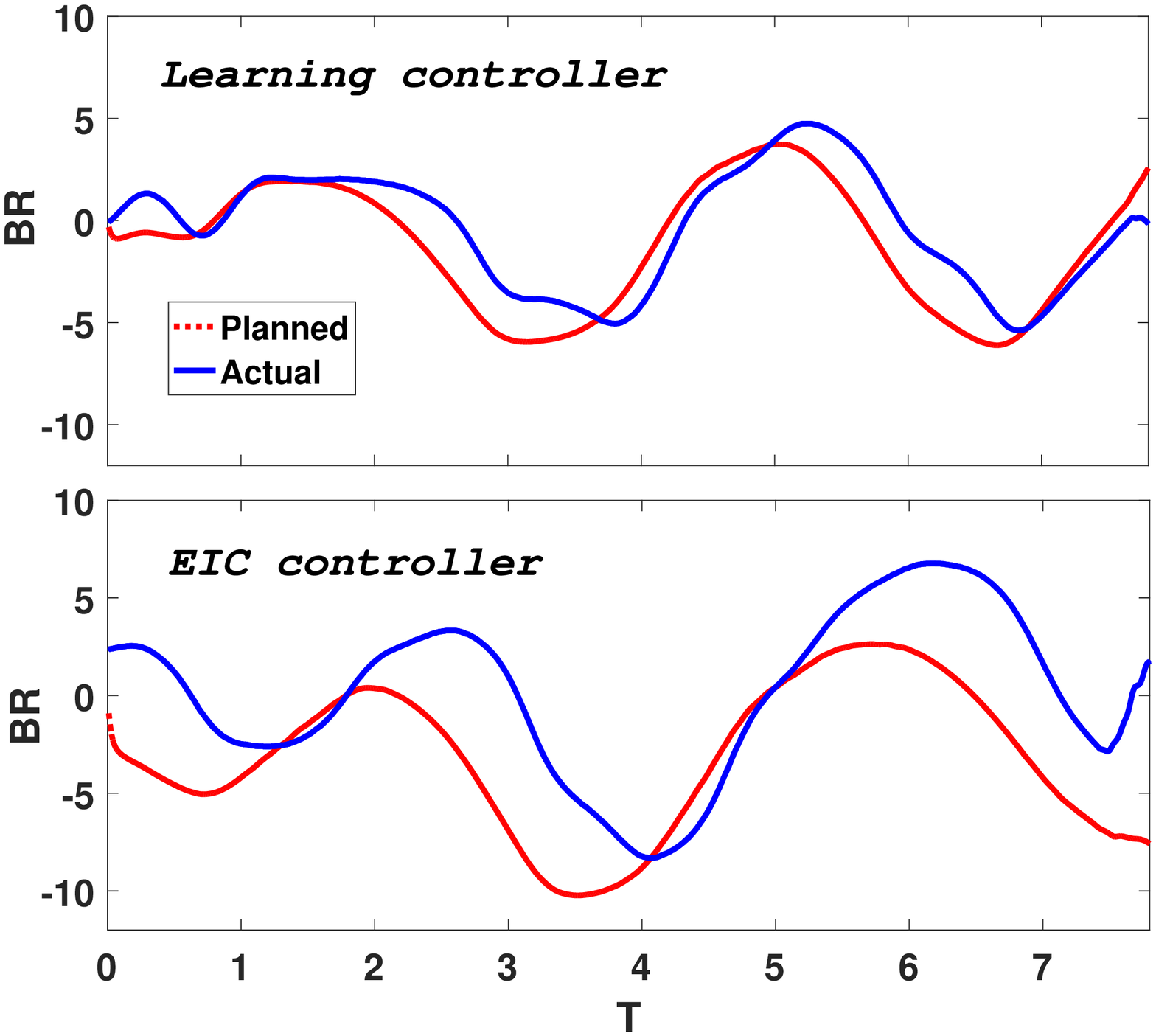}}
	\hspace{-1mm}
	\subfigure[]{
		\label{circle_roll0}
		\psfrag{T}[][]{\small  Time (s)}
		\psfrag{BR}[][]{\small  $\varphi$ (deg)}
		\includegraphics[width=2.23in]{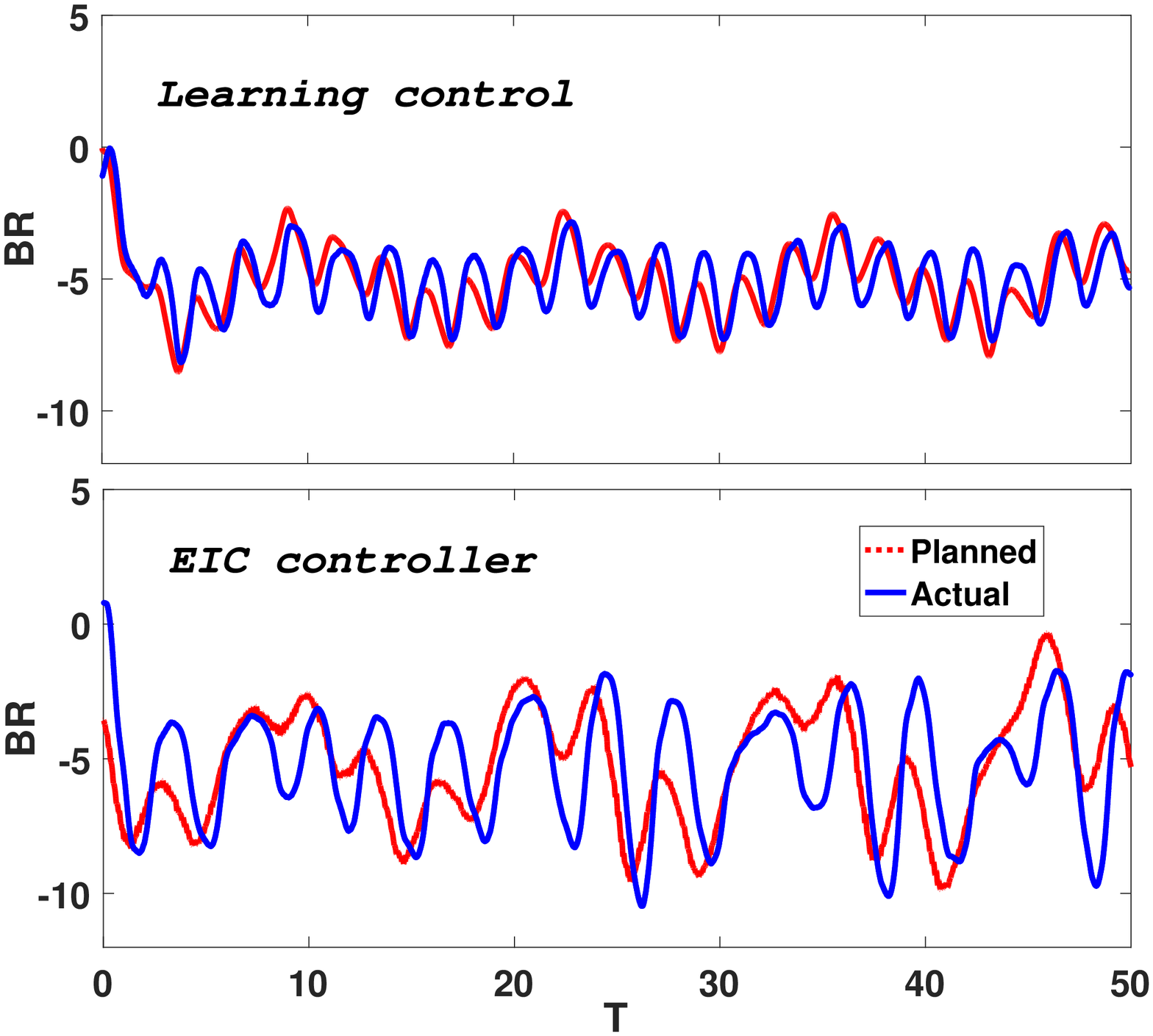}}
	\caption{Performance comparison of the bikebot tracking under the learning-based control and the EIC-based control designs for one experimental run. (a)-(c) for $X$-$Y$ position tracking profiles and (d)-(f) for roll angle profiles for straight-line, sinusoidal and circular trajectories, respectively.}
	\label{traj0}
\end{figure*}

\begin{figure*}[htb!]
\hspace{-4mm}
	\subfigure[]{
	\label{straight_track}
		\psfrag{X}[][]{\small  $X$ (m)}
		\psfrag{Y}[][]{\small  $|e_y|$ (m)}
	\includegraphics[width=2.2in]{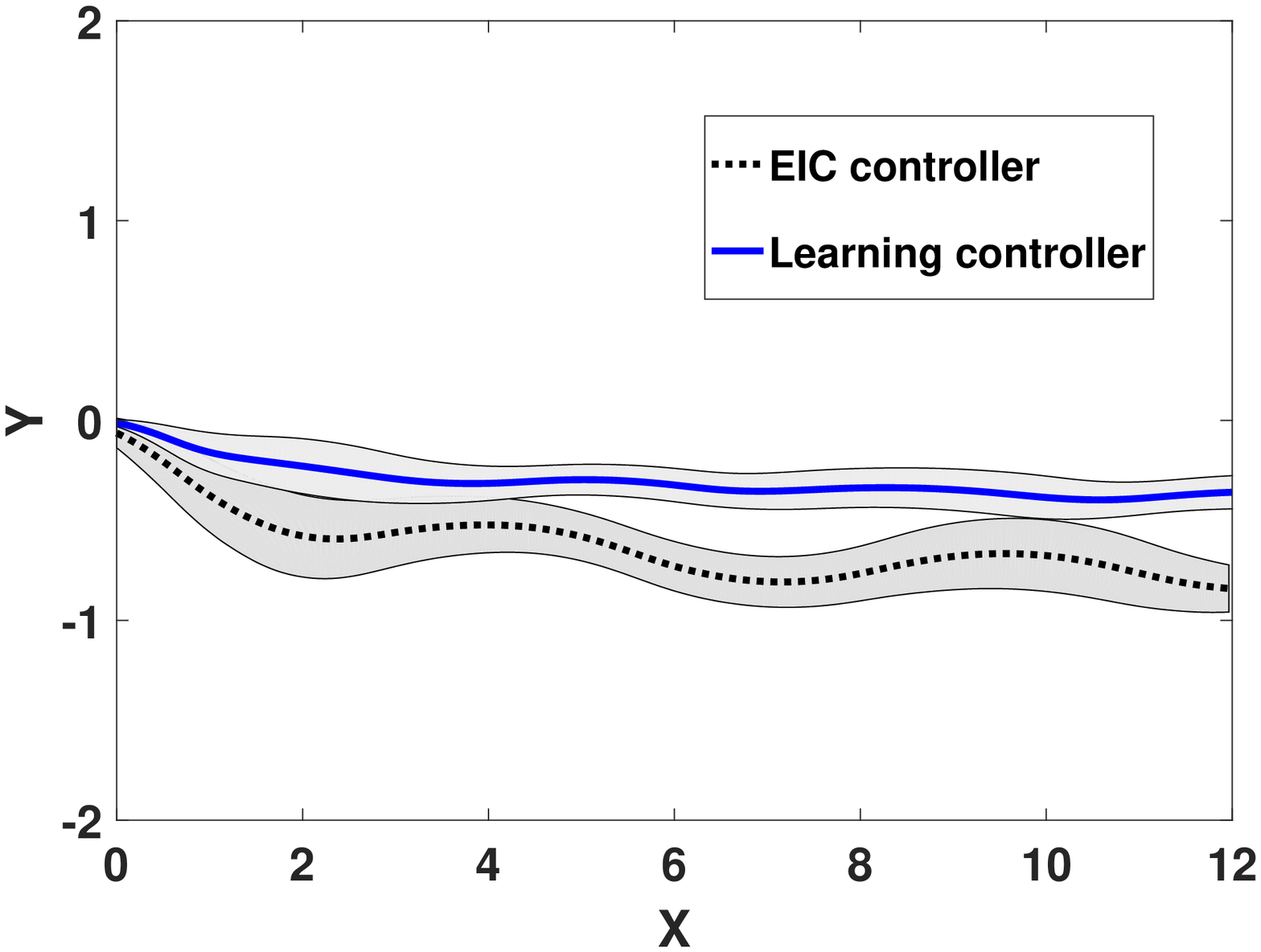}}
\hspace{-1mm}
	\subfigure[]{
	\label{sine_track}
		\psfrag{X}[][]{\small  $X$ (m)}
		\psfrag{Y}[][]{\small  $|e_y|$ (m)}
	\includegraphics[width=2.2in]{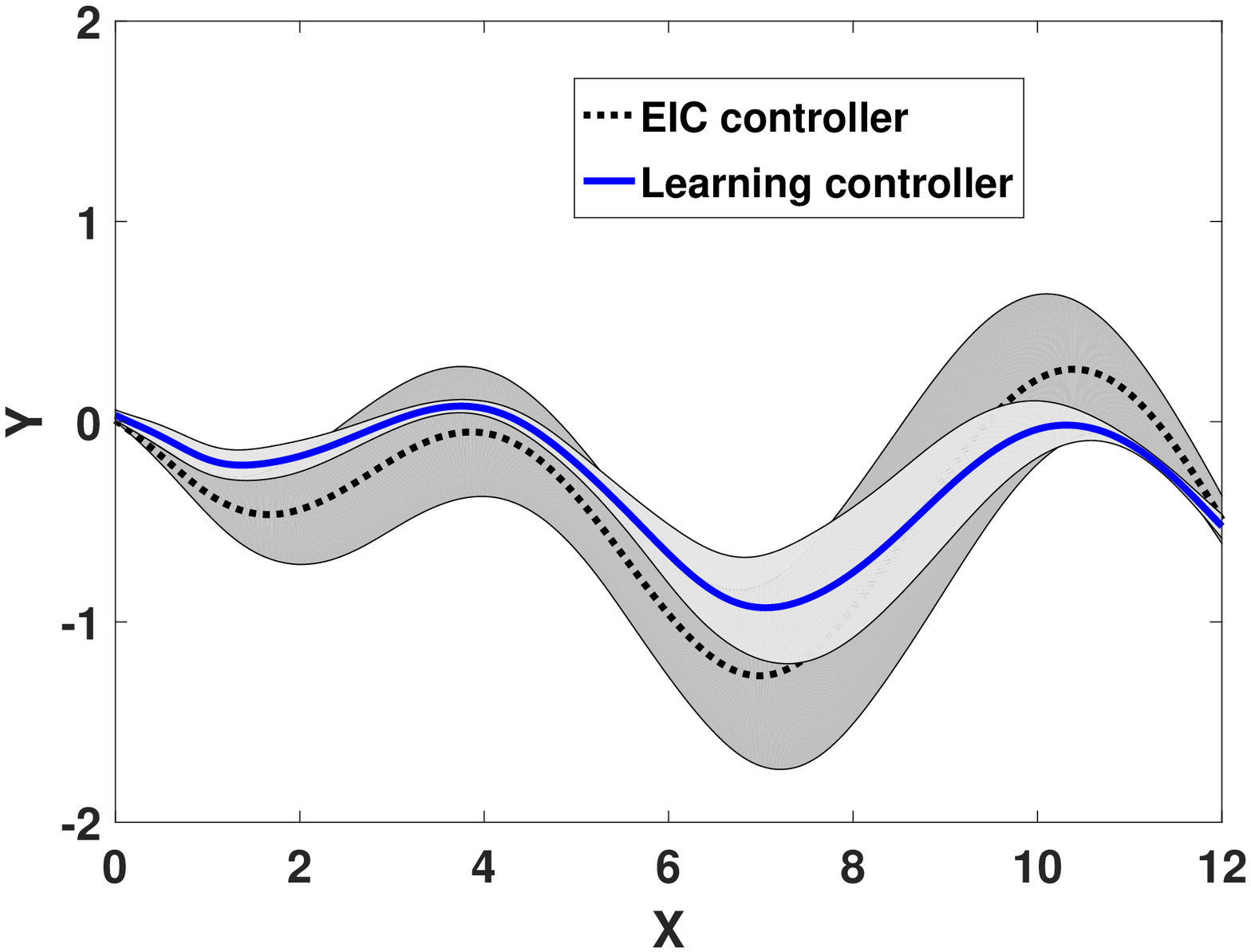}}
\hspace{-1mm}
	\subfigure[]{
	\label{circle_track}
		\psfrag{X}[][]{\small  $s$ (m)}
\psfrag{Y}[][]{\small  $|e|$ (m)}
	\includegraphics[width=2.2in]{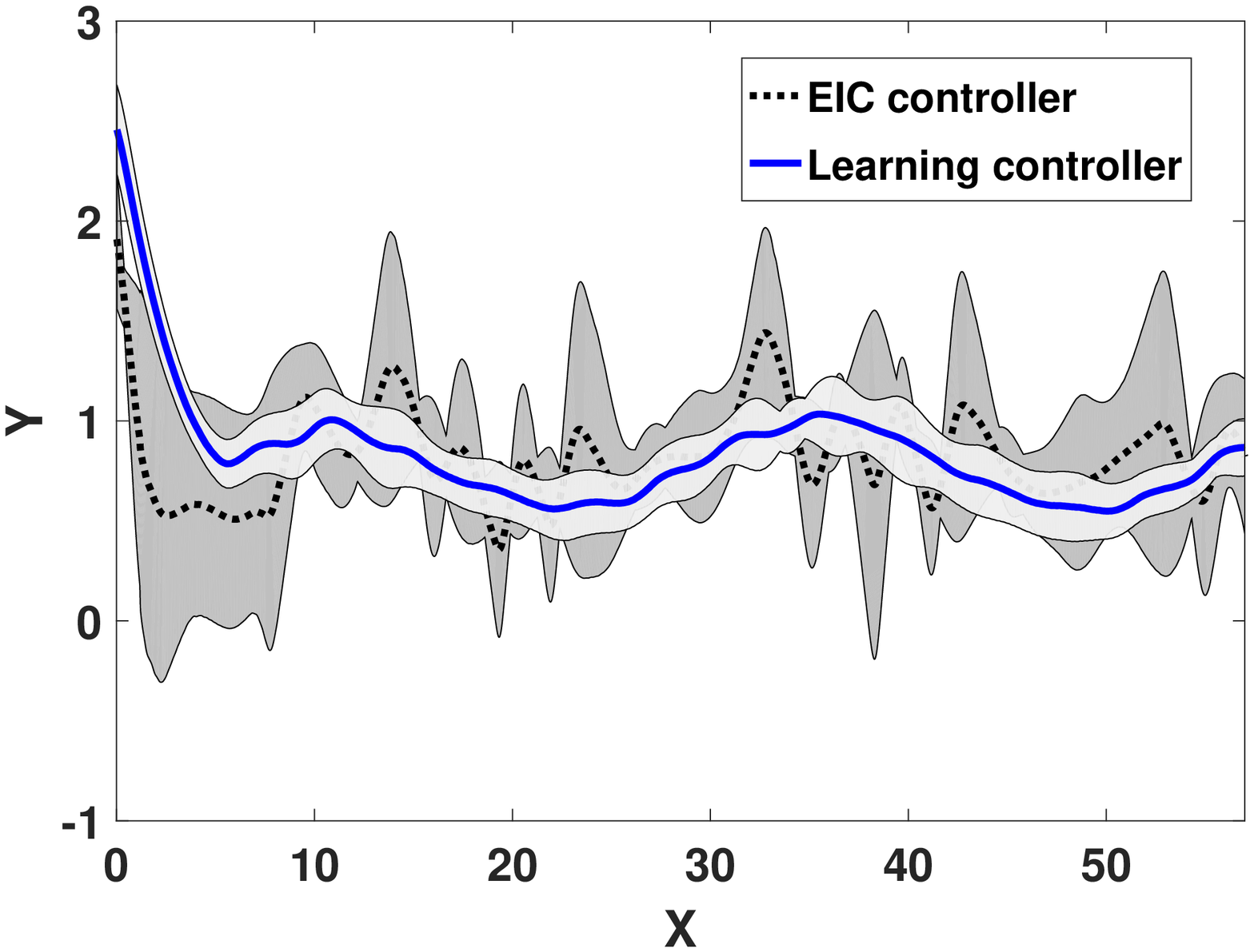}}
\hspace{-4mm}
	\subfigure[]{
	\label{straight_roll}
\psfrag{T}[][]{\small  Time (s)}
\psfrag{RE}[][]{\small  $e_\varphi$ (deg)}
	\includegraphics[width=2.21in]{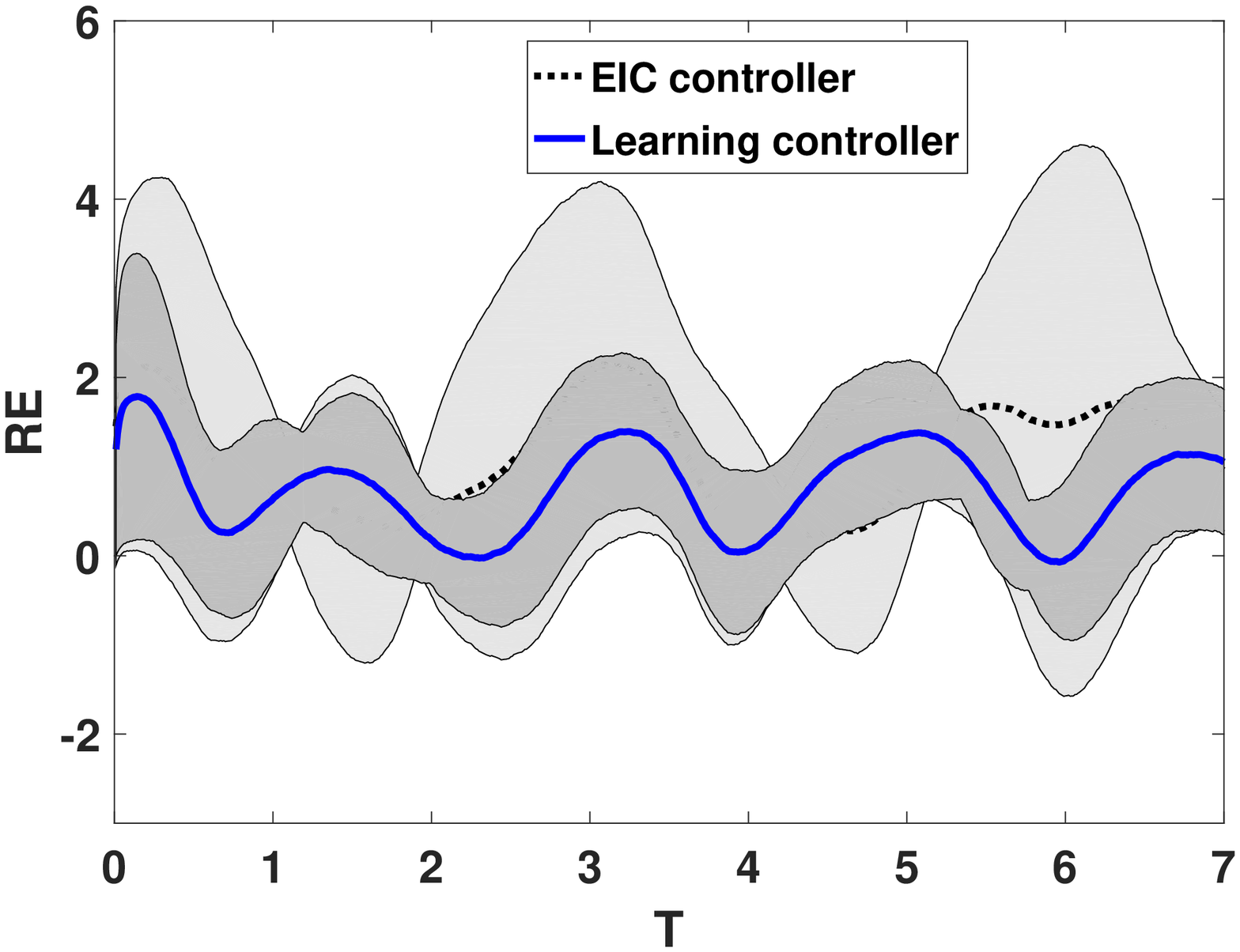}}
\hspace{-1mm}
	\subfigure[]{
	\label{sine_roll}
\psfrag{T}[][]{\small  Time (s)}
\psfrag{RE}[][]{\small  $e_\varphi$ (deg)}
	\includegraphics[width=2.21in]{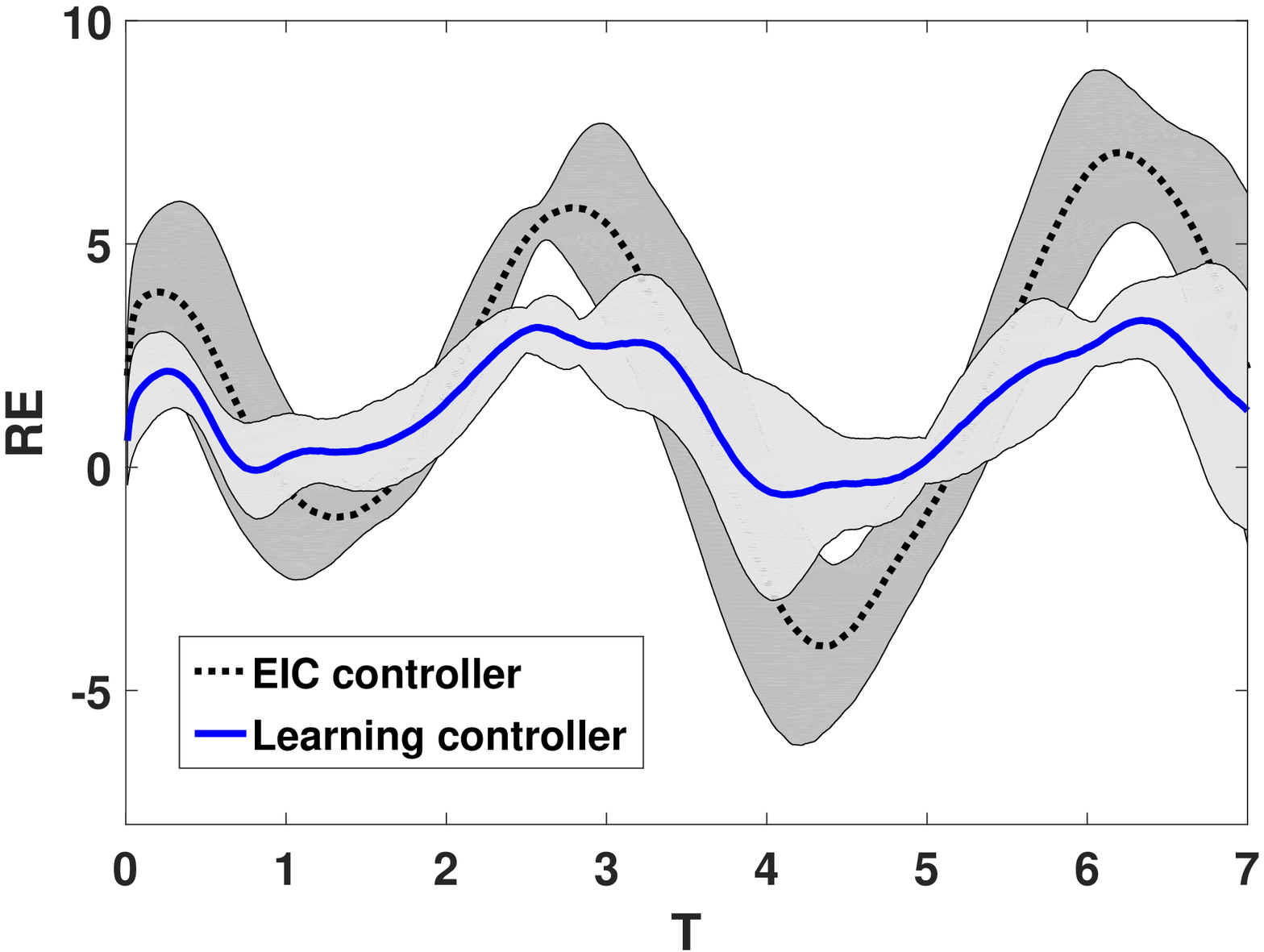}}
\hspace{-1mm}
	\subfigure[]{
	\label{circle_roll}
\psfrag{T}[][]{\small  Time (s)}
\psfrag{RE}[][]{\small  $e_\varphi$ (deg)}
	\includegraphics[width=2.21in]{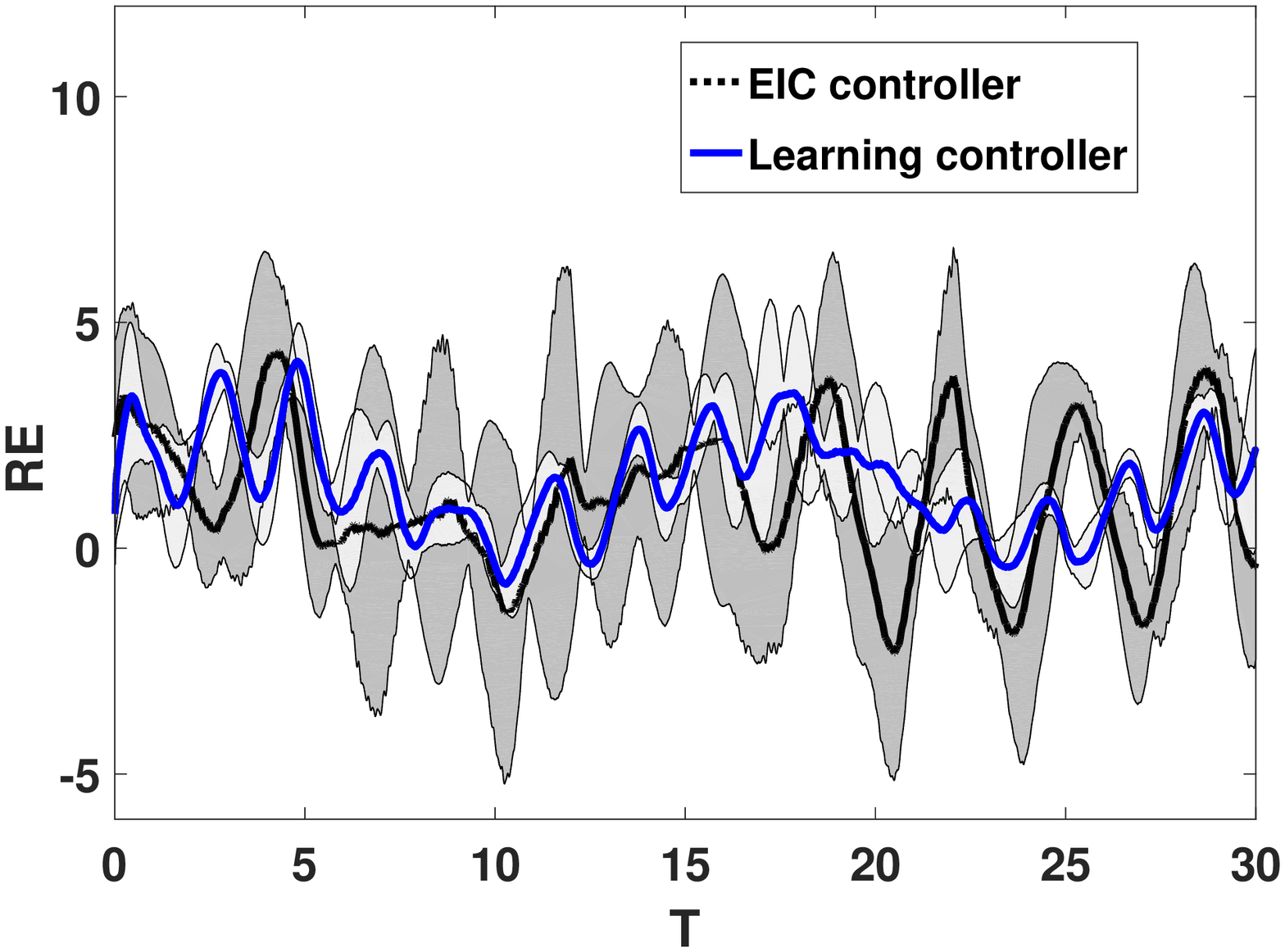}}
	\caption{The bikebot position and roll angle tracking error profiles with multiple experimental runs under the learning-based and the EIC-based controllers. (a)-(c) for $Y$ position tracking error profiles and (d)-(f) for roll angle error profiles for straight-line, sinusoidal and circular trajectories, respectively.}
	\label{traj1}
\end{figure*}

A set of 800 points are collected and used as the training data. In testing and validation experiments, the desired external trajectory was designed as $\theta_d=0.6\sin(t)+0.4\sin(4t)$ rad. We chose this smooth curve as a representative profile to demonstrate the performance. Figure~\ref{time_track:a} shows the tracking results of the external subsystem base angle $\theta$ and Fig.~\ref{time_track:b} for the internal subsystem roll angle $\alpha$. For comparison purpose, the physical model-based EIC control performance~\cite{GetzPhD} is implemented and included in the figure. The EIC-based control is used as the benchmark and other physical model-based control designs (e.g., sliding mode control~\cite{Park2009b}, orbital stabilization~\cite{Shiriaev2007,Freid2009}, etc.) produce similar performance. The parameter values of the physical model are obtained from the vendor's manual and also validated by experimental tests. Figures~\ref{time_track:bb} and~\ref{time_track:c} compare the tracking errors $\bs{e}_\theta$ and $\bs{e}_\alpha$ under these two controllers. Figures~\ref{time_track:d} and~\ref{time_track:e} further shows the error mean and standard deviation profiles over multiple experimental runs. Table~\ref{table10} lists the comparison of the root mean square (RMS) errors and their deviations under these two controllers. It is clear from these results that the learning-based control design effectively captures the underactuated balance robotic dynamics and both the external tracking and internal balancing tasks are satisfactory. The performance under the learning-based design outperforms that with the physical model-based controller with more than 50\% reduction in mean values of errors and variances.     

\renewcommand{\arraystretch}{1.28}
\begin{table}[htb!]
	\vspace{-0mm}
	\begin{center}
		\setlength{\tabcolsep}{0.053in} 
		\caption{Root mean square (RMS) errors and their standard deviations of the base angle (deg) and roll angle (deg) comparison under two controllers for the rotary pendulum.}
		\vspace{2mm}
		\label{table10}
		\begin{tabular}{|c|p{17mm}|p{14mm}|p{14mm}|p{14mm}|}
			\hline\hline \multirow{2}{*}{}  & \multicolumn{2}{|c|}{\centering \hspace{4mm} EIC control \hspace{4mm}} & \multicolumn{2}{|c|}{Learning control}\\ \cline{2-5}
			& \hspace{7mm} $\theta$  & \hspace{5mm} $\alpha$ & \hspace{5mm} $\theta$ & \hspace{5mm} $\alpha$ \\ \hline
			RMSE  & $19.5 \pm 12.0$ & $3.3 \pm 2.0$ &$7.5\pm 4.8$ & $1.6 \pm 0.2$ \\
			\hline  \hline
		\end{tabular}
	\end{center}
	\vspace{-4mm}
\end{table}

\subsubsection{Bikebot experiments}

The bikebot system has high DOFs and sophisticated sensing and actuation components. Because the falling experiments would severely damage the hardware platform, for training data collection, the bikebot is controlled to track sinusoidal-shape trajectories under the EIC-based baseline controller. Different sinusoidal-shape trajectories are designed as $X_d= v_dt$, $Y_d= A_y \sin \left(\frac{2 \pi}{T_y} t\right)$, where $v_d=2$ m/s is the $x$-direction desired velocity, $A_y$ is the magnitude around the $y$-direction and $T_y=3.5$ s. The training data are collected by 7 different experiment trails and each of them lasts 7 s. In these experiments, $A_y$ values are chosen from $0.2$ m to $0.5$ m and the use of these different trajectories aims to perturb the bikebot dynamics. Figures~\ref{training_position} and~\ref{training_roll} show one trial of bikebot training data under the EIC-based controller.  

Using the trained model, we conduct the learning model-based control experiments to track various trajectories such as straight-lines, sinusoidal-shape ($0.8$ m peak-to-peak amplitude), and circular (around $3.8$ m radius) trajectories. For comparison purpose, we also conduct experiments and include the results under the physical model-based EIC controller. Figure~\ref{traj0} shows the comparison results under the learning-based and EIC controllers for one experimental run. It is clear that the trajectory tracking results under the learning-based control outperform these under the physical model-based EIC controller (Fig.~\ref{straight_track0}-\ref{circle_track0}). Similarly, the results shown in Fig.~\ref{straight_roll0}-\ref{circle_roll0} also demonstrate that the roll angles under the learning-based control oscillate less significantly than those under the EIC controller. The learning-based controller also demonstrates quicker reaction in circular tracking than the EIC controller.  

Figure~\ref{traj1} further shows the planar bikebot tracking errors and roll angle errors under the learning-based and EIC controllers. In the figure, we plot the trajectory and roll angle tracking errors and their deviations by using five experimental trials. Figures~\ref{straight_track}-\ref{circle_track} show the error and deviation profiles for straight-line, sinusoidal and circular trajectories, respectively. The root mean square errors (RMSE) in the $Y$-direction are listed in Table~\ref{table11} for both the learning-based and EIC-based controllers. It is clearly seen from these results that the learning-based control outperforms the EIC control. Figures~\ref{straight_roll}-\ref{circle_roll} show the roll angle tracking errors for three types of trajectories in multiple runs. The roll angle error magnitudes and variances under the learning-based controller are much smaller than these under the EIC controller and therefore, the learning controller results in agile and smooth tracking behaviors. In Table~\ref{table11}, we also list the RMSEs for the roll angles during these runs and these calculations confirm small variations under the learning control as shown in the figures. 

\renewcommand{\arraystretch}{1.2}
\begin{table}[htb!]
	\vspace{-0mm}
	\begin{center}
		\setlength{\tabcolsep}{0.05in} \caption{Root mean square errors (RMSEs) and their standard deviations of the tracking position (m) and roll angle (deg) comparison under two controllers for the bikebot.}
		\vspace{2mm}
		\label{table11}
		\begin{tabular}{|c|c|c|c|c|}
			\hline\hline \multirow{2}{*}{Trajectories}  & \multicolumn{2}{|c|}{\centering \hspace{4mm} EIC control \hspace{4mm}} & \multicolumn{2}{|c|}{Learning control}\\ \cline{2-5}
			  & Posit. & Roll & Posit. & Roll  \\ \hline 
Straight-line  & $0.6 \pm 0.2$ & $2.3 \pm 1.2$ &$0.3 \pm 0.1$ &$0.9 \pm 0.5$   \\ \hline 
Sinusoidal & $0.9 \pm 0.5$ & $4.1 \pm 2.3$ &$0.4 \pm 0.3$ &$2.0 \pm 1.1$  \\ \hline
Circular & $0.9 \pm 0.3$ & $2.5 \pm 1.5$ & $0.7 \pm 0.2$ & $1.7 \pm 1.1$  \\
\hline\hline
		\end{tabular}
	\end{center}
	\vspace{-3mm}
\end{table}

\begin{figure*}[htb!]
	\centering
	\subfigure[]{
		\label{trainingsize_vs_control_perform}
		\psfrag{A}[][]{\small $\|\bs{e}_\theta\|$ (rad)}
		\psfrag{B}[][]{\small $\|\bs{e}_\alpha\|$ (rad)}
		\includegraphics[width=3in]{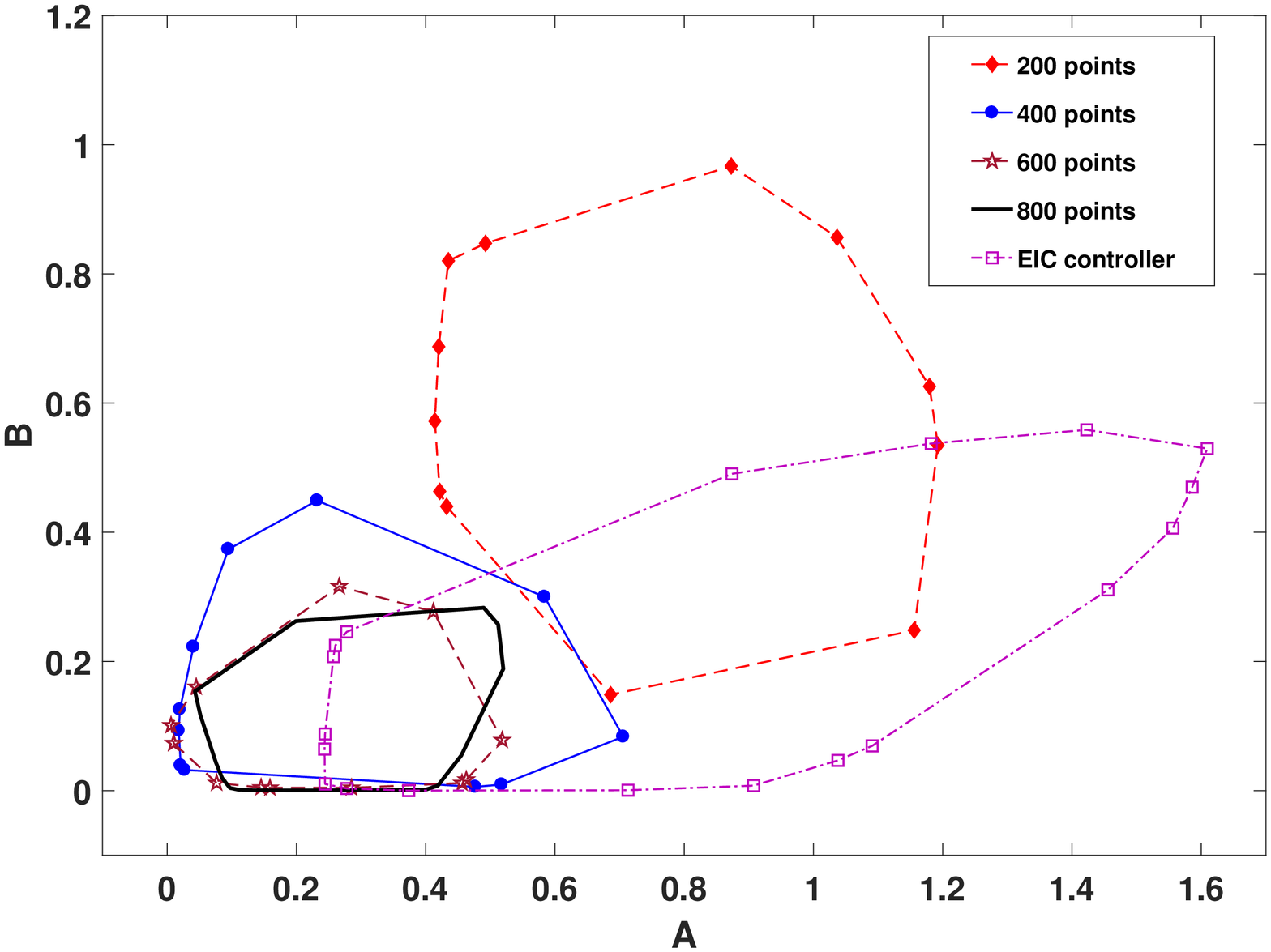}}
	\hspace{2mm}
	\subfigure[]{
		\label{tune_s}
		\psfrag{A}[][]{\small $\|\bs{e}_\theta\|$ (rad)}
		\psfrag{B}[][]{\small $\|\bs{e}_\alpha\|$ (rad)}
		\psfrag{ AAAAA}[][]{\scriptsize $\nu=0$}
		\psfrag{ BBBBB}[][]{\scriptsize $\nu=10$}
		\psfrag{ CCCCC}[][]{\scriptsize $\nu=40$}
		\psfrag{ DDDDD}[][]{\scriptsize $\nu=60$}
		\includegraphics[width=3.3in]{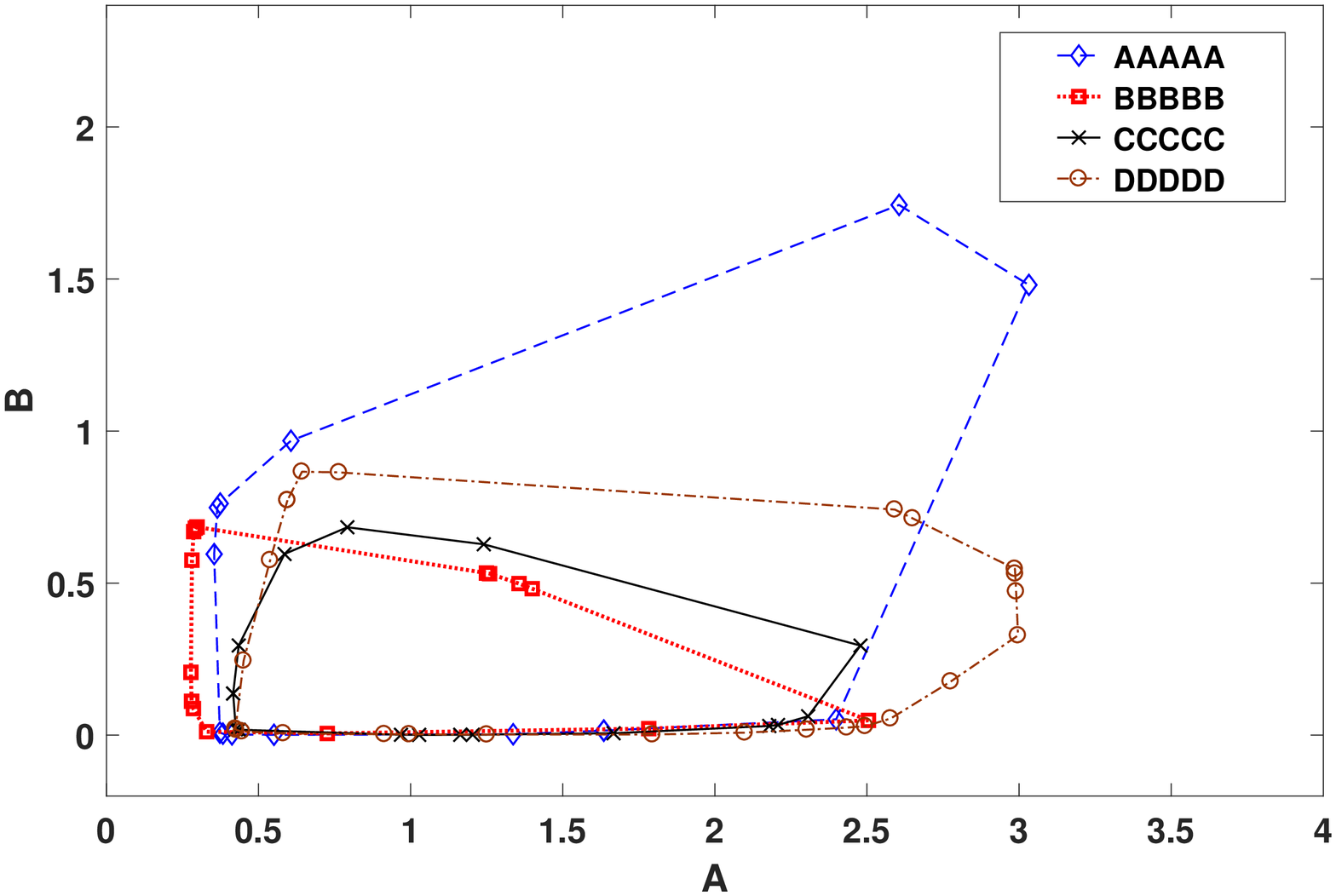}}
	\caption{(a) Comparison results of the internal balance error $\|\bs{e}_\alpha\|$ and external tracking error $\|\bs{e}_\theta\|$ under the learning-based control by various training data points and the EIC-based control for the rotary inverted pendulum. (b) Comparison of the balance and tracking errors $\|\bs{e}_\theta\|$ and $\|\bs{e}_\alpha\|$ under different values of the weight factor $\nu$.}
	\vspace{-0mm}
\end{figure*}

\begin{figure*}[htb!]
	\hspace{-2mm}
	\subfigure[]{
		\label{model_perform}
		\psfrag{Beta}[][]{\scriptsize  $\bs{\beta}\bs{\Sigma}^{\frac{1}{2}}$}
		\psfrag{NF}[][]{\scriptsize  $N_f/N_t$}
		\includegraphics[width=2.2in]{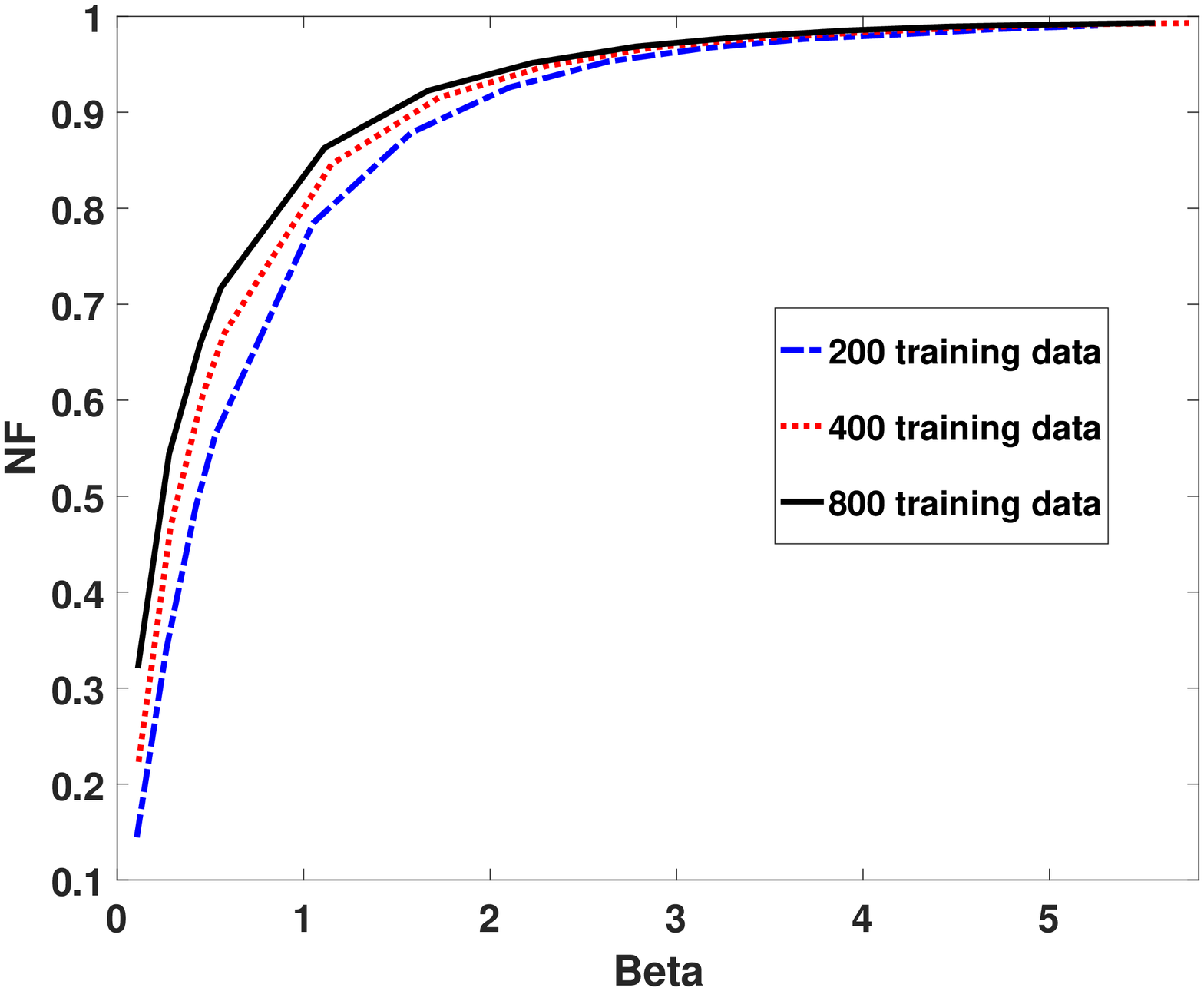}}
	\hspace{-1mm}
	\subfigure[]{
		\label{cost_perform}
		\psfrag{D}[][]{\scriptsize  $\Delta J$}
		\psfrag{NF}[][]{\scriptsize  $N_f/N_t$}
		\includegraphics[width=2.2in]{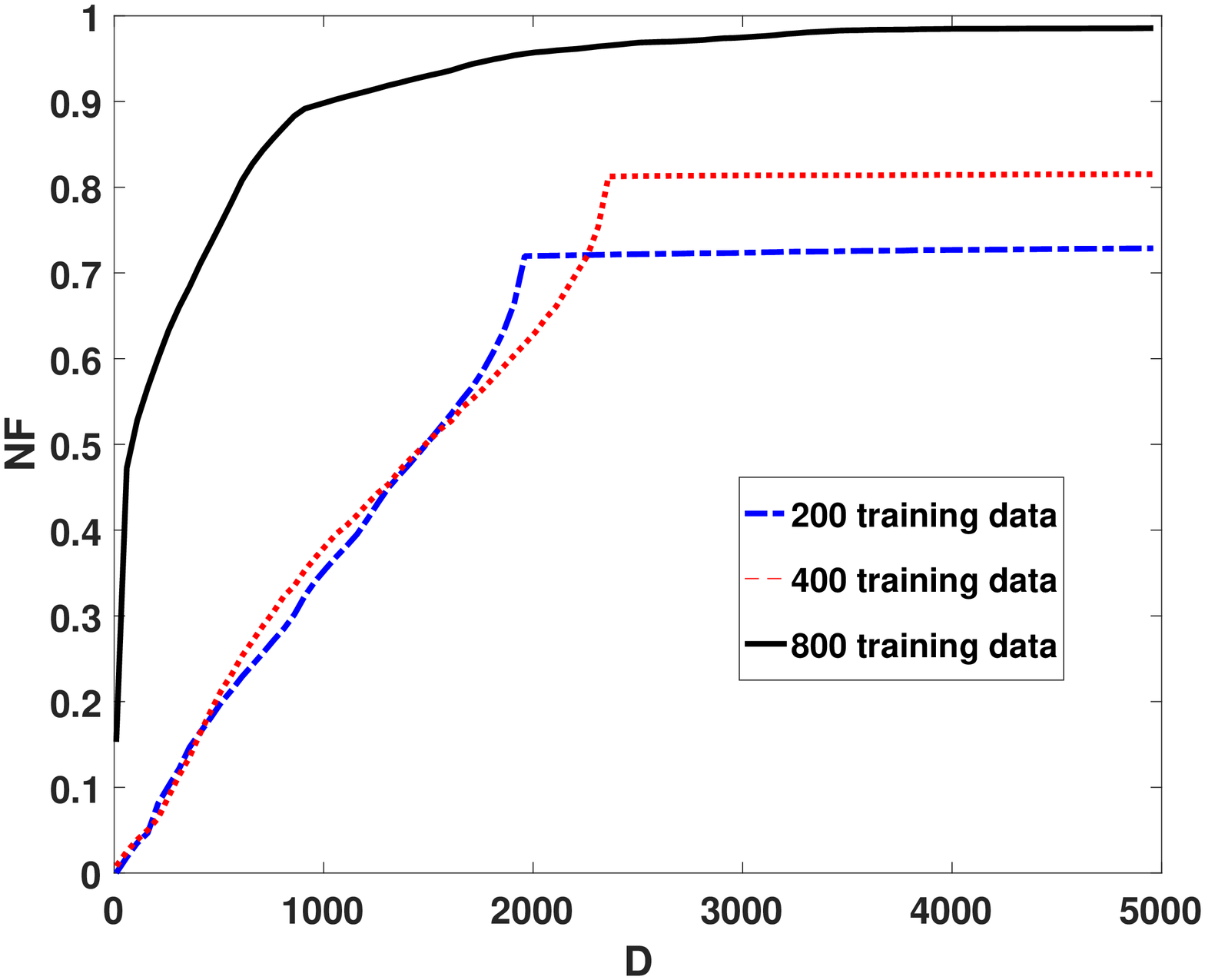}}
	\hspace{-3mm}
	\subfigure[]{
		\label{tracking_perform}
		\psfrag{E}[][]{\scriptsize  $\|\bs{e}\|_{ss}$}
		\psfrag{NF}[][]{\scriptsize  $N_f/N_t$}
		\includegraphics[width=2.2in]{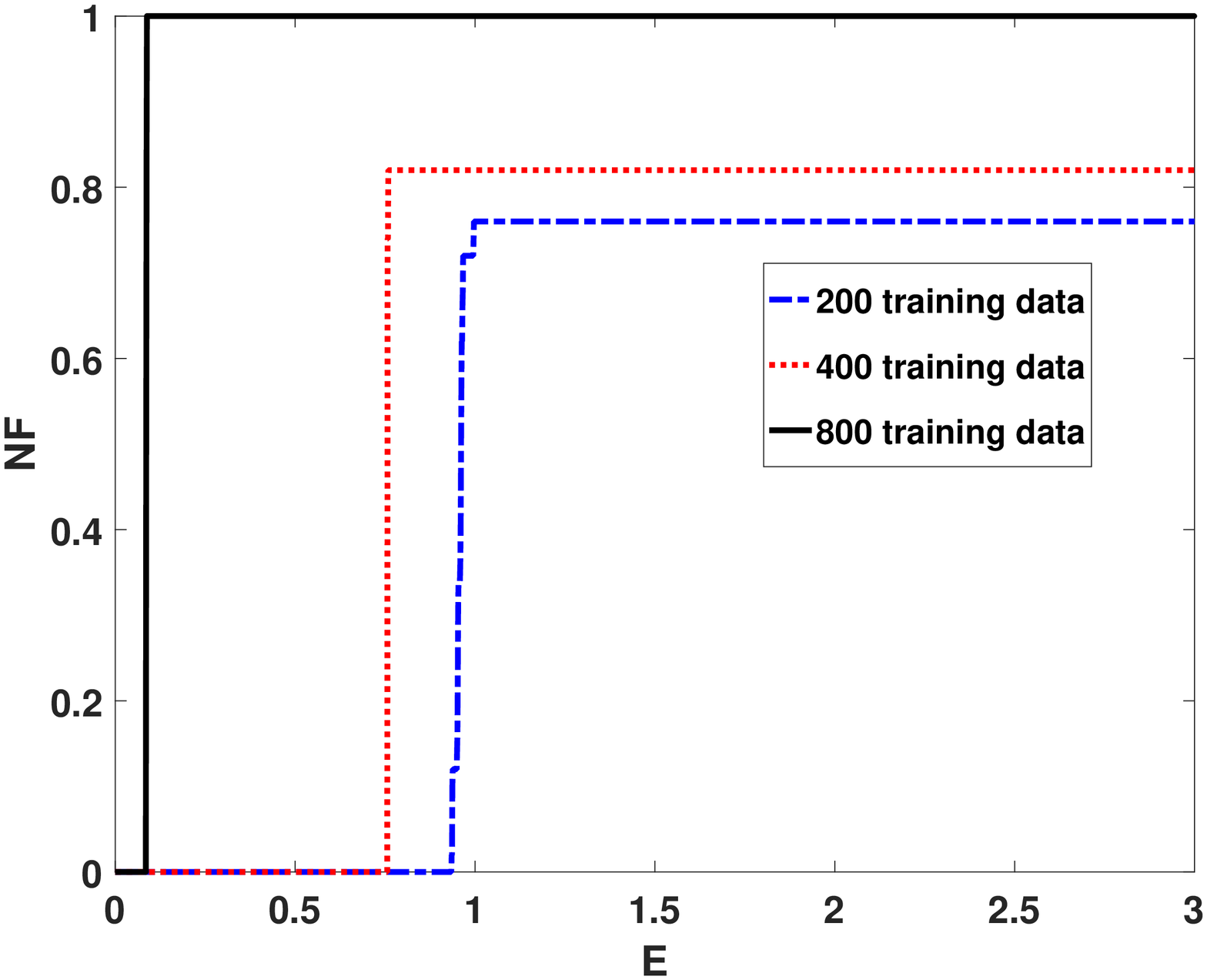}}
	\caption{(a) An approximated probability (computed as $N_f(\beta \Sigma^{\frac{1}{2}})/N_t$) of the learned model prediction accuracy from $\kappa_{\alpha}$ as function of training data variance. (b) An approximated probability (computed as $N_f(\Delta J)/N_t$)  as function of the estimated MPC cost function error. (c) An approximated probability (computed as $N_f(\|\bs{e} \|_{ss})/N_t$) as function of steady-state error.}
	\vspace{-0mm}
\end{figure*}

To understand the influence of training data on control performance, we first vary the sizes of the training data sets from $200$ to $800$ points to obtain different learned models in pendulum platform experiments. These models are used to track the same trajectory $\theta_d(t)$ as those in the experiments. Figure~\ref{trainingsize_vs_control_perform} shows the error distribution contours under different sizes of training data sets for the learning control and the EIC control. For each learned model, the plot includes the tracking errors of a 90-sec motion duration. The results clearly imply that with only 200 training data, the controller barely achieves the balancing and tracking tasks with large errors. With the increased training data points, the magnitudes of both the balancing and tracking errors decrease. With a set of 800 training data points, the learned model-based controller achieves superior performance than that under the analytical model-based controller. Theorem~\ref{stability} reveals that the error trajectory finally falls into a bounded regions and the plots in Fig.~\ref{trainingsize_vs_control_perform} demonstrate this error analysis.   
 
The trade-off between the tracking and balancing performance is tuned by the choice of $\nu$ value in the MPC objective function~(\ref{MPC_cost}). Experiments are conducted to demonstrate the performance with the same learned model under different values of $\nu$ using the rotary inverted pendulum. The learned model is obtained by using 200 training data points. We intentionally chose a slightly inaccurate learned model and the value of $\|\bs{\Sigma}_d\|$ in~(\ref{MPC_cost}) is relatively large. Figure~\ref{tune_s} shows the contours of the tracking and balancing errors with different $\nu$ values. These contours are plotted as the smallest convex cover of the corresponding error data points. When $\nu=0$, the system shows large error distributions due to the poor inverse dynamics model. With $\nu=10$, the system achieves a good trade-off between balancing and tracking tasks. But with a further increased $\nu$ value (i.e., $\nu=40,60$), the tracking performance becomes similar or slightly worse than those with $\nu=10$, and when $\nu>80$ the controller even fails to balance the pendulum. The average variances  of the inverse dynamics model for 60-second trials are $0.255$, $0.174$, $0.108$ and $0.108$ for $\nu=0,10,40,60$, respectively. The results clearly show that with increased $\nu$ values, the magnitude of $\bs{\Sigma}_d$ decreases. This confirms that the integration of $\|\bs{\Sigma}_d\|$ in the objective function helps improve the control performance.

\subsection{Discussions}

The accuracy of the learned models depends on the quality of the training data. We briefly conduct data quality analysis for control performance. We take and validate the performance analysis using rotary inverted pendulum control simulation. First, we evaluate the learned model prediction accuracy that is given in Lemma~\ref{gp_lemma} about estimate bounds. The training data is collected from the simulation of the ground truth dynamics with additive white noise. The learned model is built on the training data without any knowledge of the true model. The prediction accuracy could be quantified by the mean square errors between the outputs from the learned model and the true model. The prediction accuracy is quantified by the probability frequency of the output differences that fall in the error bound $\beta \Sigma^{\frac{1}{2}}$ as shown in  Lemma~\ref{gp_lemma}. To conduct such simulation experiment, the learned model is tested on $N_t=10,000$  independently randomly sampled  testing data. The testing data is sampled from a Gaussian distribution whose variance is larger than that of the training data. For each value of the error bound $\beta \Sigma^{\frac{1}{2}}$, we count frequency $N_f(\beta \Sigma^{\frac{1}{2}})$, i.e., the times that the difference between the outputs from the learned model and the true model falls in that error bound, and compute frequency ratio $\frac{N_f(\beta \Sigma^{\frac{1}{2}})}{N_t}$ as an approximation of the probability measure, that is, $\text{Prob}\approx \frac{N_f(\beta \Sigma^{\frac{1}{2}})}{N_t}$. Figure~\ref{model_perform} shows the frequency ratio (namely, probability) vesus the experimental cumulative distribution of the error bound $\beta \Sigma^{\frac{1}{2}}$. As the bound $\beta \Sigma^{\frac{1}{2}}$ becomes larger, the frequency ratio $\frac{N_f}{N_t}$ converges to one. As shown in Fig.~\ref{model_perform}, the converging speed of $\frac{N_f(\beta \Sigma^{\frac{1}{2}})}{N_t}$ to $1$ becomes faster when the number of training data increases. 

In MPC computation, the prediction error between the learned model and the true model is cumulated over the prediction horizon. The difference between the predicted trajectory $\bs{\hat{\theta}}(t)$  from the learned model $\bs{gp}_{\theta}$ and the actual trajectory $\bs{{\theta}}(t)$ from $f_{\theta}$ can be quantified by their cost function difference under the same input trajectory, that is, $\Delta J^k= \bar{J}_{\hat{\theta},\hat{W}^*}^k-\bar{J}_{{\theta},\hat{W}^*}^k$. In the simulation experiment, MPC is applied on $N_t$ values independently sampled from the Gaussian distribution. Similar to the above case, ratio $\frac{N_f(\Delta J)}{N_t}$ denotes the frequency (i.e., probability) that the cost function error is smaller than $\Delta {J}$. Figure~\ref{cost_perform} shows the experimental cumulative distribution of $\Delta {J}$.

Finally, the proposed controller is tested on $N_t$ trials whose initial conditions are independently sampled from the Gaussian distribution. The steady-state error $\|\bs{e} \|_{ss}$ for each trial is collected to quantify the control performance. For each value of $\|\bs{e} \|_{ss}$, the frequency of trials that end up with a steady-state error smaller than that value is counted as $\frac{N_f(\|\bs{e} \|_{ss})}{N_t}$. Figure~\ref{tracking_perform} shows the chosen initial condition frequency as the distribution of steady-state error  magnitude $\|\bs{e} \|_{ss}$. For example, the model trained from a data set of 400 points can drive the steady-state error below $0.76$ for about $82\%$ of the initial conditions. However, the system diverges for the other $18\%$ initial conditions due to the inaccuracy of the learned model prediction. The simulation results demonstrate the probabilistic behavior of the proposed learning-based controller. In practical sense, a learned model trained with 800 training data could stabilize the system for most of the situations.

\section{Conclusion and Future Work}
\label{concl}

This paper proposed a learning model-based control framework for underactuated balance robots. One characteristic of underactuated balance robots is that the equilibra of the internal subsystem depends on and varies according to the external subsystem trajectory tracking. The control design consisted an integrated trajectory tracking of the external subsystem and stabilization of the internal subsystem. The trajectory tracking of the external subsystem was designed through an MPC approach, while an inverse dynamics controller was used to simultaneously stabilize the planned internal subsystem trajectory. The GPs models were used to estimate the system dynamics and provide predictive distribution of model uncertainties. The control design explicitly incorporated prediction variances with tracking and stabilization performance through online optimization. The learned GP models were obtained without need of prior knowledge about the robotic systems dynamics nor successful balance demonstration. Moreover, the stability and closed-loop control performance were guaranteed through comprehensive closed-loop control systems analysis. We demonstrated the control systems design independently using two underactuated balance robotic platforms: a rotary inverted pendulum and a bikebot.  

We are currently working on testing the bikebot system on various terrain conditions to explore the performance under complex, dynamic environments. Real-time machine learning techniques are currently designed and developed on dedicated hardware to improve the control performance. Finally, quantitative analysis of training data quality is also among the future research directions. 

\section*{Acknowledgment}

The authors would like to thank Dr. Pengcheng Wang for valuable discussions on theoretical analysis and his help on bikebot experiments. The authors are also grateful to Yongbin Gong of Rutgers University for his implementation help of the vision-based localization system for bikebot experiments. 

\appendix
	
\newtheorem{lemm}{Lemma}[section]

\vspace{-2mm}
\section{Some Basic Properties}
		
\subsection{Gaussian Process}
\label{app_GP}

A Gaussian process (GP) is a collection of random variables, any finite number of which have a joint Gaussian distribution. A real value process $f(\bs{x}): \mathbb{R}^n \rightarrow \mathbb{R}$ is determined by its mean value function $\mu(\bs{x})$ and covariance function $k(\bs{x},\bs{x}')$ as $\mu(\bs{x})=\mathbb{E}[f(\bs{x})]$ and $k(\bs{x},\bs{x}')=\mathbb{E}[(f(\bs{x})-\mu(\bs{x}))(f(\bs{x}')-\mu(\bs{x}'))]$. Suppose the training data set contains $N$ input output data pairs $\mathcal{D}=\{\bs{x}_i,y_i \}_{i=1}^N$. The observed output $y_i$ is a noisy observation of the underlying function value with zero mean Gaussian noise $\varepsilon$,  i.e., $y_i=f(\bs{x}_i)+\varepsilon$ with $\varepsilon \sim \mathcal{N}(0,\sigma^2)$. The observation vector is denoted as $\bs{y}=[y_1 \, \cdots \,y_N]^T$ and the input design matrix is denoted as $\bs{X}=[\bs{x}^T_1\,\cdots\,\bs{x}^T_N]^T$. At a testing point $\bs{x}^* \in \mathbb{R}^n$, the function value $f^*$ is predicted by the observed training data $\mathcal{D}$. The joint distribution of $\bs{y}$ and the testing output ${f}^*$ is
\begin{align*}
\begin{bmatrix} \bs{y}\\ {f}^* \end{bmatrix} \sim
\mathcal{N}\left(\bs{0}, \begin{bmatrix} \bs{K}(\bs{X},\bs{X})+\sigma^2 \bs{I}_{N} & \bs{k}(\bs{X},\bs{x}^*) \\
\bs{k}(\bs{X},\bs{x}^*)^T & k(\bs{x}^*,\bs{x}^*)  \end{bmatrix}\right),
\end{align*}
where $\bs{K}(\bs{X},\bs{X})$ is an $N \times N$ kernel matrix whose element is $\bs{K}_{i,j}(\bs{X},\bs{X})=k(\bs{x}^i,\bs{x}^j)$. $\bs{k}(\bs{X},\bs{x}^*)$ is an $N\times1$ column vector whose element is $\bs{k}_i(\bs{X},\bs{x}^*)=k(\bs{x}^i,\bs{x}^*)$.

The probabilistic prediction of ${f}^*$ is given by the conditional distribution
\begin{equation*}
{f}^*|\bs{x}^*,\mathcal{D}  \sim \mathcal{N}(\mu(\bs{x}^*), \Sigma(\bs{x}^*) )
\end{equation*}
where $\mathcal{N}(\cdot,\cdot)$ represents a normal distribution, $\mu(\bs{x}^*)$ and $\Sigma(\bs{x}^*)$ are the posterior mean and covariance functions as
\begin{align}
\mu(\bs{x}^*)=&\bs{k}(\bs{X},\bs{x}^*)^T[\bs{K}(\bs{X},\bs{X})+\sigma^2\bs{I}_N]^{-1}\bs{y}, \nonumber \\
\Sigma(\bs{x}^*)=&k(\bs{x}^*,\bs{x}^*)-\bs{k}(\bs{X},\bs{x}^*)^T[\bs{K}(\bs{X},\bs{X})+ \nonumber \\
&\sigma^2 \bs{I}_N]^{-1}\bs{k}(\bs{X},\bs{x}^*).
\label{predictive_distribution}
\end{align}
GPs can also be applied to learn $n$-dimensional vector-valued function $\bs{f}(\bs{x}): \mathbb{R}^n \rightarrow \mathbb{R}^n$. In such cases, GPs are adopted to learn each function $f_i(\bs{x}), i=1,\dots,n$, as ${f}_i^*|\bs{x}^*,\mathcal{D}  \sim \mathcal{N}(\mu_i(\bs{x}^*), \Sigma_i(\bs{x}^*) )$  independently. The predictive distribution is written as 
\begin{equation}
\bs{f}^*|\bs{x}^*,\mathcal{D} \sim \mathcal{N}(\bs{\mu}(\bs{x}^*), \bs{\Sigma}(\bs{x}^*) ), 
\label{pred_dist_n}
\end{equation}
where $\bs{\mu}(\bs{x}^*)=[\mu_1(\bs{x}^*) \cdots \mu_n(\bs{x}^*)]^T$ and $\bs{\Sigma}(\bs{x}^*)=\diag\{\Sigma_1(\bs{x}^*), \cdots,\Sigma_n(\bs{x}^*)\}$. 

\subsection{GP-based estimation error bounds} 

The Gaussian process is determined by the covariance function (also called kernel function), which corresponds to a set of basis feature function in regression problem. The choice of the covariance function form depends on the input data and the commonly used covariance function is the squared exponential (SE) function as 
\begin{equation}
k(\bs{x}_i,\bs{x}_j)=\sigma_f^2 \exp\left(-\frac{1}{2}\Delta\bs{x}_{ij}^T\bs{W}\Delta\bs{x}_{ij}\right)+\sigma^2 \delta_{ij},
\label{SEkernel}
\end{equation}
where $\Delta \bs{x}_{ij}=\bs{x}_i-\bs{x}_j$, $\bs{W}$ is a positive definite weighting matrix, $\sigma_f^2$ and $\sigma^2$ are hyper-parameters, $\delta_{ij}=1$ if $i=j$; otherwise $\delta_{ij}=0$. The values of the above SE covariance function only depend on the distance between two points $\|\Delta \bs{x}_{ij}\|$. The following result gives the upper bound of covariance of the SE kernel. 
\begin{lemm}
\label{bound_sigma}
For any testing point $\bs{x}^* \in \mathbb{R}^n$, the posterior covariance $\Sigma(\bs{x}^*)$ of the SE kernel is bounded by $\Sigma(\bs{x}^*) \leq \sigma_f^2 + \sigma^2$, where $\sigma_f$ and $\sigma$ are the hyper-parameters in~(\ref{SEkernel}). For $n$-dimensional function $\bs{f}$ in~(\ref{pred_dist_n}), $\| \bs{\Sigma}(\bs{x}^*) \| \leq \max_{1\leq i \leq n} (\sigma_{f_i}^2 + \sigma_i^2)$,	where $\sigma_{f_i}$ and $\sigma_i$ are the hyper-parameters for corresponding $f_i$. 
\end{lemm}
\begin{IEEEproof}
From~(\ref{predictive_distribution}) and~(\ref{SEkernel}), noting the positive definiteness of $\bs{K}(\bs{X},\bs{X})+\sigma^2 \bs{I}_N$, we have  $\Sigma(\bs{x}^*) \leq k(\bs{x}^*,\bs{x}^*) \leq \sigma_f^2 + \sigma^2$. Since $\bs{\Sigma}(\bs{x}^*)=\diag\{\Sigma_1(\bs{x}^*),\cdots,\Sigma_n(\bs{x}^*)\}$, by the definition of matrix norm $\|\bs{\Sigma}(\bs{x}^*)\|= \lambda_{\max}(\bs{\Sigma}(\bs{x}^*)) \leq \max_{1\leq i \leq n} (\sigma_{f_i}^2 + \sigma_i^2)$. This proves the lemma.
\end{IEEEproof}

For a testing point $\bs{x}$, the predictive distribution conditioned on observations $\mathcal{D}$ is $\mathcal{N}(\mu(\bs{x}),\Sigma(\bs{x}))$. The following lemma gives the learning error bound.
\begin{lemm}[{\cite[Theorem~6]{GPtheory}}]Let $\delta \in (0,1)$, then 
	\label{gp_lemma}
\begin{equation*}
	\Pr \{| \mu(\bs{x}) - f(\bs{x}) | \leq \beta\Sigma^{\frac{1}{2}}(\bs{x}) \}\geq 1-\delta
\end{equation*}
with $\beta=\sqrt{ 2\|f\|_k^2+300\gamma \ln^3(\frac{N+1}{\delta}) }$, $\gamma \in \mathbb{R}$ is the maximum information gain defined as $\gamma=\max_{\bs{X}} I_g (\bs{y};f)$. 
\end{lemm}
The information gain in the above lemma is defined as ${I}_g(\bs{y};f)=H(\bs{y})-H(\bs{y}|f)$, where $H(\cdot)$ is the entropy function. In GP context, the prior distribution $\bs{y} \sim \mathcal{N}(\bs{0}, \bs{K}+\sigma^2 \bs{I}_{N})$ and the conditional distribution $\bs{y}|f \sim \mathcal{N}(\bs{0}, \sigma^2 \bs{I}_{N})$, the entropies are $H(\bs{y})=\frac{1}{2} \log \{ \det[2\pi e (\bs{K}+\sigma^2 \bs{I}_{N}) ] \}  $ and $H(\bs{y}|f)=\frac{1}{2} \log ( 2\pi e \sigma^2)$. Therefore, the information gain is ${I}_g(\bs{y};f)= \frac{1}{2}\log \det (\bs{I}_{N}+\sigma^{-2}\bs{K})$. According to~\cite{GPtheory}, the maximum information gain $\gamma$ for the SE kernel is in the order of $O((\ln(N)^{n+1} ))$. For $n$-dimensional vector function $\bs{f}(\bs{x})$, if every dimension is independent of each other, the results in Lemma~\ref{gp_lemma} are extended to the following lemma. 
\begin{lemm}[{\cite[Lemma~1]{Beckers}}] Let $\delta \in (0,1)$, then
\label{gp_lemma2}
\begin{equation*}
	\Pr\{\| \bs{\mu}(\bs{x}) - \bs{f}(\bs{x}) \| \leq \| \bs{\beta}^T\bs{\Sigma}^{\frac{1}{2}}(\bs{x}) \| \}\geq (1-\delta)^n,
	\end{equation*}
	where $\bs{\mu}(\cdot)$ and $\bs{\Sigma}(\cdot)$ are defined in~(\ref{pred_dist_n}), vector $\bs{\beta} \in \mathbb{R}^n$ and its $i$th element $\beta_i=\sqrt{ 2\|f_i\|_k^2+300\gamma_i \ln^3(\frac{N+1}{\delta})}$, and $\gamma_i$ is the maximum information gain for $f_i$.
\end{lemm}

\vspace{-2mm}
\section{Proofs of Main Results}

\subsection{Proof of Lemma ~\ref{lemma1}}
\label{proof_lemma1}

We calculate $\bs{\mu}_{\alpha}-\bs{\kappa}_{\alpha}(\dot{\bs{\alpha}}_2)$ by Tayor expansion as	
\begin{align*}
&\bs{\mu}_{\alpha}-\bs{\kappa}_{\alpha}(\dot{\bs{\alpha}}_2)=\bs{\mu}_{\alpha}-\bs{\kappa}_{\alpha}(\bs{v})+\bs{\kappa}_{\alpha}(\bs{v})-\bs{\kappa}_{\alpha}(\dot{\bs{\alpha}}_2)  \\
=&	\bs{\mu}_{\alpha}-\bs{\kappa}_{\alpha}(\bs{v})-\frac{\partial \bs{\kappa}_{\alpha}}{\partial \bs{v}} ( \dot{\bs{\alpha}}_2-\bs{v} )+O(\|\dot{\bs{\alpha}}_2-\bs{v} \|^2) \\
=&\bs{\mu}_{\alpha}-\bs{\kappa}_{\alpha}(\bs{v})-\frac{\partial \bs{ \kappa}_{\alpha}}{\partial \bs{v}} [ \bs{\mu}_{\alpha}- \bs{\kappa}_{\alpha} (\dot{\bs{\alpha}}_2) ]+ O(\|\dot{\bs{\alpha}}_2-\bs{v} \|^2). 	
\end{align*}
The third equality results from~(\ref{alpha_sub}). Note that $O(\|\dot{\bs{\alpha}}_2-\bs{v}\|^2)\leq c_2\|\bs{e}_{\alpha}\|^2+c_1\|\bs{e}_{\alpha}\|+c_0$, with constants $c_i>0$, $i=0,1,2$. 
	
Note that $\bs{A}_\kappa=\bs{I}+\frac{\partial \bs{ \kappa}_{\alpha}}{\partial \bs{v}}$ is the linearization of the left-hand side of the second equation (i.e., $\bs{\alpha}_2$ subdynamics) of~(\ref{robot_dyn_reshape}) with respect to $\dot{\bs{\alpha}}_2$. It serves like the inertia matrix of $\bs{\alpha}_2$-dynamics and therefore, $\bs{A}_\kappa$ is non-singular and also positive definite. From the above equation and $\dot{\bs{\alpha}}_2$ in~(\ref{robot_dyn_reshape}), we have 
\begin{equation*}
\bs{\mu}_{\alpha}-\bs{\kappa}_{\alpha}(\dot{\bs{\alpha}}_2)=\bs{A}^{-1}_\kappa[O(\|\dot{\bs{\alpha}}_2-\bs{v}\|^2)
	+\bs{\mu}_{\alpha}- \bs{\kappa}_{\alpha}(\bs{v})]. 
\end{equation*}
Taking the norm and applying the results in Lemma~\ref{gp_lemma2} to $\bs{\mu}_{\alpha}-\bs{\kappa}_{\alpha}(\bs{v})$, we have 
\begin{align*}
\Pr \Bigl\{\bs{\Pi}=&\Bigl\{\|\bs{\mu}_{\alpha}-\bs{\kappa}_{\alpha}(\dot{\bs{\alpha}}_2)\|\leq \lambda_{\min}^{-1}(\bs{A}_\kappa)\bigl(\sum_{i=0}^2 c_i\|\bs{e}_{\alpha}\|^i+\\
&\|\bs{\beta}_{\alpha}^T \bs{\Sigma}_{\alpha}^{\frac{1}{2}}\|\bigr)\Bigr\}\Bigr\} \geq (1-\delta)^n,
\end{align*}
where $\lambda_{\min}(\bs{A}_\kappa)>0$ is the smallest eigenvalue of $\bs{A}_\kappa$. Defining 
\begin{equation}
\rho(\bs{e}_{\alpha},\bs{\theta})=\lambda_{\min}^{-1}(\bs{A}_\kappa)
\Bigl(\sum_{i=0}^2 c_i\|\bs{e}_{\alpha}\|^i+\|\bs{\beta}_{\alpha}^T \bs{\Sigma}_{\alpha}^{\frac{1}{2}}\|\Bigr),
\label{rho2}
\end{equation}
we prove the lemma. 

\vspace{-1mm}
\subsection{Proof of Lemma~\ref{e_lemma}}
\label{proof_e_lemma}

Defining $\lambda_1=-\frac{1}{2}(k_d-\sqrt{k_d^2-4k_p})<0$ and $\lambda_2=-\frac{1}{2}(k_d+\sqrt{k_d^2-4k_p})<0$, we first show that $\bs{A}$ is diagonalizable with $n$-eigenvalue as $\frac{\lambda_1}{\epsilon}$ and the other $n$-eigenvalue as $\frac{\lambda_2}{\epsilon}$. To see that, introducing nonsingular matrix $\bs{M}$ and diagonal matrix $\bs{\Lambda}$ as   
\begin{equation}
\bs{M}=\begin{bmatrix}
	\epsilon \bs{I}_n & \epsilon \bs{I}_n \\
	\lambda_1 \bs{I}_n & \lambda_2 \bs{I}_n
	\end{bmatrix}, \;
	\bs{\Lambda}=
\begin{bmatrix}
\frac{\lambda_1}{\epsilon} \bs{I}_n & \bs{0}\\
\bs{0} & 		\frac{\lambda_2}{\epsilon} \bs{I}_n
\end{bmatrix},
\label{cond00}
\end{equation}
it is straightforward to verify that $\bs{A}=\bs{M} \bs{\Lambda} \bs{M}^{-1}$. To assess the convergence property of $\bs{e}_{\alpha}$, we introduce $\bs{e}_{\alpha}= \bs{M} \bs{e}_{\alpha'}$ and error dynamics~(\ref{track_err}) becomes
\begin{equation}
\dot{\bs{e}}_{\alpha'}=\bs{\Lambda} \bs{e}_{\alpha'}+\bs{M}^{-1}\bs{B}[\bs{r}(t) +\bs{\mu}_{\alpha}-\bs{\kappa}_{\alpha}(\dot{\bs{\alpha}}_2)].
\label{ey_dyn}
\end{equation}
Since $\bs{\Lambda}$ is Hurwitz, there exists a positive definite matrix $\bs{P}_\alpha$ such that $\bs{\Lambda}^T\bs{P}_\alpha+\bs{P}_\alpha\bs{\Lambda}=-\bs{I}_{2n}$. We choose the Lyapunov function candidate $V_{\alpha'}=\bs{e}_{\alpha'}^T \bs{P}_\alpha \bs{e}_{\alpha'}$ for~(\ref{ey_dyn}) and then 
\begin{align*}
\dot{V}_{\alpha'}=&-\bs{e}_{\alpha'}^T\bs{e}_{\alpha'}+2(\bs{B}^T \bs{M}^{-T}\bs{P}_\alpha \bs{e}_{\alpha'})^T[\bs{r}+\bs{\mu}_{\alpha}-\nonumber \\
& \bs{\kappa}_{\alpha}(\dot{\bs{\alpha}}_2)].
\end{align*}
Auxiliary control $\bs{r}(t)$ is designed as $\bs{r}(t)=-\rho(\bs{e}_\alpha,\bs{\theta}) \frac{\bs{B}^T \bs{M}^{-T}\bs{P}_\alpha \bs{e}_{\alpha'}}{\|\bs{B}^T \bs{M}^{-T}\bs{P}_\alpha \bs{e}_{\alpha'}\|}$ if $\|\bs{B}^T \bs{M}^{-T}\bs{P}_\alpha \bs{e}_{\alpha'}\|> \xi$; $\bs{r}(t)=-\frac{\rho(\bs{e}_\alpha,\bs{\theta})}{\xi} {\bs{B}^T \bs{M}^{-T}\bs{P}_\alpha \bs{e}_{\alpha'}}$ if $\|\bs{B}^T \bs{M}^{-T}\bs{P}_\alpha \bs{e}_{\alpha'}\| \leq \xi$ for constant $\xi>0$ and $\rho(\bs{e}_\alpha,\bs{\theta})$ is defined by~(\ref{rho2}). 

According to Lemma~\ref{e_lemma}, with the above design and choosing $\xi=\frac{\lambda_{\min}(\bs{A}_\kappa)}{c_2\|\bs{M}\|^2}$, we obtain 
\begin{align}
\dot{V}_{\alpha'} \leq& -\|\bs{e}_{\alpha'}  \|^2 +\frac{\xi \rho(\bs{e}_\alpha,\bs{\theta})}{2}= -\frac{1}{2}\|\bs{e}_{\alpha'}  \|^2+\nonumber \\
&\frac{c_1}{2c_2\|\bs{M}\|}\|\bs{e}_{\alpha'}\|+\frac{c_0}{2c_2\|\bs{M}\|^2}+\frac{\|\beta_\alpha^T\bs{\Sigma}_\alpha^{1/2}\|}{2c_2\|\bs{M}\|^2} \nonumber \\
=&-\frac{1}{4}\|\bs{e}_{\alpha'}\|^2-\frac{1}{4}\left(\|\bs{e}_{\alpha'}\|-\frac{c_1}{c_2\|\bs{M}\|}\right)^2+c_3 \nonumber \\
\leq & -\frac{1}{4}\|\bs{e}_{\alpha'}\|^2+c_3, \label{cond11}
\end{align}
where $c_3=\frac{1}{4}\frac{c_1^2}{c^2_2\|\bs{M}\|^2}+\frac{1}{2}\frac{c_0}{c_2\|\bs{M}\|^2}+\frac{1}{2} \frac{\|\beta_\alpha^T\bs{\Sigma}_\alpha^{1/2}\|}{2c_2\|\bs{M}\|^2}>0$. Since $\bs{P}_\alpha$ is the solution of Lyapunov equation with $\bs{\Lambda}$ in~(\ref{cond00}), it has $n$-eigenvalue at $-\frac{\epsilon}{2\lambda_2}>0$ and the other $n$-eigenvalue at $-\frac{\epsilon}{2\lambda_1}>0$. Thus, we obtain ${V}_{\alpha'}\leq -\frac{\epsilon}{2\lambda_1} \|\bs{e}_{\alpha'}\|^2 $. Using this result, from~(\ref{cond11}), we obtain $\dot{V}_{\alpha'}\leq \frac{\lambda_1}{2\epsilon}{V}_{\alpha'}+c_3$ and therefore, 
\begin{equation}
{V}_{\alpha'}(t) \leq {V}_{\alpha'}(0) e^{\frac{\lambda_1}{2\epsilon}t}-\frac{2\epsilon}{\lambda_1}c_3.
\label{cond12} 
\end{equation}

Considering $V_{\alpha}(t)=\bs{e}_{\alpha}(t)^T \bs{P}\bs{e}_{\alpha}(t)$ and positive definiteness of $\bs{P}=\bs{M}^{-T}\bs{P}_\alpha\bs{M}^{-1}$, it is straightforward to check that $V_{\alpha}(t)=\bs{e}_{\alpha'}(t)^T \bs{M}^T \bs{P}\bs{M}\bs{e}_{\alpha'}(t)=\bs{e}_{\alpha'}(t)^T \bs{P}_\alpha \bs{e}_{\alpha'}(t) =V_{\alpha'}(t)$. Defining $\bs{Q}=\bs{M}^{-T}\bs{M}^{-1}$, $\bs{P}$ is the solution of Lyapunov equation $\bs{A}^T\bs{P}+\bs{P}\bs{A}=-\bs{Q}$. Using~(\ref{cond11}), we obtain 
\begin{equation}
\dot{V}_\alpha(t) \leq -\frac{1}{4}\bs{e}_\alpha^T\bs{Q}\bs{e}_\alpha+c_3.
\label{cond111}
\end{equation}
Using $\bs{P}$ and $\bs{e}_{\alpha}$, we write the control input $\bs{r}(t)$ as in~(\ref{cond01}). Noting that $\lambda_{\min}(\bs{P}) \|\bs{e}_{\alpha}\|^2 \leq V(t)_\alpha  \leq \lambda_{\max}(\bs{P}) \|\bs{e}_{\alpha}\|^2$, from~(\ref{cond12}), we have
\begin{align*}
\| \bs{e}_{\alpha}(t)\| \leq& \sqrt{\frac{\lambda_{\max}(\bs{P})}{\lambda_{\min}(\bs{P})}}\| \bs{e}_{\alpha}(0)\|e^{\frac{\lambda_1}{4\epsilon}t}+\sqrt{-\frac{2\epsilon c_3}{\lambda_1\lambda_{\min}(\bs{P})}}\\
=&d_1 \| \bs{e}_{\alpha}(0)\|e^{\frac{\lambda_1}{4\epsilon}t}+d_2
\end{align*}
with $d_1$ and $d_2$ are given in the lemma.    

\vspace{-1mm}
\subsection{Proof of Lemma~\ref{Sigma_theta_bound}}
\label{proof_Sigma_theta_bound}

Taking norm on both sides of~(\ref{mean_var}) and applying the upper-bound of the gradient $\|\frac{\partial \bs{\mu}_{gp_{\theta}} }{\partial \bs{\theta} }\| \leq L_1$, we obtain
\begin{align*}
\|\bs{\Sigma}_{\hat{\theta}}(k+i+1|k)\| \leq & (\|\bs{F}\|^2+\|\bs{G}\|^2L_1^2)\|\bs{\Sigma}_{\hat{\theta}}(k+i|k)\|\\ 
&+\|\bs{G}\|^2\|\bs{\Sigma}_{gp_{\theta}}\| \\
\text{\hspace{-0mm}}\leq & (\|\bs{F}\|^2+\|\bs{G}\|^2 L_1^2)\|\bs{\Sigma}_{\hat{\theta}}(k+i|k)\| \\
& + \|\bs{G}\|^2  \sigma^{2}_{\bs{f}\max}.
\end{align*}
Applying the above process iteratively with $\bs{\Sigma}_{\hat{\theta}}(k|k)=\bs{0}$, we have 
\begin{equation*}
\|\bs{\Sigma}_{\hat{\theta}}(k+i|k)\|\leq	
	\frac{1-(\|\bs{F}\|^2+\|\bs{G}\|^2L_1^2)^i}{1-(\|\bs{F}\|^2+\|\bs{G}\|^2L_1^2)}\|\bs{G}\|^2
\sigma^{2}_{\bs{f}\max}.
\end{equation*}
From~(\ref{FG}), the norms of $\bs{F}$ and $\bs{G}$ are calculated as 
\begin{equation*}
\|\bs{F}\|=\sqrt{1+\frac{\Delta t}{2}\left(\Delta t+\sqrt{(\Delta t)^2+4}\right)}, \; 
\|\bs{G}\|=\Delta t.
\end{equation*}
For $\Delta t \ll 1$, taking approximation $\|\bs{F}\|\approx 1$ and fact that $(1+x)^n\approx 1+nx$ for $|x|\ll 1$, we obtain the upper-bound as shown in the lemma.  

\vspace{-1mm}
\subsection{Proof of Lemma~\ref{lemma_terminal}}
\label{proof_lemma_terminal}

It is straightforward to obtain 
\begin{align*}
l_f(k+H+2)=&l_f^*(k+H+2)+\tr(\bs{Q}_3 \bs{\Sigma}_{\hat{\theta}}(k+H+2))\\
\leq& l_f^*(k+H+1)-l_s^*(k+H+1)\\
& +\tr(\bs{Q}_3 \bs{\Sigma}_{\hat{\theta}}(k+H+2)) \\
\leq& l_f(k+H+1)-l_s^*(k+H+1)\\
&+\tr(\bs{Q}_3 \bs{\Sigma}_{\hat{\theta}}(k+H+2)).
\end{align*}
Under conditions~(\ref{input_domain}), we obtain $l_s(k+H+1)\leq l_s^*(k+H+1)+\tr(\bs{Q}_3 \bs{\Sigma}_{\hat{\theta}}(k+H+1))$ and combining with the above inequality, the proof is completed.  

\vspace{-1mm}
\subsection{Proof of Lemma~\ref{mpc_lemma}}
\label{proof_mpc_lemma}

We first show the decreasing property of $J^k_{\hat{\theta}^*,\hat{W}^*}$. Inspired by the approach in~\cite{MayneMPCbook}, we take the technique to construct a following intermediary policy $\hat{\bs{W}}^e(k+1)$ extended from $\hat{\bs{W}}^*(k)$ as
\begin{align*}
\hat{\bs{W}}^e(k+1)=&\{ \hat{\bs{\alpha}}^e(k+1), \hat{\bs{w}}^e(k+i+1), \bs{u}_f^e(k+i+1), \\
&i=0,\ldots,H\},
\end{align*}
where $\hat{\bs{\alpha}}_1^e(k+1)=\hat{\bs{\alpha}}_1^*(k)+\hat{\bs{\alpha}}_2^*(k) \Delta t$,  $\hat{\bs{\alpha}}_2^e(k+1)=\hat{\bs{\alpha}}_2^*(k)+\hat{\bs{W}}^*(k) \Delta t$,  and $\hat{\bs{w}}^e(k+i)=\hat{\bs{W}}^*(k+i) $, $\bs{u}_f^e(k+i)=\bs{u}_f^*(k+i)$ for $i=1,\ldots,H$, and $\hat{\bs{w}}^e(k+H+1)$ and $\bs{u}_f^e(k+H+1)$ satisfy~(\ref{input_domain}). The choice of the above design guarantees that inputs \{$\hat{\bs{\alpha}}^e(k+i)$,$\hat{\bs{w}}^e(k+i)$,$\bs{u}_f^e(k+i)$\} of $\hat{\bs{W}}^e_H(k+1)$ are the same as \{$\hat{\bs{\alpha}}^*(k+i)$,$\hat{\bs{W}}^*(k+i)$,$\bs{u}_f^*(k+i)$\} of $\hat{\bs{W}}^*(k)$ for $i=1,\ldots,H$. Consequently, the predicted states $\bs{\mu}_{\hat{\theta}}^e(k+i)$, $\bs{\Sigma}_{\hat{\theta}}^e(k+i)$ by~(\ref{mean_var}) under $\hat{\bs{W}}^e(k+1)$ are the same as these under control $\hat{\bs{W}}^*(k)$ at these steps. Let $l_s^e(k+i)$ ($l_f^e(k+i)$) and $l_s^*(k+i)$ ($l_f^*(k+i)$)  denote the stage and terminal costs under controls~$\hat{\bs{W}}^e(k+1)$ and $\hat{\bs{W}}^*(k)$, respectively. It is then straightforward to obtain that	$l_s^e(k+i)= l_s^*(k+i)$ for $i=1,\ldots,H$ and	$l_f^e(k+H+1)=l_f^*(k+H+1)$. Therefore,
\begin{align*}
J^{k+1}_{\hat{\theta}^e,\hat{W}^e}-&J^k_{\hat{\theta}^*,\hat{W}^*}=l_s^e(k+H+1)
+ l_f^e(k+H+2)\\
&-l_s^*(k)-l_f^e(k+H+1)+ \nu \Delta \bs{\Sigma}_{dk}^{e*}+\Delta \hat{\bs{\alpha}}_{\bs{Q}_2k}^*,
\end{align*}
where $\Delta \bs{\Sigma}_{dk}^{e*}= \| \bs{\Sigma}_d^{\hat{W}^e}(k+1) \|- \| \bs{\Sigma}_d^{\hat{W}^*}(k)) \| $ and
$\Delta \hat{\bs{\alpha}}_{\bs{Q}_2k}^*=\|\hat{\bs{\alpha}}^e(k+1) \|_{\bs{Q}_2}^2 -\|\hat{\bs{\alpha}}^*(k) \|_{\bs{Q}_2}^2$. Noting that $\hat{\bs{w}}^e(k+H+1)$ and $\bs{u}_f^e(k+H+1)$ satisfy~(\ref{input_domain}), by Lemma~\ref{lemma_terminal}, we have 
\begin{align*}
J^{k+1}_{\hat{\theta}^e,\hat{W}^e}&-J^k_{\hat{\theta}^*,\hat{W}^*} \leq -l_s^*(k)+ \nu \Delta \bs{\Sigma}_{dk}^{e*}+\Delta \hat{\bs{\alpha}}_{\bs{Q}_2k}^*+ \\
&  \tr(\bs{Q}_1 \bs{\Sigma}_{\hat{\theta}}(k+H+1))+\tr(\bs{Q}_3 \bs{\Sigma}_{\hat{\theta}}(k+H+2)).
\end{align*}
Because of $J^{k+1}_{\hat{\theta}^*,\hat{W}^*}\leq J^{k+1}_{\hat{\theta}^e,\hat{W}^e}$, from the above result, we have 
\begin{align}
J^{k+1}_{\hat{\theta}^*,\hat{W}^*}-J^k_{\hat{\theta}^*,\hat{W}^*} \leq &-\lambda_{\min}(\bs{Q}_1) \|\bs{e}_{\theta}(k) \|^2+ \nu \Delta \bs{\Sigma}_{dk}^{e*} \nonumber \\
&+\Delta \hat{\bs{\alpha}}_{\bs{Q}_2k}^*+\tr(\bs{Q}_1 \bs{\Sigma}_{\hat{\theta}}(k+H+1)) \nonumber  \\
&+\tr(\bs{Q}_3 \bs{\Sigma}_{\hat{\theta}}(k+H+2)).
\label{J_decreasing}
\end{align}
From Lemma~\ref{Sigma_theta_bound}, $\bs{\Sigma}_{\hat{\theta}}(k+H+1)\leq (H+1)(\Delta t)^2 \sigma^2_{{f}\max}$. From Lemma~\ref{bound_sigma}, $\Delta \bs{\Sigma}_{dk}^{e*}\leq \| \bs{\Sigma}_d^{\hat{W}^e}(k+1))\|\leq \max_{1\leq i \leq n} (\sigma_{{\alpha_i}}^2 + \sigma_i^2):=\sigma^2_{\kappa\max}$. Letting $\|\bs{\alpha}(k+1)\|_{\bs{Q}_2}^2 \leq \alpha_{\max}^2$ as the constant upper-bound, we have $\Delta \hat{\bs{\alpha}}_{\bs{Q}_2k}^*  \leq \|\hat{\bs{\alpha}}^e(k+1) \|_{\bs{Q}_2}^2 \leq \alpha_{\max}^2$ and thus, 
\begin{align}
J^{k+1}_{\hat{\theta}^*,\hat{W}^*}-J^k_{\hat{\theta}^*,\hat{W}^*} \leq& -\lambda_{\min}(\bs{Q}_1) \|\bs{e}_{\theta}(k) \|^2+ \nu \sigma^2_{\bs{\kappa}\max} \nonumber \\
&+ \alpha_{\max}^2+m [\lambda_{\max}(\bs{Q}_1)+\lambda_{\max}(\bs{Q}_3) ] \nonumber \\
&(H+2)(\Delta t)^2 \sigma^2_{\bs{f}\max}.
\label{value_function_decreasing}
\end{align}
Furthermore, from the definition of $J^k_{\hat{\bs{\theta}}^*,\hat{\bs{W}}^*}$, we have $J^{k+1}_{\hat{\theta}^*,\hat{W}^*} \geq \lambda_{\min}(\bs{Q}_1)\|\bs{e}_{\bs{\mu}_{\hat{\theta}}}(k+1)\|^2$. By the monotonicity of the value function (Lemma 2.15 in~\cite{MayneMPCbook}), we have 
\begin{align*}
J^k_{\hat{\theta}^*,\hat{W}^*} \leq &  l_f(k)+\|\hat{\bs{\alpha}}^*(k)\|_{\bs{Q}_2}^2+\nu\|\bs{\Sigma}_d^{\hat{W}^*}(k))\| \nonumber \\
\leq& \lambda_{\max}(\bs{Q}_3) \| \bs{e}_{\theta}(k)\|^2+\alpha_{\max}^2+ \nu \sigma_{\kappa \max}^2. 
\label{value_function_monotocity}
\end{align*}
Substituting the above inequalities into~(\ref{value_function_decreasing}) to cancel $\|\bs{e}(k) \|^2$, we obtain $J^{k+1}_{\hat{\theta}^*,\hat{W}^*} \leq d_3 J^{k}_{\hat{\theta}^*,\hat{W}^*} + d_4 $ with $d_3=1-\frac{\lambda_{\min}(\bs{Q}_1)}{\lambda_{\max}(\bs{Q}_3)} <1$ and $d_4=\left[1+\frac{\lambda_{\min}(\bs{Q}_1)}{\lambda_{\max}(\bs{Q}_3)}\right](\nu \sigma^2_{{\kappa}\max} +  \alpha_{\max}^2)+ m\lambda_m(H+2)\Delta^2 t \sigma^2_{{f}\max} $ where $\lambda_m=\lambda_{\max}(\bs{Q}_1)+\lambda_{\max}(\bs{Q}_3)$. Therefore, 
\begin{equation}
J^{k+i}_{\hat{\theta}^*,\hat{W}^*} \leq d_3^i J^{k}_{\hat{\theta}^*,\hat{W}^*} + d_4 \frac{1-d_3^i}{1-d_3},
\end{equation}
and consequently, $\|\bs{e}_{\bs{\mu}_{\hat{\theta} } }(k+i)\|\leq a_4(i)\|\bs{e}_{\theta}(k)\|+a_5(i)$ where $a_4(i)=d_3^{\frac{i}{2}}\sqrt{\frac{\lambda_{\max}(\bs{Q}_3)}{\lambda_{\min}(\bs{Q}_1)} }$ and $a_5(i)=\sqrt{\frac{d_3^i(\alpha_{\max}^2+ \nu \sigma_{\kappa \max}^2)+d_4\frac{1-d_3^i}{1-d_3}}{\lambda_{\min}(\bs{Q}_1)  }}$. This proves the lemma. 

\vspace{-1mm}
\subsection{Proof of Lemma~\ref{err1_lemma}}
\label{proof_err1_lemma}

Plugging the iterative relation~(\ref{mean_var}) for $\bs{\mu}_{\hat{\theta}}(k+i|k)$ and counterpart for $\bs{\mu}_{{\theta}}(k+i|k)$ into $\tilde{\bs{\mu}}_\theta(k+i)$, the difference is then  
\begin{align*}
\tilde{\bs{\mu}}_\theta(k+i)=&\|\bs{F}\| \|\tilde{\bs{\mu}}_\theta(k+i-1)\| +\|\bs{G}\| \\
& \|\bs{\mu}_{gp_{\theta}}(\bs{\mu}_{\theta}(k+i-1),\bs{\alpha}(k+i-1))-\\
&\bs{\mu}_{gp_{\theta}}(\bs{\mu}_{\hat{\theta}}(k+i-1|k),\hat{\bs{\alpha}}(k+i-1|k))\| \\
\leq& \|\bs{F}\| \|\tilde{\bs{\mu}}_\theta(k+i-1)\|
	+ L_2 \|\bs{e}_{\alpha}(k+i-1)\|  \\
& + L_3\|\bs{G}\| \|\tilde{\bs{\mu}}_\theta(k+i-1)\| \\    
=& (\|\bs{F}\|+L_3\|\bs{G}\|) \|\tilde{\bs{\mu}}_\theta(k+i-1)\|+\\
&L_2 \|\bs{G}\| \|\bs{e}_{\alpha}(k+i-1)\|.  
\end{align*}
In the above derivations, we use the Lipshitz assumptions. For small sampling period $\Delta t \ll 1$, $\|\bs{F}\| \approx 1$ and $\|\bs{G}\|=\Delta t$, when $i=1$, with the fact that $\bs{\mu}_{\theta}(k|k)=\bs{\mu}_{\hat{\theta}}(k|k)$, we have $\tilde{\bs{\mu}}_\theta(k+1) \leq L_2 \Delta t \|\bs{e}_{\alpha}(k)\|$. For $i\geq 2$, applying the above process iteratively, we obtain 
\begin{align*}
\tilde{\bs{\mu}}_\theta(k+i) \leq&  \sum_{j=0}^{i-1} (\|\bs{F}\|+L_3\|\bs{G}\|)^{i-j-1}
	L_2\|\bs{G}\| \\
& \Bigl[ d_1 e^{ \frac{\lambda_1}{4 \epsilon} j\Delta t } \|\bs{e}_{\alpha}(k)\| 
	+d_2 \Bigr],
\end{align*}
where the result in Lemma~\ref{e_lemma} is used to obtain $\|\bs{e}_{\alpha}(k+j)\|  \leq  d_1 e^{ \frac{\lambda_1}{4 \epsilon} j\Delta t } \|\bs{e}_{\alpha}(k)\|+d_2$. Using approximation $(1+L_3 \Delta t)^{i-j-1} \approx 1+(i-j-1)L_3 \Delta t$ for small $L_3 \Delta t \ll 1$, we obtain the upper-bound as shown in the lemma. 

\vspace{-1mm}
\subsection{Proof of Lemma~\ref{err2_lemma}}
\label{proof_err2_lemma}

Substituting the iterative model similar to~(\ref{mean_var}) for both $\bs{\mu}_{{\theta}}(k+i|k)$ and $\bs{\theta}(k+i)$, the error calculation is reduced to
\begin{align*}
\bs{\theta}_\mu(k+i)=&\bs{F} \bs{\theta}_\mu(k+i-1)+\bs{G}[\bs{\mu}_{gp_{\theta}}(k+i-1|k)-\\
& \bs{f}_{\theta}(k+i-1)].
\end{align*}
Taking the norm on both sides of the above equation and using approximation $\|\bs{F}\|\approx 1$, $\|\bs{G}\|=\Delta t$, and assumption~(\ref{assumption_thetamu}), we obtain the iterative relationship of the error bound as $\|\bs{\theta}_\mu(k+i)\|\leq \|\bs{\theta}_\mu(k+i-1)\|+\Delta t\|\bs{\beta}_{\theta}^T \bs{\Sigma}_{gp_\theta}^{\frac{1}{2}}(k+i-1|k )\|$. With the initial condition $\bs{\mu}_{\theta}(k|k)=\bs{\theta}(k)$,  we then obtain that $\| \bs{\theta}_\mu(k+i)\| \leq  \Delta t \sum_{j=0}^{i-1} \|\bs{\beta}_{\theta}^T \bs{\Sigma}_{gp_\theta}^{\frac{1}{2}}(k+j|k)\|$. 

\vspace{-1mm}
\subsection{Proof of Lemma~\ref{Edifflemma}}
\label{proof_Edifflemma}

To assess $\bar{J}_{\hat{\theta}^*,\hat{W}^*}^{k}-\bar{J}_{\theta,\hat{W}^*}^k$, we use~(\ref{e_obj})-(\ref{e_obj2}) and obtain  
\begin{align}
&\bar{J}_{\hat{\theta}^*,\hat{W}^*}^{k}-\bar{J}_{\theta,\hat{W}^*}^k \nonumber \\
=& \sum_{i=0}^{H} \{\mathbb{E}[\| \bs{e}_{\hat{{\theta}}}(k+i|k)\|^2_{\bs{Q}_1}] 
	- \| \bs{e}_{\theta}(k+i)\|^2_{Q_1}  \}+ \nonumber \\
&\mathop{\mathbb{E}}[\|\bs{e}_{\hat{{\theta}}}(k+H+1)\|^2_{Q_3}] 
	- \|\bs{e}_{{\theta}}(k+H+1)\|^2_{Q_3} \nonumber \\
=&\sum_{i=0}^{H}\{\|\bs{e}_{\mu_{\hat{\theta}}}(k+i)\|^2_{Q_1} + \tr(\bs{Q}_1\bs{\Sigma}_{\hat{\theta}}(k+i))- \nonumber \\
& \| \bs{e}_{\theta}(k+i)\|^2_{Q_1}\}+\|\bs{e}_{\mu_{\hat{\theta}}}(k+H+1)\|^2_{Q_1}+ \nonumber\\
&\tr(\bs{Q}_3\bs{\Sigma}_{\hat{\theta}}(k+H+1))-\|\bs{e}_{\theta}(k+H+1)\|^2_{Q_3}\nonumber \\
=&\sum_{i=0}^{H}\{-\|\tilde{\bs{\theta}}_{\mu}(k+i)\|^2_{Q_1} + \tr(\bs{Q}_1\Sigma_{\hat{\theta}}(k+i))+ \nonumber\\
&2 \tilde{\bs{\theta}}^T_{\mu}(k+i) \bs{Q}_1\bs{e}_{\bs{\mu}_{\hat{\theta}}}(k+i) \}-\|\tilde{\bs{\theta}}_{\mu}(k+H+1)\|^2_{Q_3}+\nonumber \\
&\tr(\bs{Q}_3\bs{\Sigma}_{\hat{\theta}}(k+H+1))+ \nonumber \\
&2\tilde{\bs{\theta}}^T_{\mu}(k+H+1) \bs{Q}_3\bs{e}_{\mu_{\hat{\theta}}}(k+H+1). \label{Ediff}
\end{align}
The above last equality comes from the observation:
\begin{align*}
&\|\bs{e}_{\mu_{\hat{\theta}}}\|^2_{Q_1}-\| \bs{e}_\theta\|^2_{Q_1}=\|\bs{e}_{\mu_{\hat{\theta}}}\|^2_{Q_1}
- \| \bs{e}_{\mu_{\hat{\theta}}}-\tilde{\bs{\theta}}_\mu\|^2_{Q_1}\\
=&\|\bs{e}_{\mu_{\hat{\theta}}} \|^2_{Q_1}- \|\bs{e}_{\mu_{\hat{\theta}}}\|^2_{Q_1}-\|\tilde{\bs{\theta}}_\mu\|^2_{Q_1}
	+2\tilde{\bs{\theta}}^T_\mu\bs{Q}_1 \bs{e}_{\mu_{\hat{\theta}}}\\
=&-\|\tilde{\bs{\theta}}_\mu\|^2_{Q_1}
	+2\tilde{\bs{\theta}}^T_\mu\bs{Q}_1 \bs{e}_{\mu_{\hat{\theta}}}
\end{align*}
with $\bs{e}_{\mu_{\hat{\theta}}}=\bs{\mu}_{\hat{\theta}}-\bs{\theta}_d$, $\tilde{\bs{\theta}}_\mu=\bs{\mu}_{\hat{\theta}}-\bs{\theta}$ and $\bs{e}_{\theta}=\bs{\theta}-\bs{\theta}_d$. The rationale to use the above formulation is to put a bound on  $\bar{J}_{\hat{\theta}^*,\hat{W}^*}^{k}-J_{\theta,\hat{W}^*}^k$ by terms $\|\tilde{\bs{\theta}}_\mu\|$ and $\|\bs{e}_{\mu_{\hat{\theta}}}\|$.
	
Since $\lambda_{\max}(\bs{Q}_1)<\lambda_{\max}(\bs{Q}_3)$, from~(\ref{Ediff}), we have
\begin{align}
|&\bar{J}_{\hat{\theta}^*,\hat{W}^*}^{k}-\bar{J}_{\theta,\hat{W}^*}^k| \leq \lambda_{\max}(\bs{Q}_3) \sum_{i=0}^{H+1}  \bigl\{\|\tilde{\bs{\theta}}_\mu(k+i)\|^2 + \nonumber \\
&\tr(\bs{\Sigma}_{\hat{\theta}}(k+i))+ 2 \| \bs{e}_{\mu_{\hat{\theta}}}(k+i)\| \| \tilde{\bs{\theta}}_\mu(k+i) \| \bigr\}.
\label{cond55}
\end{align}
 From Lemma~\ref{mpc_lemma},  $\| \bs{e}_{\mu_{\hat{\theta}}}(k+i) \|\leq a_4(i)\|\bs{e}_{\theta}(k)\|+a_5(i)$. From Lemma~\ref{Sigma_theta_bound}, $\|\bs{\Sigma}_{\hat{\theta}}(k+i)\| \leq i(\Delta t)^2\sigma_{\bs{f}\max}^2$. Noting that matrix $\bs{\Sigma}_{\hat{\theta}}(k+i)$ is diagonal, we obtain $\tr(\bs{\Sigma}_{\hat{\theta}}(k+i)) \leq m\| \bs{\Sigma}_{\hat{\theta}}(k+i)\| \leq i m (\Delta t)^2\sigma_{\bs{f}\max}^2$. Adding the above upper bounds for each term in~(\ref{cond55}), we obtain that $ |\bar{J}_{\hat{\theta}^*,\hat{W}^*}^{k}-\bar{J}_{\theta,\hat{W}^*}^k| \leq \rho_J(\bs{e}_{\alpha},\bs{e}_{\theta})$, where $\rho_J(\bs{e}_{\alpha},\bs{e}_{\theta})$ is given by~(\ref{cond66}).  
	
\vspace{-1.5mm}
\subsection{Proof of Theorem~\ref{stability}}
\label{proof_stability}

First, it is straightforward to obtain the lower-bound of $V(k)$ by the fact that $V(k) \geq \lambda_{\min}(\bs{Q}_1) \|\bs{e}_{\theta}(k) \|^2+ \zeta \lambda_{\min}(\bs{Q}) \| \bs{e}_{\alpha}(k)\|^2\geq \underline{\lambda} \|\bs{e}(k)\|^2$ and similarly for upper-bound $V(k) \leq \overline{\lambda} \|\bs{e}(k)\|^2$. Therefore, we have
\begin{equation}
\underline{\lambda} \|\bs{e}(k)\|^2 \leq V(k) \leq \overline{\lambda} \|\bs{e}(k)\|^2.
\label{Vbound}
\end{equation}
From~(\ref{lya_decrease}), we obtain 
\begin{align}
\Delta V(k)& \leq |\bar{J}_{\hat{\theta}^*,\hat{W}^*}^{k+1}-\bar{J}_{\theta,\hat{W}^*}^{k+1}|+
|\bar{J}_{\hat{\theta}^*,\hat{W}^*}^{k}-\bar{J}_{\theta,\hat{W}^*}^k|+ \nonumber \\
&\zeta \left[V_\alpha(k+1)-V_\alpha(k)\right]+\nu \bigl[\|\bs{\Sigma}_d({\hat{W}^*}(k))\|- \nonumber \\
& \|\bs{\Sigma}_d({\hat{W}^*}(k+1))\|\bigr]+\bigl(J_{\hat{\theta}^*,\hat{W}^*}^{k+1}- {J}_{\hat{\theta}^*,\hat{W}^*}^{k}\bigr). \label{cond33}
\end{align}
We apply the results in Lemma~\ref{Edifflemma} to the first two terms in the above equation. For the third difference term, from Lemma~\ref{e_lemma}, we consider the discrete-time form of~(\ref{cond111}) 
\begin{align*}
V_{{\alpha}}(k+1)-V_{\alpha}(k) & \leq -\frac{1}{4}\Delta t \bs{e}_{\alpha}^T(k) \bs{Q}\bs{e}_{\alpha}(k) + c_3 \Delta t \\
&\leq -\frac{1}{4} \Delta t \lambda_{\min}(\bs{Q})\|\bs{e}_{\alpha}(k)\|^2 +  c_3 \Delta t.
\end{align*}
For the last two difference terms in~(\ref{cond33}), by~(\ref{J_decreasing}) we have
\begin{align*}
&J_{\hat{\theta}^*,\hat{W}^*}^{k+1}-J_{\hat{\theta}^*,\hat{W}^*}^{k}
	+\nu \bigl[\|\bs{\Sigma}_d({\hat{W}^*}(k))\|- \\
	& \|\bs{\Sigma}_d({\hat{W}^*}(k+1))\|\bigr] \\
\leq&  -\lambda_{\min} (\bs{Q}_1)\|\bs{e}_{\theta}(k)\|^2+\Delta \hat{\bs{\alpha}}^*_{Q_2k}+\\
&\nu \bigl[\|\bs{\Sigma}_{d}({\hat{W}^e}(k+1))\|- \|\bs{\Sigma}_d({\hat{W}^*}(k+1))\|\bigr]+ \\
&\tr(\bs{Q}_1 \bs{\Sigma}_{\hat{\theta}}(k+H+1) )+\tr(\bs{Q}_3 \bs{\Sigma}_{\hat{\theta}}(k+H+2) )\\
\leq &  -\lambda_{\min} (\bs{Q}_1)\|\bs{e}_{\theta}(k)\|^2 + \hat{\alpha}_{\max}^2 + \nu \sigma_{\bs{\kappa} \max}^2+ \\
&m \lambda_m(H+2)(\Delta t)^2 \sigma^2_{\bs{f}\max}.
\end{align*}
In the above last inequality, we use the facts that  $ \|\hat{\bs{\alpha}}^*(k+1)\|^2_{\bs{Q}_2} \leq \hat{\alpha}_{\max}^2$ and $\|\bs{\Sigma}_d({\hat{W}^e}(k+1))\| \leq \sigma_{\bs{\kappa} \max}^2$.	 
 	
Substituting the above derivations into~(\ref{cond33}), we obtain
\begin{align*}
\text{\hspace{-0mm}}\Delta V(k)\leq& \xi_1\|\bs{e}_{\alpha}(k)\|^2 + \xi_2\|\bs{e}_{\alpha}(k)\| \|\bs{e}_{\theta}(k)\| + \xi_3 \|\bs{e}_{\alpha}(k) \| \nonumber \\
&+\xi_4 \|\bs{e}_{\theta}(k)\| +\xi_5-\frac{\zeta}{4}\Delta t \lambda_{\min}(\bs{Q})\|\bs{e}_{\alpha}(k)\|^2 +\nonumber \\
& \zeta c_3 \Delta t-\lambda_{\min} (\bs{Q}_1)\|\bs{e}_{\theta}(k)\|^2+\hat{\alpha}_{\max}^2 + \nu \sigma_{\bs{\kappa}\max}^2  \nonumber \\
&+m \lambda_{m}(H+2)(\Delta t)^2 \sigma^2_{\bs{f}\max}\nonumber \\
=&-\frac{1}{2}\Bigl(\gamma_3\|\bs{e}_{\theta}(k)\|-\frac{\xi_2}{\gamma_3}\|\bs{e}_{\alpha}(k)\|\Bigr)^2 -\frac{\gamma_3^2}{4} \|\bs{e}(k)  \|^2\nonumber \\
& - \left(\gamma_1 \|\bs{e}_\alpha(k)\|- \gamma_2\right)^2 
-{\left(\frac{\gamma_3}{2} \|\bs{e}_{\theta}(k)\|-\gamma_4\right)^2}+\gamma_5
\label{v_decrease}
\end{align*}
if~(\ref{parameter_condition_theory}) is held. Considering the above result and~(\ref{Vbound}), we have 
$V(k+1)\leq \gamma_\lambda V(k)+\gamma_5$ and this proves the theorem.
	
\vspace{-2.5mm}
\section{MPC Implementation}
\label{mpc_implement}
	
The unconstrained MPC given in~(\ref{optimal_input}) is solved at each step by a gradient decent method. The gradient of the objective function with respect to the design variables is obtained by a back propagation approach. From~(\ref{MPC_cost}) and~(\ref{e_obj}), it is straightforward to obtain 
\begin{align*}
J^k_{\hat{\theta},\hat{W}}=&\sum_{i=0}^{H} l_s(k+i)+\|\hat{\bs{\alpha}}(k)\|^2_{Q_2}+l_f(k+H+1)\\
&+ \nu \|\bs{\Sigma}_d(k)\|
\end{align*}
The partial derivatives of $J^k_{\hat{\theta},\hat{W}}$ with respect to $\hat{\bs{\theta}}_{k+i}=\hat{\bs{\theta}}(k+i|k)$, $i=0,\ldots,H+1$, is obtained as 
\begin{equation*}
\frac{\partial J^k_{\hat{\theta},\hat{W}}}{\partial \hat{\bs{\theta}}_{k+i}}=
	\begin{cases}
	\frac{ \partial l_s(k+i)}{\partial \hat{\bs{\theta}}_{k+i}}, &\text{$i=0,\ldots,H$} \\
	\frac{ \partial l_f(k+H+1)}{\partial \hat{\bs{\theta}}_{k+i}} , & \text{$i=H+1$.}
	\end{cases}
\end{equation*}
Noting that the current state $\hat{\bs{\theta}}(k+i|k)$ affects the future states $\hat{\bs{\theta}}(k+i+1|k)$, the gradient calculation has the following backward iterative relationship
\begin{equation}
\frac{d J^k_{\hat{\theta},\hat{W}}}{d \hat{\bs{\theta}}_{k+i}}= \frac{\partial J^k_{\hat{\theta},\hat{W}}}{\partial \hat{\bs{\theta}}_{k+i}}+\frac{\partial J^k_{\hat{\theta},\hat{W}}}{\partial \hat{\bs{\theta}}_{k+i+1}}
	\frac{\partial \hat{\bs{\theta}}_{k+i+1}}{ \partial \hat{\bs{\theta}}_{k+i}}
\label{iter1}
\end{equation}
with the terminal condition $\frac{d J^k_{\hat{\theta},\hat{W}}}{d \hat{\bs{\theta}}_{k+H+1}}= \frac{\partial J^k_{\hat{\theta},\hat{W}}}{\partial \hat{\bs{\theta}}_{k+H+1}}$. The similar computation is applied to obtain the partial derivatives of $J^k_{\hat{\theta},\hat{W}}$ with respect to $\hat{\bs{w}}(k+i)$ and we omit here. We conduct the computation from the terminal condition at $i=K+H+1$ and then back propagate for $i=1,\ldots,H$ in~(\ref{iter1}).    

\vspace{-2.5mm}
\section{Physical Model for Two Platforms}

\subsection{Rotary inverted pendulum}
\label{invertedpendulum}

The rotary inverted pendulum dynamics model is obtained from Lagrangian mechanics and written in the form of (\ref{lagrangian_model}) with $\bs{q}=[\theta_1 \; \alpha_1 ]^T$, $\dot{\bs{q}}=[\theta_2 \; \alpha_2 ]^T$, $\bs{u}=V_m$ and
\begin{eqnarray*}
\bs{D}=\begin{bmatrix}
D_{1} & D_{3} \\ D_{3} & D_{2}
\end{bmatrix}, 
\bs{H}=\begin{bmatrix}
H_1 \\ H_2
\end{bmatrix},
\bs{B}=\begin{bmatrix}
1 \\ 0
\end{bmatrix}, 
\end{eqnarray*}
where
\begin{align*}
D_{1} =& m_pl_r^2+\frac{1}{4}m_pl_p^2\sin^2\alpha_1+J_r, \\
D_{3}=&-\frac{1}{2}m_p l_p l_r\cos\alpha_1, D_{2}=J_p+\frac{1}{4}m_p l_p^2, \\
H_1=&\frac{1}{2}m_p l_p^2\theta_2 \alpha_2\sin\alpha_1\cos\alpha_1 +\frac{1}{2} m_p l_p l_r\alpha_2^2 \sin\alpha_1 \\
&+d_r \theta_2+ k_g^2 k_t k_m/R_m \theta_2, \\
H_2=&-\frac{1}{4}m_p l_p^2\cos\alpha_1\sin\alpha_1 \theta_2^2+d_p \alpha_2 -\frac{1}{2}m_p l_pg\sin\alpha_1,
\end{align*}
$l_r$ and $J_r$ denote the base arm's length and moment of inertia, respectively, and $l_p$, $m_p$, and $J_p$ denotes the pendulum link's length, mass, and moment of inertia, respectively. Parameters $d_r$ and $d_p$ are the viscous damping coefficients of the base arm and pendulum joints, respectively. $k_g$, $k_t$, $k_m$, and $R_m$ are DC motor's electromechanical parameters~\cite{Apk2011}. It is straightforward to write the above dynamics in the form of~(\ref{robot_dyn_reshape}) with 
\begin{displaymath}
f_{\theta}(\theta_2,\bs{\alpha},u_d)=\frac{D_{2} V_m-H_1D_{2}+D_{3}H_2}{D_{1}D_{2}-D_{3}^2}
\end{displaymath} 
and $\kappa_{\alpha}(\theta_2,\bs{\alpha},\dot{\alpha}_2)= H_1-\frac{1 }{D_{3}}[(D_{3}+D_{1}D_{2}-D_{3}^2)\dot{\alpha}_2+H_2D_{1}]$.

\subsection{Autononmous bikebot}
\label{bikebotmodel}

The physical model of bikebot is obtained from Lagrangian machanics and the bicycle kinematics relationship. As shown in Fig.~\ref{bikebot_sketch}, considering a kinematic model and a nonholonomic constraint at rear contact point $C_2$, the motion equation of $C_2$ is written as 
\begin{subequations}
\begin{align}
& \ddot{X}= \dot{v}_c \cos\psi - v_c\sin\psi \dot\psi, \label{bikebot_analytic_model:a} \\
& \ddot{Y}= \dot{v}_c \sin\psi + v_c\cos\psi \dot\psi. \label{bikebot_analytic_model:b} 
\end{align}
\label{bikebot_analytic_model}
\end{subequations}
\hspace{-1.4mm}The yaw rate is calculated from the geometric relationship between the steering and rear frames as~\cite{WangTASE2019}
\begin{equation}
\dot{\psi}= \frac{v_c \cos\xi}{l \cos\varphi}\tan\phi
\label{bikebot_analytic_model:c}
\end{equation}
and the equation of roll motion is obtained as
\begin{equation}
J_t \ddot{\varphi}= -m_b h_b\frac{v_c^2}{l}\cos\xi \tan\phi + m_b h_b g \sin\varphi, 
\label{bikebot_analytic_model:d}
\end{equation}
where $m_b$ is the total mass of the bikebot, $l$ is the wheel base (i.e., distance between wheel contact points $C_1$ and $C_2$), $\xi$ is the steering casting angle, $h_b$ is the height of mass center, $J_t=m_b h_b^2+J_b$ is the mass moment of inertia of the bikebot along the $x$ axis of frame $\mathcal{R}$ ($J_b$ is the mass moment of inertia along the $x_b$ axis of body frame $\mathcal{B}$); see Fig.~\ref{bikebot_sketch}. From bikebot model~(\ref{bikebot_analytic_model}), the steering angle input $\phi$ affects both the balancing and the position tracking tasks.

Plugging~(\ref{bikebot_analytic_model:c}) into~(\ref{bikebot_analytic_model}) and combining with~(\ref{bikebot_analytic_model:d}), we obtain the dynamic model in~(\ref{bike}) with functions  
\begin{displaymath}
\bs{f}_{\theta}(\bs{\theta},\bs{\alpha},\bs{u})= \begin{bmatrix} 
u_f \cos\psi - \frac{v_c^2\sin\psi \cos\xi}{l \cos\varphi}\tan u_d \\
u_f \sin\psi + \frac{v_c^2\cos\psi \cos\xi}{l \cos\varphi}\tan u_d 
\end{bmatrix}
\end{displaymath}
and $\kappa_{\alpha}(\bs{\theta}, \bs{\alpha},\dot{\alpha}_2,u_f)=\arctan \left(\frac{m_b h_b g l \sin\varphi-J_t l \ddot\varphi}{m_b h_b v_c^2 \cos\xi}\right) -\ddot\varphi$. Note that in the above equations, yaw angle is calculated as $\psi=\atan2(\dot{Y},\dot{X})$ from state variable $\bs{\theta}_2$.

\bibliography{\refpath YiRef}

\end{document}